\documentclass{ieeetj}
\usepackage{cite}
\usepackage{amsmath,amssymb,amsfonts}
\usepackage{algorithmic}
\usepackage{graphicx,color}
\usepackage{textcomp}
\usepackage{xcolor}
\usepackage{hyperref}
\hypersetup{hidelinks=true}
\usepackage{algorithm,algorithmic}
\def\BibTeX{{\rm B\kern-.05em{\sc i\kern-.025em b}\kern-.08em
    T\kern-.1667em\lower.7ex\hbox{E}\kern-.125emX}}
\AtBeginDocument{\definecolor{tmlcncolor}{cmyk}{0.93,0.59,0.15,0.02}\definecolor{NavyBlue}{RGB}{0,86,125}}

\def\OJlogo{\vspace{-4pt}$<$Society logo(s) and publication title will appear here.$>$}
\def\seclogo{\vspace{10pt}$<$Society logo(s) and publication title will appear here.$>$}

\def\authorrefmark#1{\ensuremath{^{\textbf{#1}}}}

\usepackage{algorithm}
\usepackage{subfigure}
\usepackage{caption}
\usepackage{cite}

\begin{document}
\receiveddate{XX Month, XXXX}
\reviseddate{XX Month, XXXX}
\accepteddate{XX Month, XXXX}
\publisheddate{XX Month, XXXX}
\currentdate{XX Month, XXXX}
\doiinfo{XXXX.2022.1234567}

\markboth{}{Author {et al.}}

\title{Clustering of Time-Varying Graphs Based on Temporal Label Smoothness}

\author{Katsuki Fukumoto\authorrefmark{1}, Koki Yamada\authorrefmark{2}, Yuichi Tanaka\authorrefmark{3}, and Hoi-To Wai\authorrefmark{4}}
\affil{Graduate School of Bio-Applications and Systems Engineering, Tokyo University of Agriculture and Technology}
\affil{Department of Electrical Engineering, Tokyo University of Science}
\affil{Graduate School of Engineering, Osaka University}
\affil{Department of Systems Engineering and Engineering Management, The Chinese University of Hong Kong}
\corresp{Corresponding author: Yuichi Tanaka (email: ytanaka@comm.eng.osaka-u.ac.jp).}
\authornote{This work was supported in part by JST PRESTO under Grant JPMJPR1935 and JSPS KAKENHI under Grant 20H02145.}

\begin{abstract}
  We propose a node clustering method for time-varying graphs based on the assumption that the cluster labels are changed smoothly over time.
  Clustering is one of the fundamental tasks in many science and engineering fields including signal processing, machine learning, and data mining.
  Although most existing studies focus on the clustering of nodes in static graphs, we often encounter time-varying graphs for time-series data, e.g., social networks, brain functional connectivity, and point clouds.
  In this paper, we formulate a node clustering of time-varying graphs as an optimization problem based on spectral clustering, with a smoothness constraint of the node labels.
  We solve the problem with a primal-dual splitting algorithm.
  Experiments on synthetic and real-world time-varying graphs are performed to validate the effectiveness of the proposed approach.
\end{abstract}

\begin{IEEEkeywords}
Spectral clustering, graph signal processing, time-varying graphs, point clouds.
\end{IEEEkeywords}

\maketitle

\section{INTRODUCTION}
\IEEEPARstart{C}{lustering} is an important and fundamental technique in signal processing, machine learning, and data mining \cite{xuSurveyClusteringAlgorithms2005,saxenaReviewClusteringTechniques2017,minSurveyClusteringDeep2018}.
Features of the data points belonging to the same community are expected to have similar characteristics and vice versa.
That is, the intra-community relationship is stronger than that for the inter-community.
Such an underlying relationship is often given by a graph, where nodes of the graph correspond to the data points, and the relationships among the data points are given by edges.
Therefore, clustering nodes of the graph is an important and widely studied problem \cite{nascimento2011spectral,schaeffer2007graph,rossettiCommunityDiscoveryDynamic2018}.

Many methods have been proposed for graph node clustering \cite{shiNormalizedCutsImage2000,ngSpectralClusteringAnalysis2002,newmanFastAlgorithmDetecting2004,newmanFindingEvaluatingCommunity2004,newmanFindingCommunityStructure2006,shenDetectOverlappingHierarchical2009}.
In this paper, we focus on unsupervised learning where it is assumed that we have a graph, but we do not have training data associated with it.

Unsupervised clustering can be classified into two approaches.
One is the node-domain approach, where the nodes are clustered based on several node-wise features.
Its examples include clustering based on a goodness-of-partition metric called modularity \cite{newmanFastAlgorithmDetecting2004}.
The other is the spectral-domain approach, where the node clustering is performed based on the spectral characteristics of graphs, e.g., the polarities of elements in eigenvectors of a graph operator, e.g., graph Laplacian \cite{vonluxburgTutorialSpectralClustering2007}.
Its representative method is the well-known spectral clustering (SC) \cite{whiteSpectralClusteringApproach2005}.
Note that the node- and spectral-domain approaches are related to each other.
For example, SC with normalized graph Laplacian is an approximation of the minimum cut problem \cite{shiNormalizedCutsImage2000,ngSpectralClusteringAnalysis2002}.

These existing node clustering methods are mostly designed for static graphs.
However, we often encounter time-varying (TV) graphs, i.e., a set of graphs where each graph represents a relationship between nodes at a certain time slot \cite{yamadaTimevaryingGraphLearning2020,casteigtsTimevaryingGraphsDynamic2012}.
Examples of TV graphs include social networks \cite{zhangDiffusionSocialNetworks2014}, brain functional connectivity \cite{pretiDynamicFunctionalConnectome2017}, and point clouds\cite{wang2019dynamic}.
TV graph clustering is a crucial problem since TV graphs are found in many applications mentioned above: For example, clustering of social networks can help us understand human behavior \cite{tabassum2018social},
that of brain functional connectivity could contribute to understanding the brain activity \cite{lynn2019physics}, and 
that of point clouds could be utilized for autonomous driving \cite{shi2020point}.

The static approach can be used for TV graph clustering by applying it independently at each time slot, however, this does not capture the temporal relationship or evolution of the clusters.
Nevertheless, there exist few clustering methods for TV graph clustering, and most existing algorithms are designed in an ad-hoc extension of the static methods (please see Section \ref{rel}-\ref{TVGC} for details) \cite{hopcroftTrackingEvolvingCommunities2004,aynaudStaticCommunityDetection2010,sunGraphscopeParameterfreeMining2007,liuGlobalSpectralClustering2018,karaaslanliConstrainedSpectralClustering2020}.

In this paper, we propose a TV graph clustering method as an extension of the (static) SC for capturing the smooth temporal evolution of clusters.
To capture temporal evolution, we assume a \textit{label smoothness} property: only a small number of nodes changed their cluster membership at each time slot.
Broadly speaking, time-series data points could change their relationships smoothly over time.
For example, friendships, environments measured by sensor networks, and pixel values in videos, may change gradually over time.
Feature values in machine learning may also gradually change over time in a high-dimensional space.
Therefore, it is natural to assume their memberships, i.e., clusters, to be smoothly changed (or not-so-frequently changed) over time as well as their feature values.
To reflect such a smoothness assumption, we propose a new regularization term that is incorporated with the proposed SC formulation.
We then rewrite the problem so that it can be solved using a primal-dual splitting algorithm \cite{condatPrimalDualSplitting2013}.

We perform our proposed TV graph clustering for synthetic graphs and real-world point cloud data.
In both cases, the proposed method efficiently clusters TV graphs compared to the existing several static and TV graph clustering methods.

In our preliminary study \cite{fukumoto2021node}, we proposed a basic methodology for the proposed TV graph clustering.
In this paper, we significantly extend the preliminary version by showing a relationship between the static version of SC and the proposed TV graph clustering, reformulating the objective function, and performing comprehensive experiments.

The remainder of this paper is organized as follows.
Notation used throughout the paper is defined in the following.
Section \ref{rel} describes related works for graph clustering.
The proposed method is described in Section \ref{pro}.
Section \ref{exp} provides experimental results with synthetic and real-world data.
Finally, we conclude the paper in Section \ref{con}.

\subsection*{Notation}
 In this paper, we consider a weighted undirected graph $\mathcal{G} = (V, E, \mathbf{W})$, where $V := \{v_0, v_1, \dots\}$ is a set of nodes, $E$ is a set of edges, and $\mathbf{W} \in \mathbb{R}^{N\times N}$ is a weighted adjacency matrix.
The number of nodes is given by $N = |V|$.
Each element of the weighted adjacency matrix is defined as
\begin{equation}
[\mathbf{W}]_{i j}=\begin{cases}
w_{i j} & \text { if } v_i \text { and } v_j \text { are connected, } \\
0 & \text { otherwise.}
\end{cases}
\end{equation}
That is, $w_{i j} \geq 0$ is the weight of the edge between $v_i$ and $v_j$.
In addition, the degree of the $i$th node, $d_i$, is given by
\begin{equation}
  d_i = \sum^{N}_{j=1} w_{ij}
\end{equation}
where $\mathbf{D} := \text{diag}(d_0, d_1, \dots)$ is called a degree matrix.
Using $\mathbf{D}$ and $\mathbf{W}$, a combinatorial graph Laplacian is given by
\begin{equation}
  \mathbf{L} = \mathbf{D} - \mathbf{W}.
\end{equation}

A graph signal is defined as $f: V \longrightarrow \mathbb{R}$ \cite{shuman2013emerging,ortega2018graph}.
This can be written by a vector $\mathbf{f} \in \mathbb{R}^N$ whose $i$th element, $f_i$, is a signal value at the node $v_i$.
The quadratic form of the graph Laplacian is given by
\begin{equation}
  \label{quad}
  \mathbf{f}^{\top} \mathbf{L} \mathbf{f}=\frac{1}{2} \sum_{i, j=1}^{N} w_{i j}\left(f_{i}-f_{j }\right)^{2}.
\end{equation}
This is widely used for a smoothness measure of the graph signal $\mathbf{f}$ on $\mathcal{G}$ \cite{ortega2018graph}.

\section{RELATED WORK}
\label{rel}
 In this section, we review existing works for static and TV graph clustering especially focusing on SC \cite{whiteSpectralClusteringApproach2005} since our work is based on the SC algorithm.

\subsection{SPECTRAL CLUSTERING}
\label{rel_sc}
 Here, we describe the formulation of SC for static graphs \cite{vonluxburgTutorialSpectralClustering2007}.
The goal of clustering $V$ of $\mathcal{G}$ is to divide the nodes so that they are strongly connected within the same cluster and are weakly connected between different clusters.

First, let us define the strength of the connection between clusters.
For a set of nodes $A \subset V$ and $\bar{A} := V\setminus A$, the connection strength between $A$ and $\bar{A}$ is defined as
\begin{equation}
  \label{waa}
c(A, \bar{A}):=\sum_{i \in A, j \in \bar{A}} w_{i j}.
\end{equation}
For a given $K$, clustering should be performed by choosing a partition of $A_1,...,A_K$ which minimizes \textit{Cut} given by
\begin{equation}
  \label{cut}
\operatorname{Cut}\left(A_{1}, \ldots, A_{K}\right):=\frac{1}{2} \sum_{k=1}^{K} c\left(A_{k}, \bar{A}_{k}\right).
\end{equation}

We often need to balance the number of nodes in $|A_i|$ to avoid very small clusters.
For this purpose, RatioCut is proposed in \cite{hagenNewSpectralMethods1992, vonluxburgTutorialSpectralClustering2007}.
Its cost function is defined as follows:
\begin{equation}
  \label{rcut}
\begin{aligned}
\operatorname{RatioCut}\left(A_{1}, \ldots, A_{K}\right) :=&\sum_{k=1}^{K} \frac{\operatorname{Cut}\left(A_{k}, \bar{A}_{k}\right)}{\left|A_{k}\right|}.
\end{aligned}
\end{equation}

Hereafter, we assume $K=2$ for simplicity.
This can be straightforwardly generalized for $K>2$.
In RatioCut, the problem to be solved is formulated by using \eqref{rcut} as follows:
\begin{equation}
  \label{minrcut}
\min _{A \subset V} \operatorname{RatioCut}(A, \bar{A}).
\end{equation}
This problem is combinatorial and NP-hard \cite{wagnerMinCutGraph1993}.
Therefore, many relaxation methods to solve \eqref{minrcut} have been proposed \cite{hagenFastSpectralMethods1991,chanSpectralKwayRatiocut1994,roxboroughGraphClusteringUsing1997}.

The typical relaxation is based on Laplacian quadratic form in \eqref{quad} \cite{whiteSpectralClusteringApproach2005}.
First, we define a vector $\mathbf{c} \in \mathbb{R}^N$ which we call the cluster vector as follows:

\begin{equation}
  \label{fi}
c_i=\begin{cases}
\sqrt{|\bar{A}| /|A|} & \text { if } v_{i} \in A, \\
-\sqrt{|A| /|\bar{A}|} & \text { if } v_{i} \in \bar{A}.
\end{cases}
\end{equation}
When we can find an appropriate $\mathbf{c}$, clustering can be immediately done by using the polarity of $c_i$.
Furthermore, we can relate the Laplacian quadratic form \eqref{quad} to the RatioCut in \eqref{rcut} as follows:
\begin{equation*}
\begin{split}
\mathbf{c}^{\top} \mathbf{L} \mathbf{c} =& \frac{1}{2} \sum_{i, j=1}^{N} w_{i j}\left(c_i-c_j\right)^{2} \\
=&\frac{1}{2} \sum_{i \in A, j \in \bar{A}} w_{i j}\left(\sqrt{\frac{|\bar{A}|}{|A|}}+\sqrt{\frac{|A|}{|\bar{A}|}}\right)^{2} \\
&+\frac{1}{2} \sum_{i \in \bar{A}, j \in A} w_{i j}\left(-\sqrt{\frac{|\bar{A}|}{|A|}}-\sqrt{\frac{|A|}{|\bar{A}|}}\right)^{2} \\
=& \operatorname{cut}(A, \bar{A})\left(\frac{|A|+|\bar{A}|}{|A|}+\frac{|A|+|\bar{A}|}{|\bar{A}|}\right) \\
=&|V| \cdot \operatorname{RatioCut}(A, \bar{A}).
\end{split}
\end{equation*}
Note that $\mathbf{c}$ satisfies

\begin{align}
\sum_{i=1}^{N} \mathbf{c}_i &=\sum_{i \in A} \sqrt{\frac{|\bar{A}|}{|A|}}-\sum_{i \in \bar{A}} \sqrt{\frac{|A|}{|\bar{A}|}} \nonumber \\
&=|A| \sqrt{\frac{|\bar{A}|}{|A|}}-|\bar{A}| \sqrt{\frac{|A|}{|\bar{A}|}}=0
\end{align}
and
\begin{equation}
\|\mathbf{c}\|_2^{2}=\sum_{i=1}^{N} c_i^{2}=|A| \frac{|\bar{A}|}{|A|}+|\bar{A}| \frac{|A|}{|\bar{A}|}=|\bar{A}|+|A|=N.
\end{equation}
These results in that $\mathbf{c}$ is orthogonal to $\mathbf{1}$ and $\|\mathbf{c}\|_2^2 = N$.

Consequently, \eqref{minrcut} can be rewritten as follows:
\begin{equation}
\begin{aligned}
\label{beforeapx}
&\min _{\mathbf{c} \in \mathbb{R}^{N}, A \subseteq V} \mathbf{c}^{\top} \mathbf{L} \mathbf{c}\\
&\text { subject to } \mathbf{c} \perp \mathbf{1},\  \mathbf{c}_i \text { in \eqref{fi}, } \|\mathbf{c}\|_2^2=N.
\end{aligned}
\end{equation}
However, it is still combinatorial and NP-hard because the values in $\mathbf{c}$ are binary in \eqref{fi}.
To approximately solve \eqref{beforeapx}, the condition on $\mathbf{c}$ is relaxed to have arbitrary real values.
This leads to the following optimization problem.
\begin{equation}
  \label{sc}
\min _{\mathbf{c} \in \mathbb{R}^{N}, A \subseteq V} \mathbf{c}^{\top} \mathbf{L} \mathbf{c} \quad \text {subject to } \mathbf{c} \perp \mathbf{1},\  \|\mathbf{c}\|_2^2=N.
\end{equation}
It is well known that the solution of \eqref{sc} is given by \textit{Fiedler vector}, the eigenvector corresponding to the second smallest eigenvalue of $\mathbf{L}$.
After solving \eqref{sc}, the cluster labels can be obtained by using the polarity of $\mathbf{c}$.

For the case of $K>2$, we need to find $\ell$ cluster vector $\{\mathbf{c}^{(\ell)} \}_{\ell=1}^{K}$.
In this case, solutions are given as $K$ eigenvectors of $\mathbf{L}$ corresponding to the smallest eigenvalues. Then, $K$-means clustering is performed for the set of eigenvectors.

\subsection{TV GRAPH CLUSTERING}
\label{TVGC}
 TV graph clustering can be classified into three main categories \cite{rossettiCommunityDiscoveryDynamic2018}.
The first method consists of a two-step process.
In the two-step methods, first, clustering is performed at each time step.
Then, for each step, the clusters at step $t$ are aligned with the clusters at step $t-1$.
As an example, a method to track the evolution of TV clusters based on agglomerative clustering has been proposed \cite{hopcroftTrackingEvolvingCommunities2004}.
Another clustering method in this category has been proposed based on SC \cite{liuGlobalSpectralClustering2018}.
The method first computes the eigenvectors of the graph in all time slots and then performs smoothing of the eigenvectors along the temporal direction.
The advantage of this type of method is that static clustering methods can be used without modification.
However, this does not take into account temporal evolution.

The second method is to iteratively compute the clusters of the TV graph.
As an example, a method based on the Louvin method \cite{blondel2008fast} is proposed \cite{aynaudStaticCommunityDetection2010}.
In this method, clustering at $t$ is repeated using the Louvin method using the clusters obtained at  $t-1$ as the initial value.
Another clustering method in the second category formulates optimization problems so that the clusters change smoothly \cite{gong2012community}.
This method maximizes modularity in the graph at each time slot as well as maximizes the normalized mutual information \cite{lancichinetti2008benchmark} of clusters at successive time slots.
This approach aims to reflect the temporal dependency in clustering like our proposed approach, however, the initial cluster at $t=0$ highly affects the resulting TV clusters.

The third method transforms a set of TV graphs into a single large network and then performs clustering for static graphs. 
As an example, a method based on the dynamic stochastic block model (SBM) is proposed in \cite{yang2009bayesian,yang2011detecting}.
This method extends the static SBM to a dynamic method.
Specifically, the evolution of clusters is captured by modeling the cluster transitions for each node and by estimating the parameters of the model.
However, this does not properly reflect temporal evolution like the first approach.

In contrast to existing approaches, our method obtains clusters of the entire TV graphs by solving an optimization problem as an extension of the static SC.
We describe this in the following section.

\section{TV GRAPH CLUSTERING BASED ON LABEL SMOOTHNESS}
\label{pro}
In this section, we formulate a node clustering problem for TV graphs based on the label smoothness assumption.
First, we extend an SC problem for TV graphs.
Second, its optimization algorithm is introduced.

\subsection{FORMULATION}

As mentioned previously, SC methods are mostly designed for static graphs, therefore, it is difficult to perform node clustering taking into account temporal evolution.
To tackle the problem, we extend SC for TV graphs.

Suppose that the number of clusters is $K=2$ (it will be relaxed later in Section \ref{pro}-\ref{ext}) and we have a set of TV graphs, $\{\mathcal{G}_t = (V_t, E_t, \mathbf{W}_t)\}_{t=1}^T$, where $\mathcal{G}_t$ is a graph at time instance $t$.
We assume that the number of nodes in $\mathcal{G}_t$ is the same $N$ for all $t$ and node registration has already been performed.
That is, node mappings among time instances are already known.

Without considering the temporal smoothness, \eqref{minrcut} may be simply extended to handle TV graphs as follows:
\begin{equation}
  \label{sc2}
  \min_{A_t \subset V_t} \sum^{T}_{t=1} \operatorname{RatioCut}(A_t, \bar{A}_t).
\end{equation}
However, \eqref{sc2} does not reflect the temporal variation of node labels.
This may lead to all node labels between $\mathcal{G}_t$ and $\mathcal{G}_{t-1}$ being significantly different.

Here, we assume that the cluster labels change smoothly over time.
This is reasonable as long as the time-series data are obtained with a sufficiently high sampling frequency.
Based on the label smoothness assumption, the \textit{label mismatch function} between clusters $A_t$ and $A_{t-1}$ can be defined as:

\begin{align}
  \label{mismatch}
  \operatorname{mismatch}(A_t,A_{t-1}) & = \sum^{N}_{i=1}u_i,
\end{align}
where
\begin{align}
      u_i & =\begin{cases}
  1 & \text { if } v_i \in A_t \text{ and } v_i \notin A_{t-1}, \\
  0 & \text { otherwise }.
  \end{cases}
\end{align}
Eq.~\eqref{mismatch} counts the number of nodes included in cluster $A_t$ but not in cluster $A_{t-1}$.
Temporal variations of cluster labels can be smoothed by promoting
\eqref{mismatch} smaller.

By combining \eqref{sc2} and \eqref{mismatch}, we formulate a TV graph clustering based on SC as follows:
\begin{equation}
\begin{aligned}
  \label{NPhard}
  \min_{A \subset V} &\sum^{T}_{t=1} \operatorname{RatioCut}(A_t, \bar{A}_t) \\
  &+
  \alpha \sum^{T}_{t=2} \left(\operatorname{mismatch}(A_t,A_{t-1}) + \operatorname{mismatch}(\bar{A}_t,\bar{A}_{t-1})\right).
\end{aligned}
\end{equation}
where $\alpha \in \mathbb{R}_{\ge 0}$ is a parameter.
As clearly observed, the optimization problem in \eqref{NPhard} is NP-hard as in the classical SC problem \eqref{minrcut}.
In the following subsection, we relax the problem into a more computationally-friendly form.

\subsection{RELAXATION}

In the first step of the relaxation, we define a vector $\mathbf{c}_t \in \mathbb{R}^N$ like the static SC as follows:
\begin{equation}
  \label{cti}
[\mathbf{c}_t]_i=\begin{cases}
\sqrt{|\bar{A}_t| /|A_t|} & \text { if } v_{i} \in A_t, \\
-\sqrt{|A_t| /|\bar{A}_t|} & \text { if } v_{i} \in \bar{A}_t.
\end{cases}
\end{equation}
To rewrite the $\operatorname{mismatch}(\cdot)$ using $\mathbf{c}_t$, the difference vector between $\mathbf{c}_t$ and $\mathbf{c}_{t-1}$ can be written as follows:
\begin{align}
  \label{c}
  [\mathbf{c}_t-\mathbf{c}_{t-1}]_i & =\begin{cases}
  \sqrt{|\bar{A_t}| /|A_t|} - \sqrt{|\bar{A}_{t-1}| /|A_{t-1}|} \\ \qquad \qquad \text { if } v_i \in A_t \text { and } v_i \in A_{t-1}, \\
  \sqrt{|\bar{A_t}| /|A_t|} - (-\sqrt{|A_{t-1}| /|\bar{A}_{t-1}|}) \\ \qquad \qquad \text { if } v_i \in A_t \text { and } v_i \in \bar{A}_{t-1}, \\
  -\sqrt{|A_t| /|\bar{A}_t|} - \sqrt{|\bar{A}_{t-1}| /|A_{t-1}|} \\ \qquad \qquad \text { if } v_i \in \bar{A}_t \text { and } v_i \in A_{t-1}, \\
  -\sqrt{|A_t| /|\bar{A}_t|} - (-\sqrt{|A_{t-1}| /|\bar{A}_{t-1}|}) \\ \qquad \qquad \text { if } v_i \in \bar{A}_t \text { and } v_i \in \bar{A}_{t-1}. \\
  \end{cases}
\end{align}
Based on the label smoothness assumption, only a few nodes change their membership between $t-1$ and $t$.
Therefore, we can approximate the number of nodes in clusters at successive time slots as follows:
\begin{align}
    \label{Aapx}
    |A_t| \approx |A_{t-1}|,\ |\bar{A}_t| \approx |\bar{A}_{t-1}|.
\end{align}
By substituting \eqref{Aapx} into \eqref{c}, it can be rewritten as
\begin{align}
  \label{cnext}
  [\mathbf{c}_t-\mathbf{c}_{t-1}]_i \sim [\mathbf{d}_t]_i := \begin{cases}
  0 \hspace{5mm} \text { if } v_i \in A_t \text { and } v_i \in A_{t-1}, \\
  2\sqrt{|\bar{A_t}| /|A_t|} \\ \qquad \text { if } v_i \in A_t \text { and } v_i \in \bar{A}_{t-1}, \\
  -2\sqrt{|A_t| /|\bar{A}_t|} \\ \qquad \text { if } v_i \in \bar{A}_t \text { and } v_i \in A_{t-1}, \\
  0 \hspace{5mm} \text { if } v_i \in \bar{A}_t \text { and } v_i \in \bar{A}_{t-1}. \\
  \end{cases}
\end{align}
It can be observed that $[\mathbf{d}_t]_i$ becomes nonzero only if $v_i$ changes its membership.
By using \eqref{cnext}, $\operatorname{mismatch}(\cdot)$ can be rewritten using $\mathbf{c}_t$ as follows:
\begin{equation}
    \label{misrewrite}
    \operatorname{mismatch}(A_t,A_{t-1}) + \operatorname{mismatch}(\bar{A}_t,\bar{A}_{t-1}) = \|\mathbf{d}_t\|_0,
\end{equation}
where $\|\cdot\|_0$ represents the $\ell_0$ pseudo norm.

As a result, we can rewrite \eqref{NPhard} by using \eqref{misrewrite} and the approximation of RatioCut shown in Section \ref{rel}-\ref{rel_sc} as follows:
\begin{equation}
  \label{sctime0}
\begin{aligned}
  &\min _{\mathbf{c}_t \in \mathbb{R}^{N}} \sum_{t=1}^{T} \mathbf{c}_t^{\top} \mathbf{L}_t \mathbf{c}_t + \alpha \sum_{t=2}^T\|\mathbf{d}_t\|_0 \\
  &\text { subject to } \mathbf{c}_t \perp \mathbf{1},\|\mathbf{c}_t\|_2^2=N.
\end{aligned}
\end{equation}
This is a natural extension of RatioCut for TV graphs.
Note that solving \eqref{sctime0} directly requires a combinatorial optimization due to the $\ell_0$ norm constraint.
Instead, we consider to solve the following relaxed optimization problem using the $\ell_1$ norm regularization:
\begin{equation}
  \label{sctime}
\begin{aligned}
  &\min _{\mathbf{c}_t \in \mathbb{R}^{N}} \sum_{t=1}^{T} \mathbf{c}_t^{\top} \mathbf{L}_t \mathbf{c}_t + \alpha\sum_{t=2}^T\|\mathbf{d}_t\|_1 \\
  &\text { subject to } \mathbf{c}_t \perp \mathbf{1},\|\mathbf{c}_t\|_2^2=N,
\end{aligned}
\end{equation}
where $\|\cdot\|_1$ represents the $\ell_1$ norm.

We further rewrite \eqref{sctime} by introducing the concatenated label vector $\mathbf{c}_{\text{all}}=[\mathbf{c}_1^{\top}, \mathbf{c}_2^{\top}, \dots, \mathbf{c}_T^{\top}]^{\top} \in \mathbb{R}^{NT}$ for a simple expression as follows.
\begin{equation}
  \label{sctime2}
\min _{\mathbf{c}_{\text{all}} \in \mathbb{R}^{NT}} \mathbf{c}_{\text{all}}^{\top} \mathbf{\mathbf{L}} \mathbf{c}_{\text{all}} + \alpha\|\boldsymbol{\Phi} \mathbf{c}_{\text{all}}\|_1 \quad \text {subject to } \mathbf{c}_t \perp \mathbf{1},\|\mathbf{c}_t\|_2^2=N,
\end{equation}
where $\mathbf{L}:= \text{diag}(\mathbf{L}_{1}, \dots, \mathbf{L}_{T})$ and $\boldsymbol{\Phi}$ is a linear operator satisfying $\boldsymbol{\Phi} \mathbf{c}_{\text{all}} = \mathbf{c}_{\text{all}} - \hat{\mathbf{c}}_{\text{all}}$, in which $\hat{\mathbf{c}}_{\text{all}} = [\mathbf{c}_1^{\top}, \mathbf{c}_1^{\top}, \mathbf{c}_2^{\top}, \dots, \mathbf{c}_{T-1}^{\top}]^{\top}$.
In the following subsection, we present an algorithm to solve \eqref{sctime2}.

\subsection{OPTIMIZATION}
Here, we rewrite \eqref{sctime2} with a more tractable form to an existing solver.
First, \eqref{sctime2} is rewritten as follows by introducing indicator functions:

\begin{equation}
  \label{sctime3}
  \begin{aligned}
  \min_{\mathbf{c}_{\text{all}} \in \mathbb{R}^{NT}} &\frac{1}{2}\mathbf{c}_{\text{all}}^{\top} \mathbf{L} \mathbf{c}_{\text{all}} + \alpha\|\boldsymbol{\Phi} \mathbf{c}_{\text{all}}\|_1 \\ 
  & + \sum_{j=1}^T \left\{\iota_{A_{\mathbf{1} \epsilon}}([\mathbf{c}_{\text{all}}]_j) + \iota_{B_{N}}([\mathbf{c}_{\text{all}}]_j)\right\},
  \end{aligned}
\end{equation}
where $[\mathbf{c}_{\text{all}}]_j$ corresponds to $\mathbf{c}_j$ and the indicator functions, $\iota_{A_{\mathbf{1} \epsilon}}$ and $\iota_{B_{N}}$, are respectively defined as follows:
\begin{align}
\label{iotaA}
\iota_{A_{\mathbf{1} \epsilon}}(\mathbf{x}) & =\begin{cases}
0 & \text { if } \mathbf{x} \in A_{\mathbf{1} \epsilon}, \\
\infty & \text { otherwise, }
\end{cases}\\
\iota_{B_{N}}(\mathbf{x}) & =\begin{cases}
0 & \text { if } \mathbf{x} \in B_{N}, \\
\infty & \text { otherwise, }
\end{cases}
\end{align}
in which
\begin{align}
  A_{\mathbf{1} \epsilon} &= \{\mathbf{x} \in \mathbb{R}^{N} | |\mathbf{x}^{\top} \mathbf{1}| \leq \epsilon\}, \\
  \label{bn}
  B_{N} &= \{\mathbf{x} \in \mathbb{R}^{N} | \|\mathbf{x}\|_2^2-N=0\}.
\end{align}
$\iota_{A_{\mathbf{1} \epsilon}}$ and $\iota_{B_{N}}$ represent constraints in \eqref{sctime2}.

Note that \eqref{sctime3} is a nonconvex optimization problem since $B_N$ in \eqref{bn} is a nonconvex set.
However, we can practically solve the problem using a popular primal-dual splitting (PDS) algorithm \cite{condatPrimalDualSplitting2013}.
The PDS algorithm tackles a problem in the following form:
\begin{equation}
\min _{\mathbf{c}} f_1(\mathbf{c})+f_2(\mathbf{c})+f_3(\mathbf{Mc}),
\end{equation}
where $f_1$ is a differentiable convex function with the $\beta$-Lipschitzian gradient $\nabla f_1$ for some $\beta > 0$; $f_2$ and $f_3$ are proper lower semicontinuous convex functions which are proximable; and $\mathbf{M}$ is a linear operator.

To apply PDS on \eqref{sctime3}, the latter is split into the following PDS-applicable forms:
\begin{align*}
  f_1(\mathbf{c}) &:= \mathbf{c}^{\top} \mathbf{L} \mathbf{c} \quad \text{with } \beta = \lambda_{\max}(\mathbf{L}), \\
  f_2(\mathbf{c}) &:= \sum_{j=1}^T \iota_{B_{N}}([\mathbf{c}]_j), \\
  f_3(\mathbf{d}) &:= \sum_{j=1}^T \iota_{A_{\mathbf{1} \epsilon}}([\mathbf{d}_1]_j) + \alpha\| \mathbf{d}_2\|_1, \\
  \mathbf{M} &= \begin{bmatrix}
  \mathbf{I}\\
  \mathbf{\Phi}
  \end{bmatrix},
\end{align*}
where $\lambda_{\max}(\mathbf{L})$ is the maximum eigenvalue of $\mathbf{L}$, and $\mathbf{d} := \mathbf{Mc} = [\mathbf{d}_1^{\top}, \mathbf{d}_2^{\top}]^{\top}$ is the dual variable.

The proximal operator of $f_2$ is given by
\begin{equation}
  \operatorname{prox}_{\gamma \iota_{B_{N}}} ([\mathbf{z}]_j) = \sqrt{\frac{N}{[\mathbf{z}]_j^{\top} [\mathbf{z}]_j}}[\mathbf{z}]_j.
\end{equation}
Moreover, the proximal operator of $f_3$ is calculated by dividing it into two terms.
The first term, $\sum_{j=1}^T \iota_{A_{\mathbf{1} \epsilon}}([\mathbf{d}_1]_j)$, has the following proximal operator
\begin{equation}
\operatorname{prox}_{\gamma \iota_{A_{\mathbf{1} \epsilon}}}([\mathbf{z}]_j)=
\begin{cases}
[\mathbf{z}]_j - \frac{[\mathbf{z}]_j^{\top}\mathbf{1} - \epsilon}{[\mathbf{z}]_j^{\top}[\mathbf{z}]_j} & \text { if } [\mathbf{z}]_j^{\top}\mathbf{1} > \epsilon, \\ \relax
[\mathbf{z}]_j - \frac{[\mathbf{z}]_j^{\top}\mathbf{1} + \epsilon}{[\mathbf{z}]_j^{\top}[\mathbf{z}]_j} & \text { if } [\mathbf{z}]_j^{\top}\mathbf{1} < -\epsilon, \\ \relax
[\mathbf{z}]_j & \text { otherwise. }
\end{cases}
\end{equation}
Furthermore, the proximal operator of the second term, $\|\cdot\|_1$, is known to be the element-wise soft-thresholding operation \cite{parikhProximalAlgorithms2014}:

\begin{equation}
\left[\operatorname{prox}_{\gamma\|\cdot\|_{1}}(\mathbf{z})\right]_{i}=\operatorname{sgn}\left(z_{i}\right) \max \left\{0,\left|z_{i}\right|-\gamma\right\}.
\end{equation}
The details of the algorithm are shown in Algorithm \ref{alg1} where $\sigma$ is a small real value.

\begin{algorithm}
\caption{TV Clustering with Label Smoothness}
\label{alg1}
\begin{algorithmic}
\REQUIRE $\mathbf{c}^{(0)},\mathbf{d}_1^{(0)},\mathbf{d}_2^{(0)}$
\ENSURE $\mathbf{c}^{(i)}$
\WHILE{$\|\mathbf{c}^{(i+1)}-\mathbf{c}^{(i)}\|/\|\mathbf{c}^{(i)}\|>\sigma$}
\STATE $\mathbf{c}^{(i+1)} = \operatorname{prox}_{\gamma_1 \iota_{B_N(\cdot)}}(\mathbf{c}^{(i)} - \gamma_1 (\mathbf{L} \mathbf{c}^{(i)} + \mathbf{d}_1^{(i)} + \mathbf{\Phi}^{\top} \mathbf{d}_2^{(i)}))$
\STATE $\mathbf{d}_1^{(i+1)} = \operatorname{prox}_{\gamma_2 \iota_{A_{\mathbf{1} \epsilon}}(\cdot)^*} (\mathbf{d}_1^{(i)}+\gamma_{2} (2 \mathbf{c}^{(i+1)}-\mathbf{c}^{(i)}))$
\STATE $\mathbf{d}_2^{(i+1)} = \operatorname{prox}_{\gamma_2 \| \cdot \|_1^*} (\mathbf{d}_2^{(i)}+\gamma_{2} \Phi (2 \mathbf{c}^{(i+1)}-\mathbf{c}^{(i)}))$
\ENDWHILE
\end{algorithmic}
\end{algorithm}

Although we observed that the algorithm works well in practice \cite{condat2015cadzow,andersson2014new}, a formal convergence analysis is left for future work.

\subsection{EXTENSION TO ARBITRARY NUMBER OF CLUSTERS}
\label{ext}
 In the previous subsection, we assume $K=2$.
However, the number of clusters is often greater than two.
In this subsection, we describe the proposed method for clustering graphs with arbitrary $K$.

To split $\{\mathcal{G}_t\}_{t=1}^{T}$ into multiple clusters, we need to have $\{\mathbf{c}_t^{(\ell)} \}_{\ell=1}^{K}$ as those required in the static SC.
Suppose that ${\mathbf{c}_t^{(1)}, \ldots, \mathbf{c}_t^{(\ell)}, \ldots, \mathbf{c}_t^{K}}$, is obtained before calculating the $\ell$th cluster vector $\mathbf{c}^{(\ell)}_t$.
$\mathbf{c}^{(\ell)}_t$ is defined as follows:
\begin{equation}
\label{clt}
[\mathbf{c}_t^{(\ell)}]_i=\begin{cases}
\sqrt{|[A_{\ell}]_t|} & \text { if } v_{i} \in [A_{\ell}]_t, \\
0 & \text { otherwise, }
\end{cases}
\end{equation}
where $[A_{\ell}]_t$ is the $\ell$th cluster at time $t$.
With the spirit of SC, we need to solve the following problem:
\begin{equation}
\label{extsctime}
\begin{aligned}
  &\min _{\mathbf{c}_t \in \mathbb{R}^{N}} \sum_{t=1}^{T} \mathbf{c}_t^{\top} \mathbf{L}_t \mathbf{c}_t + \alpha\sum_{t=2}^{T}\|\mathbf{c}_t - \mathbf{c}_{t-1}\|_1 \\
  &\text { subject to } \quad \mathbf{c}_t \perp \{\mathbf{c}^{(1)}_t, \dots, \mathbf{c}^{(\ell)}_t\}, \|\mathbf{c}_t\|_2^2=N, \mathbf{c}^{(\ell)}_t \text{ in } \eqref{clt}.
\end{aligned}
\end{equation}
As with the static SC, the condition on $\mathbf{c}$ is relaxed to have arbitrary real values.
This leads to the following optimization problem.
\begin{equation}
\label{extsctime}
\begin{aligned}
  &\min _{\mathbf{c}_t \in \mathbb{R}^{N}} \sum_{t=1}^{T} \mathbf{c}_t^{\top} \mathbf{L}_t \mathbf{c}_t + \alpha\sum_{t=2}^{T}\|\mathbf{c}_t - \mathbf{c}_{t-1}\|_1 \\
  &\text { subject to } \quad \mathbf{c}_t \perp \{\mathbf{c}^{(1)}_t, \dots, \mathbf{c}^{(\ell)}_t\}, \|\mathbf{c}_t\|_2^2=N.
\end{aligned}
\end{equation}

The indicator function $\iota_{A_{\mathbf{1} \epsilon}}(\cdot)$ in \eqref{iotaA} is redefined as follows so that it can be computed for any vectors:

\begin{equation}
\iota_{A_{\mathbf{v} \epsilon}}(\mathbf{x})=\begin{cases}
0 & \text { if } \mathbf{x} \in A_{\mathbf{v} \epsilon}, \\
\infty & \text { otherwise, }
\end{cases}
\end{equation}
where
\begin{equation*}
  A_{\mathbf{v} \epsilon} = \{\mathbf{x} \in \mathbb{R}^{N} | |\mathbf{x}^{\top} \mathbf{v}| \leq \epsilon\}.
\end{equation*}

The proximal operator of $\iota_{A_{\mathbf{1} \epsilon}}(\cdot)$ is given by
\begin{equation}
\operatorname{prox}_{\gamma \iota_{A_{\mathbf{v} \epsilon}}}([\mathbf{z}]_j)=
\begin{cases}
[\mathbf{z}]_j - \frac{[\mathbf{z}]_j^{\top}\mathbf{v} - \epsilon}{[\mathbf{z}]_j^{\top}[\mathbf{z}]_j} \mathbf{v} & \text { if } [\mathbf{z}]_j^{\top}\mathbf{v} > \epsilon, \\ \relax
[\mathbf{z}]_j - \frac{[\mathbf{z}]_j^{\top}\mathbf{v} + \epsilon}{[\mathbf{z}]_j^{\top}[\mathbf{z}]_j} \mathbf{v} & \text { if } [\mathbf{z}]_j^{\top}\mathbf{v} < -\epsilon, \\ \relax
[\mathbf{z}]_j & \text { otherwise. }
\end{cases}
\end{equation}

The algorithm to solve \eqref{extsctime} is similar to Algorithm \ref{alg1}.
We first determine $\{\mathbf{c}_t^{(1)}\}$ by Algorithm \ref{alg1}, and the others ${\mathbf{c}_t}_{\ell=2}^{(K-1)}$'s are sequentially computed.

\section{EXPERIMENTS}
\label{exp}

\begin{figure}[t]
\centering \relax
\includegraphics[width=80mm]{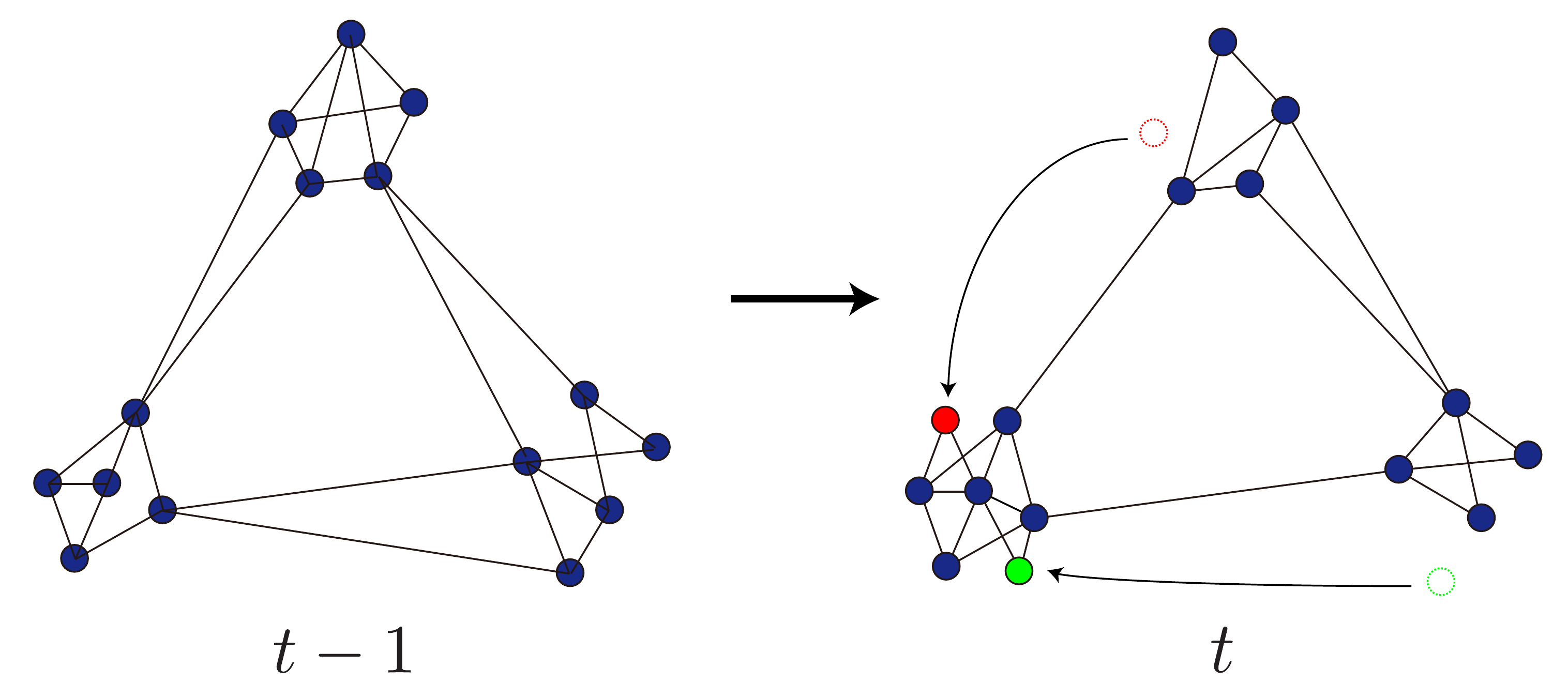}
\caption{Overview of synthetic TV graphs. The red and green nodes at $t-1$ moved to different clusters at $t$, and the nodes are reconnected with the SBM model.}
\label{fig:generate}
\end{figure}

\begin{figure*}[t]
    \centering
    \subfigure[Graph at $t=10$]{
        \includegraphics[clip, width=0.32\columnwidth]
        {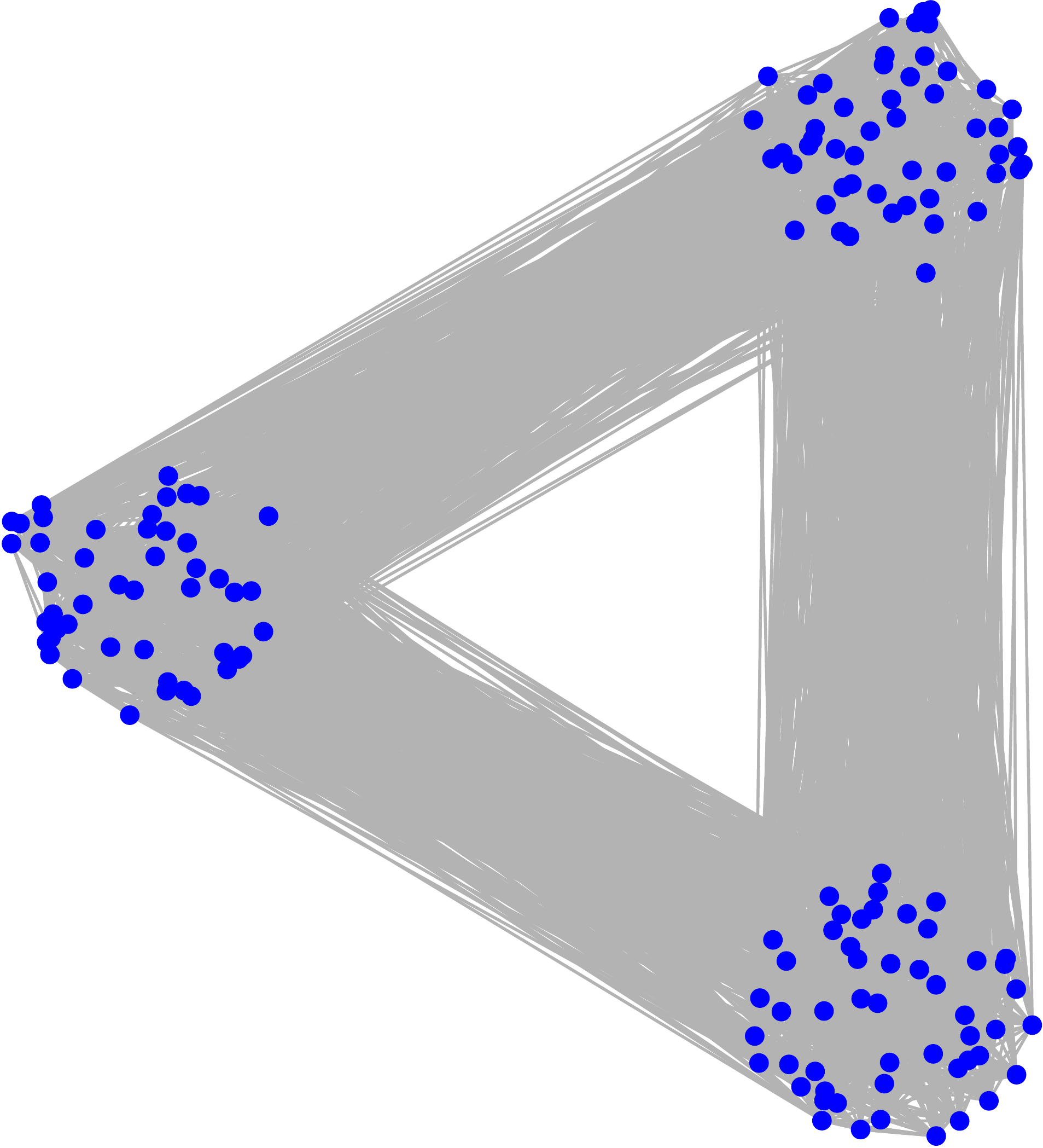}
        }
        \hspace{10pt}
    \subfigure[SC for static graph]{
        \includegraphics[clip, width=0.32\columnwidth]
        {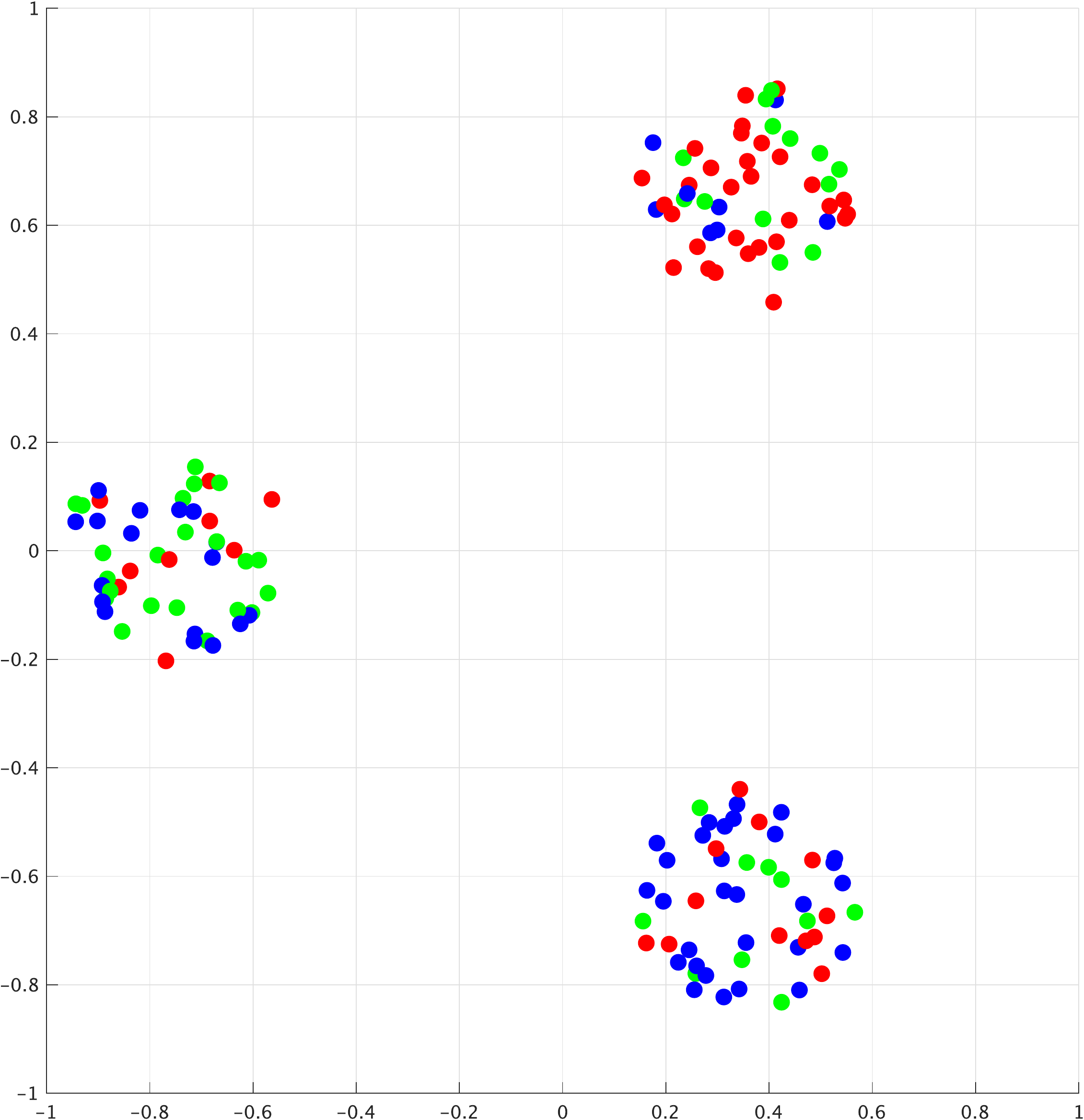}
        }
        \hspace{10pt}
    \subfigure[PisCES]{
        \includegraphics[clip, width=0.32\columnwidth]
        {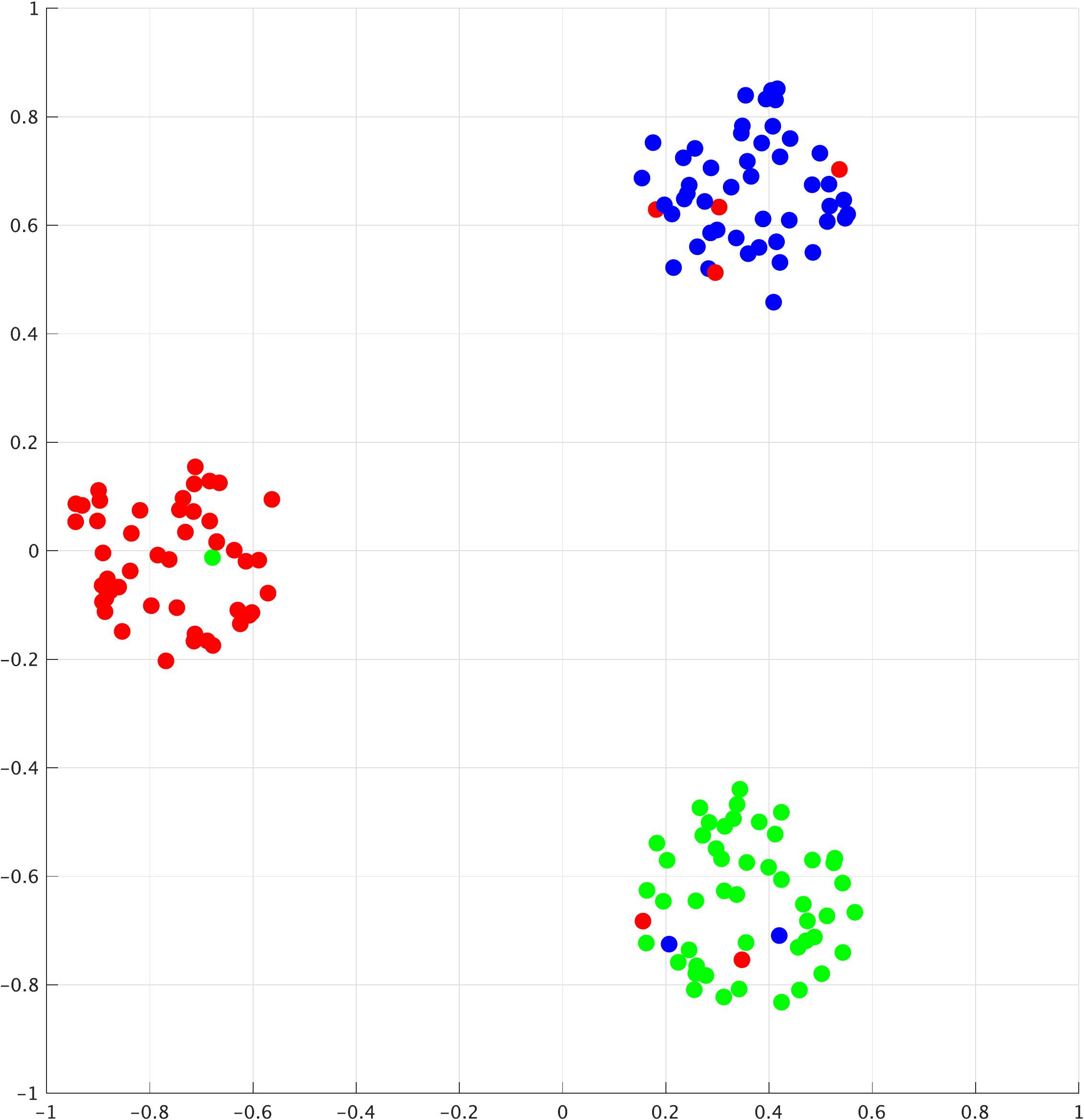}
        }
        \hspace{10pt}
    \subfigure[Proposed method]{
        \includegraphics[clip, width=0.32\columnwidth]
        {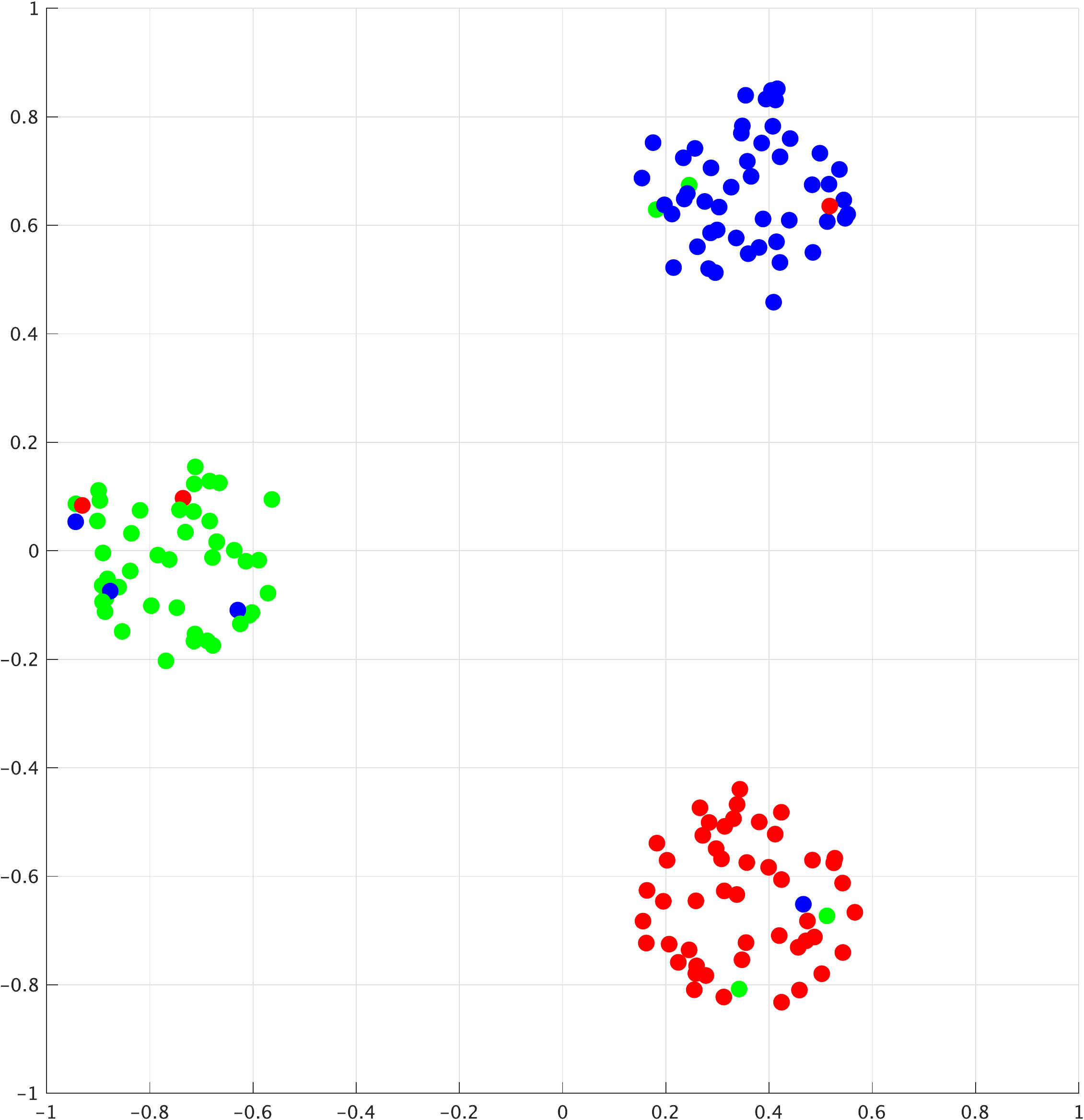}
        }
\\
    \centering
    \subfigure[Graph at $t=50$]{
        \includegraphics[clip, width=0.32\columnwidth]
        {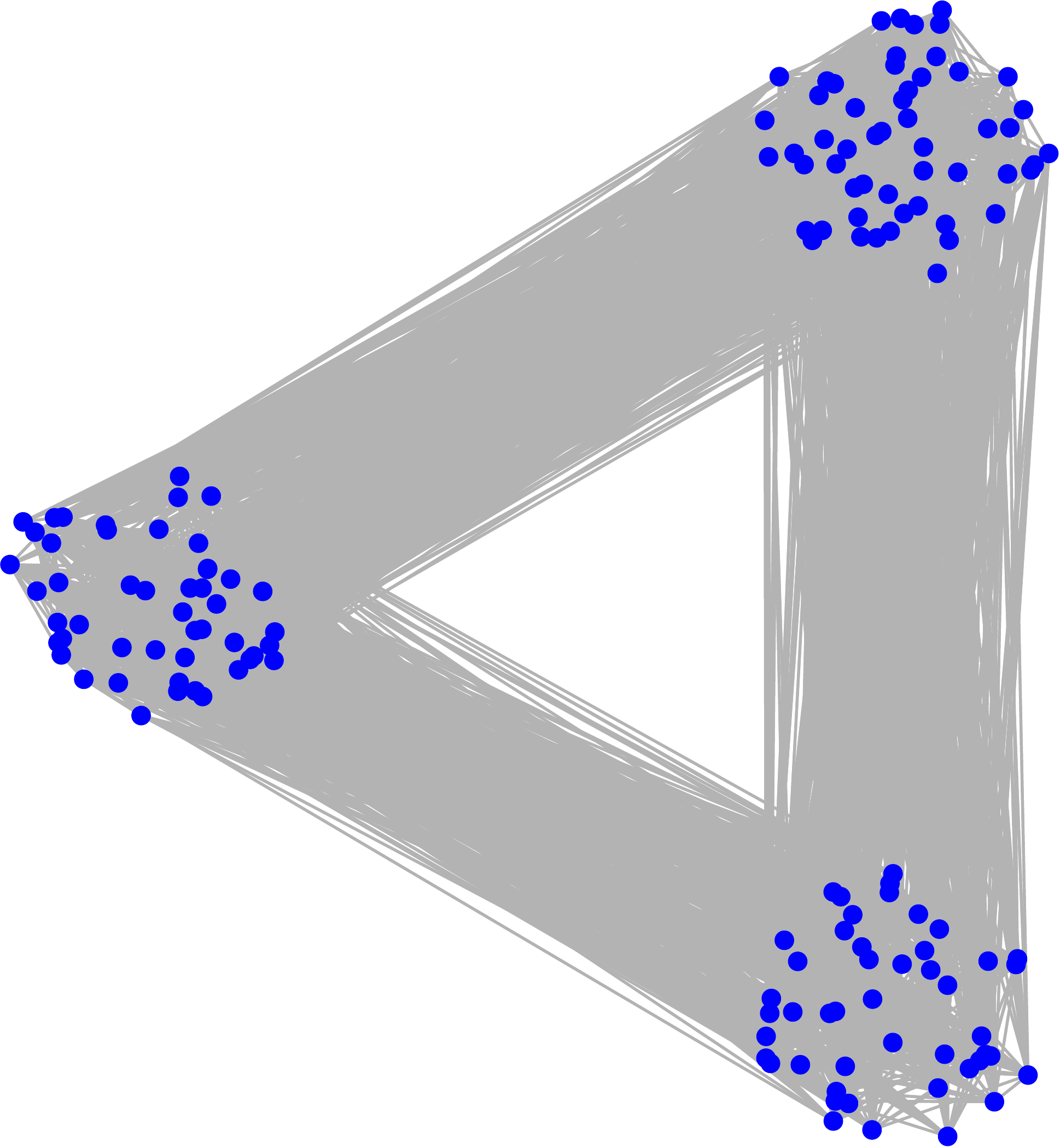}
        }
        \hspace{10pt}
    \subfigure[SC for static graph]{
        \includegraphics[clip, width=0.32\columnwidth]
        {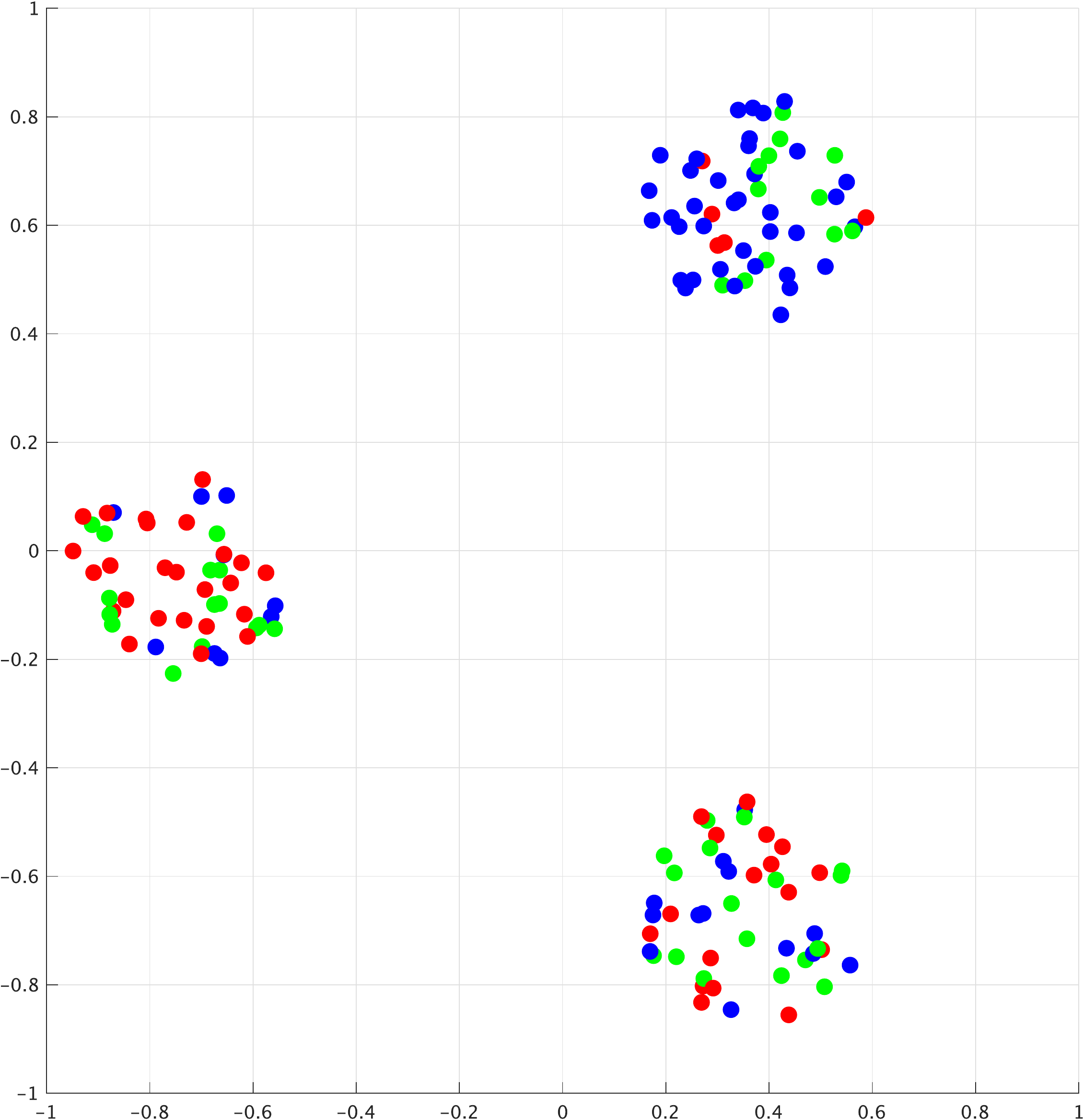}
        }
        \hspace{10pt}
    \subfigure[PisCES]{
        \includegraphics[clip, width=0.32\columnwidth]
        {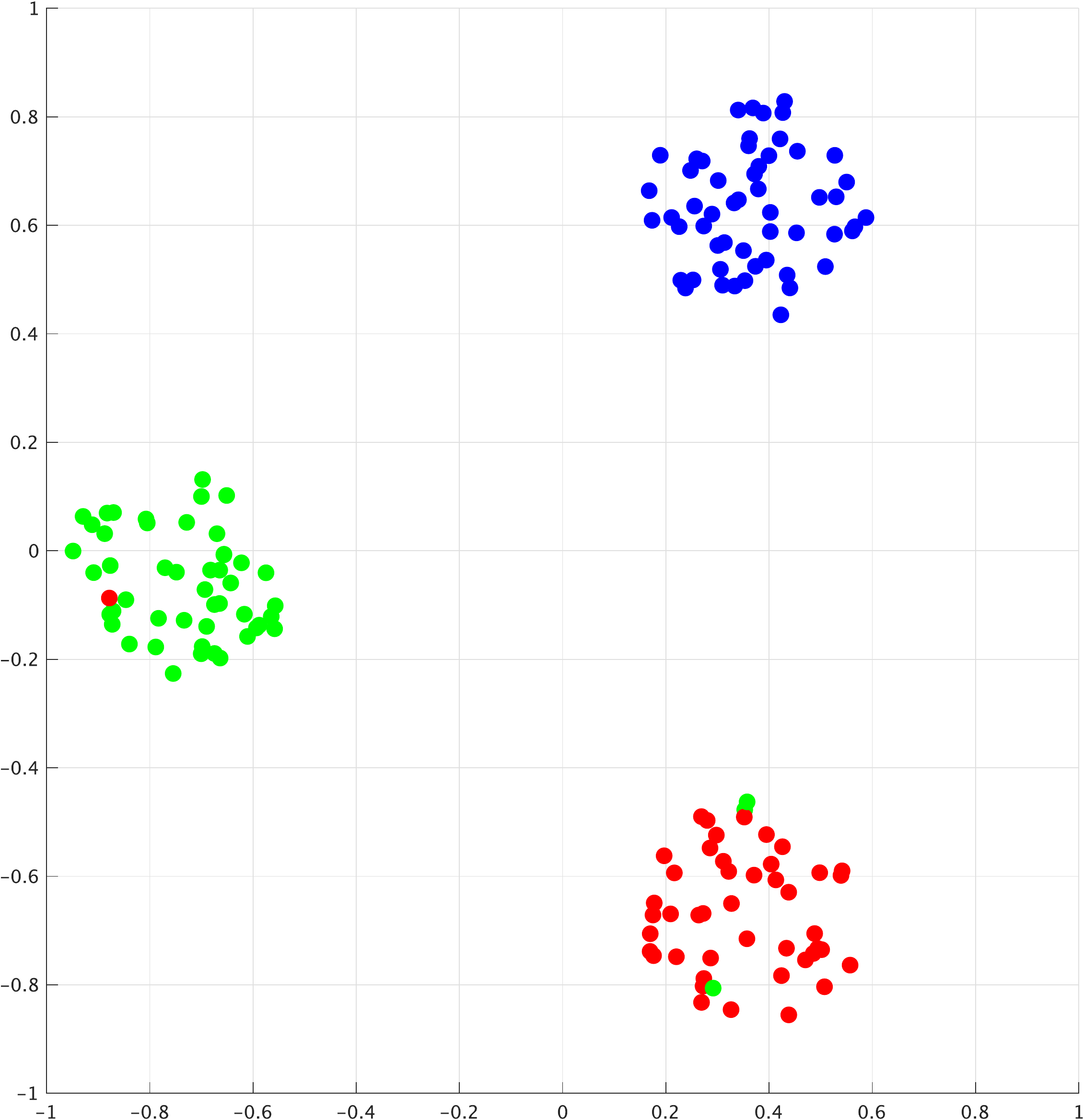}
        }
        \hspace{10pt}
    \subfigure[Proposed method]{
        \includegraphics[clip, width=0.32\columnwidth]
        {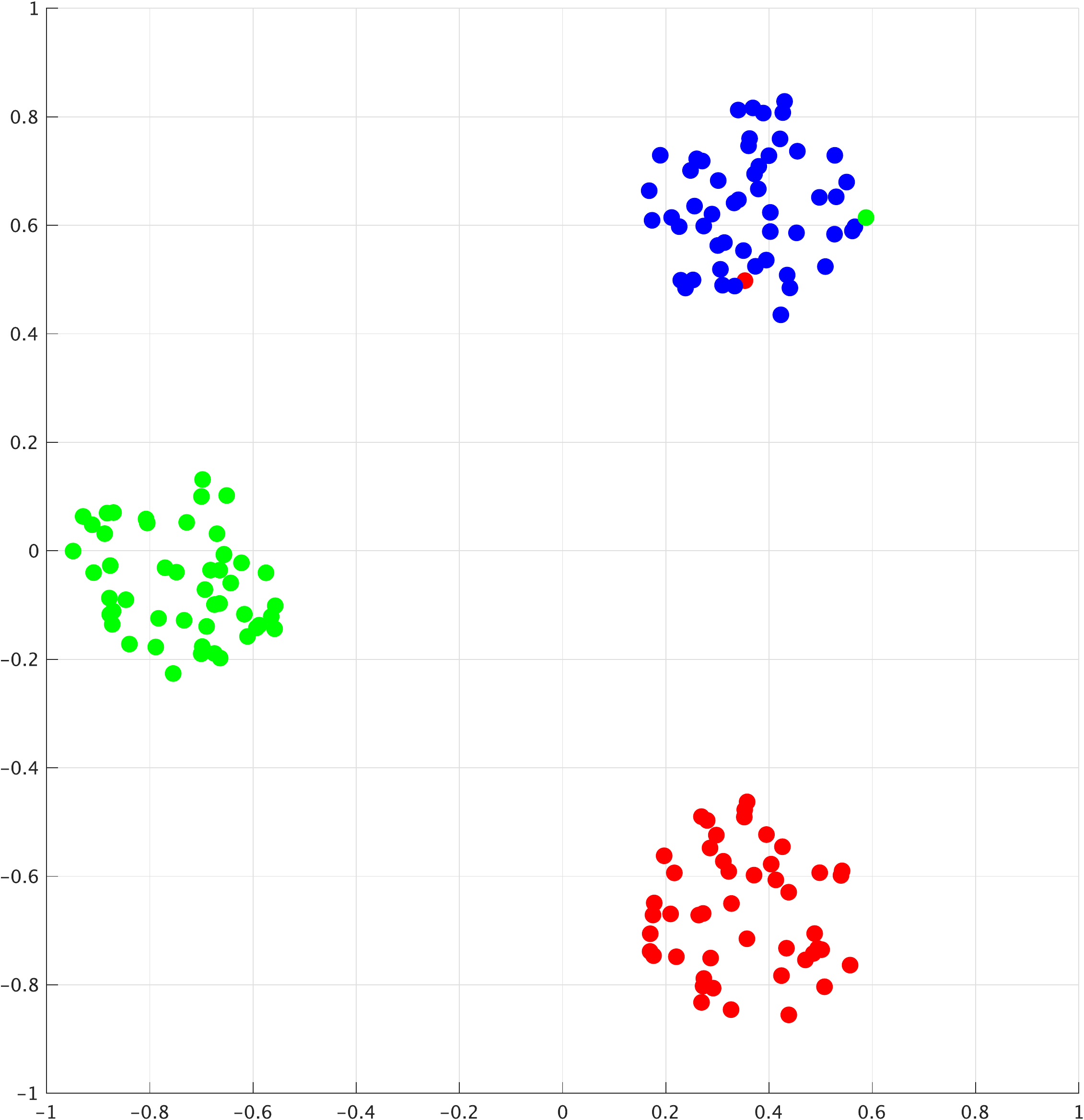}
        }
    \caption{Experimental results of the dense synthetic data. Top row: $t=10$. Bottom row: $t=50$.}
    \label{fig:sy1}
\end{figure*}

\begin{figure}[t]
\centering \relax
\includegraphics[width=80mm]{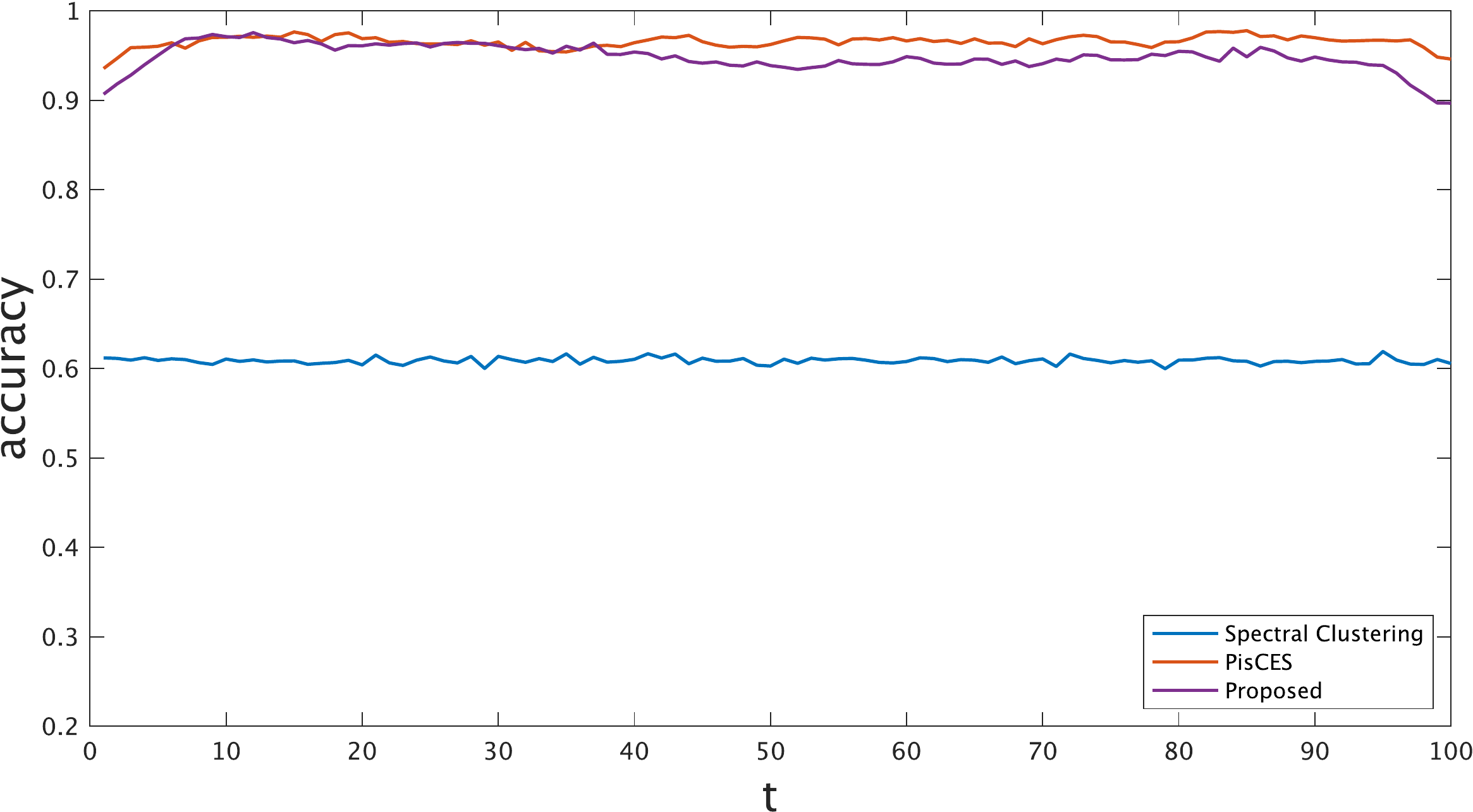}
\caption{Clustering accuracy for dense TV graphs. Averages for 100 trials are shown.}
\label{fig:accdense}
\end{figure}

\begin{figure*}[t]
    \centering
    \subfigure[Graph at $t=1$]{
        \includegraphics[clip, width=0.32\columnwidth]
        {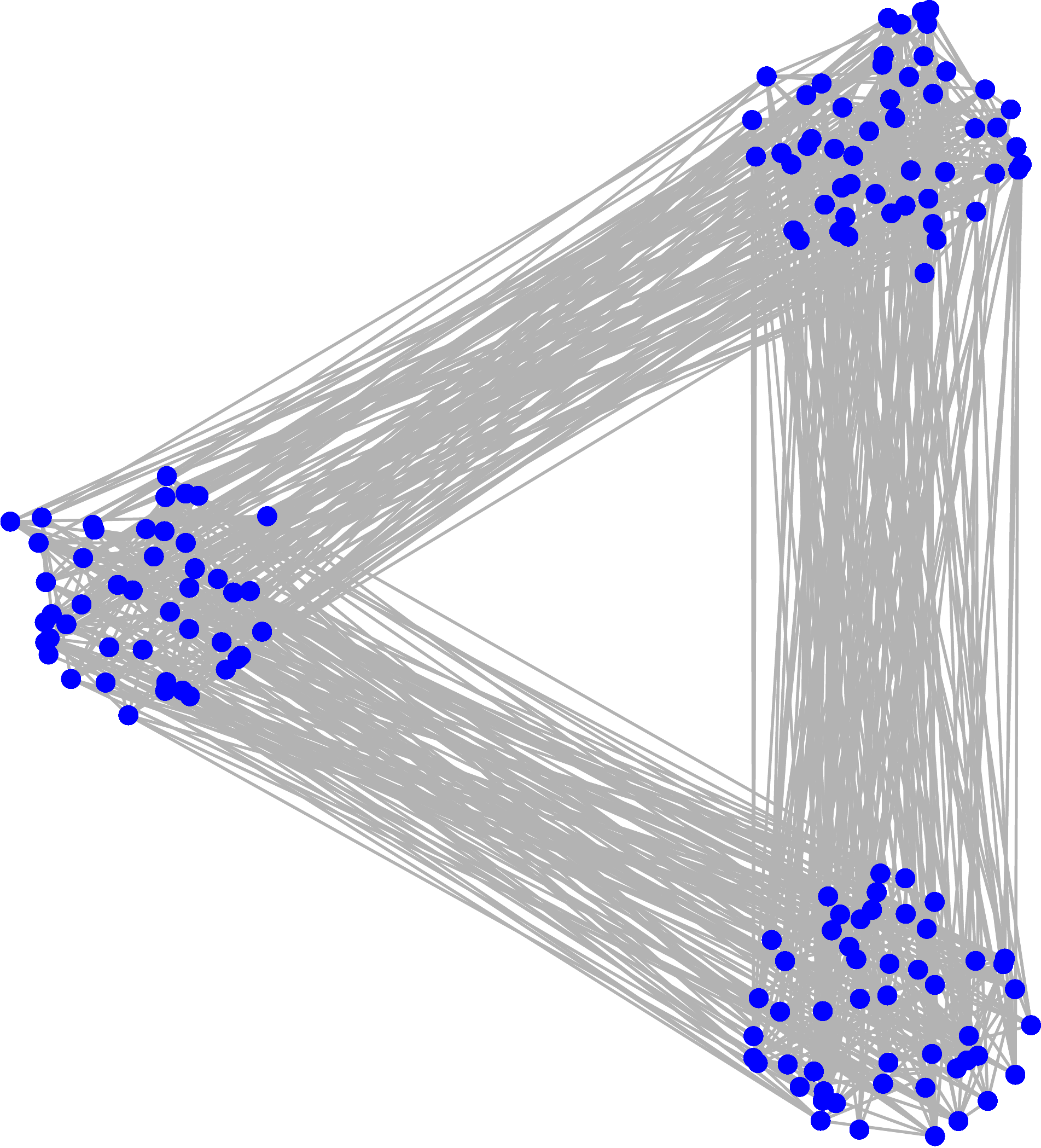}
        }
        \hspace{10pt}
    \subfigure[SC for static graph]{
        \includegraphics[clip, width=0.32\columnwidth]
        {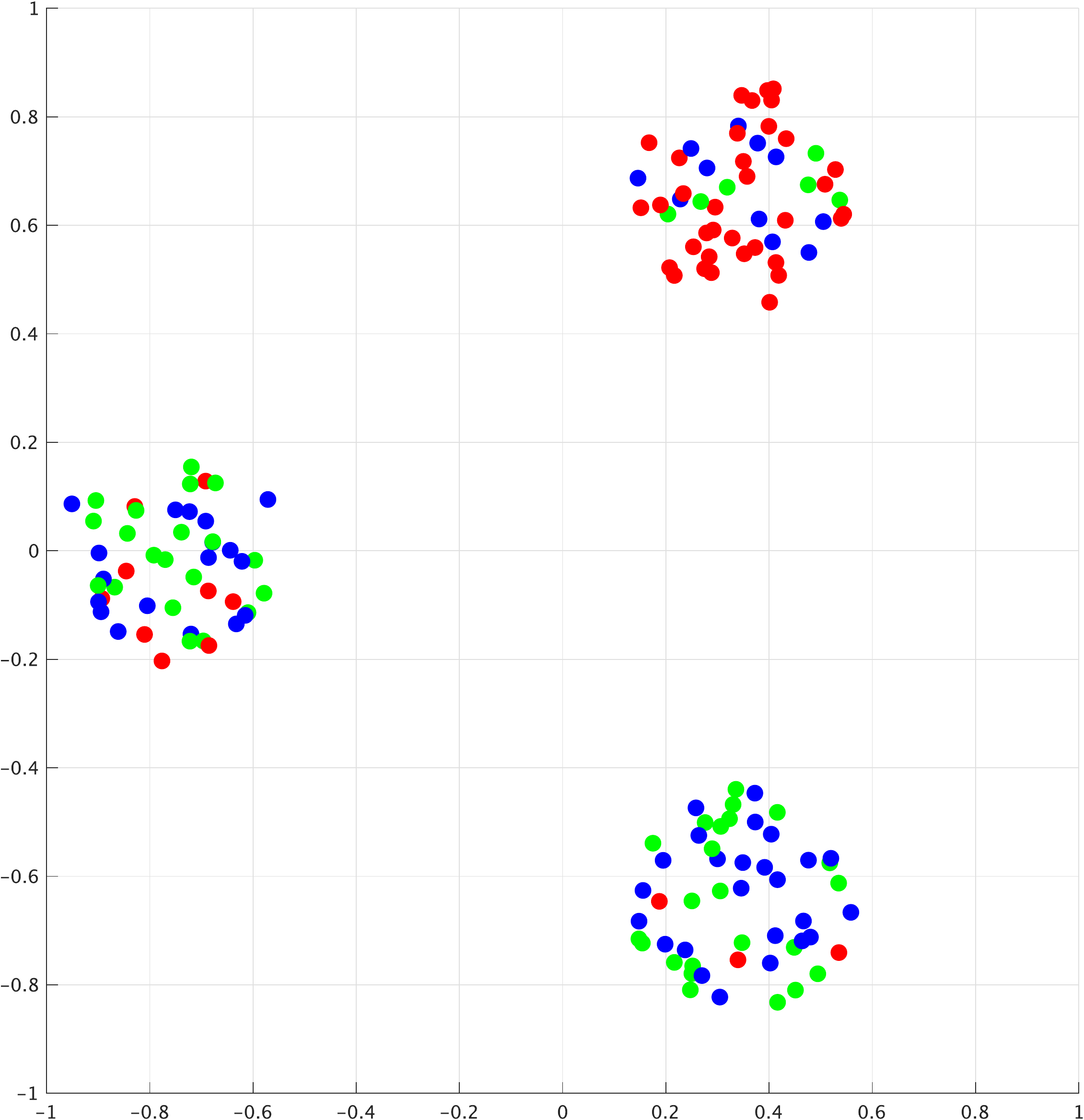}
        }
        \hspace{10pt}
    \subfigure[PisCES]{
        \includegraphics[clip, width=0.32\columnwidth]
        {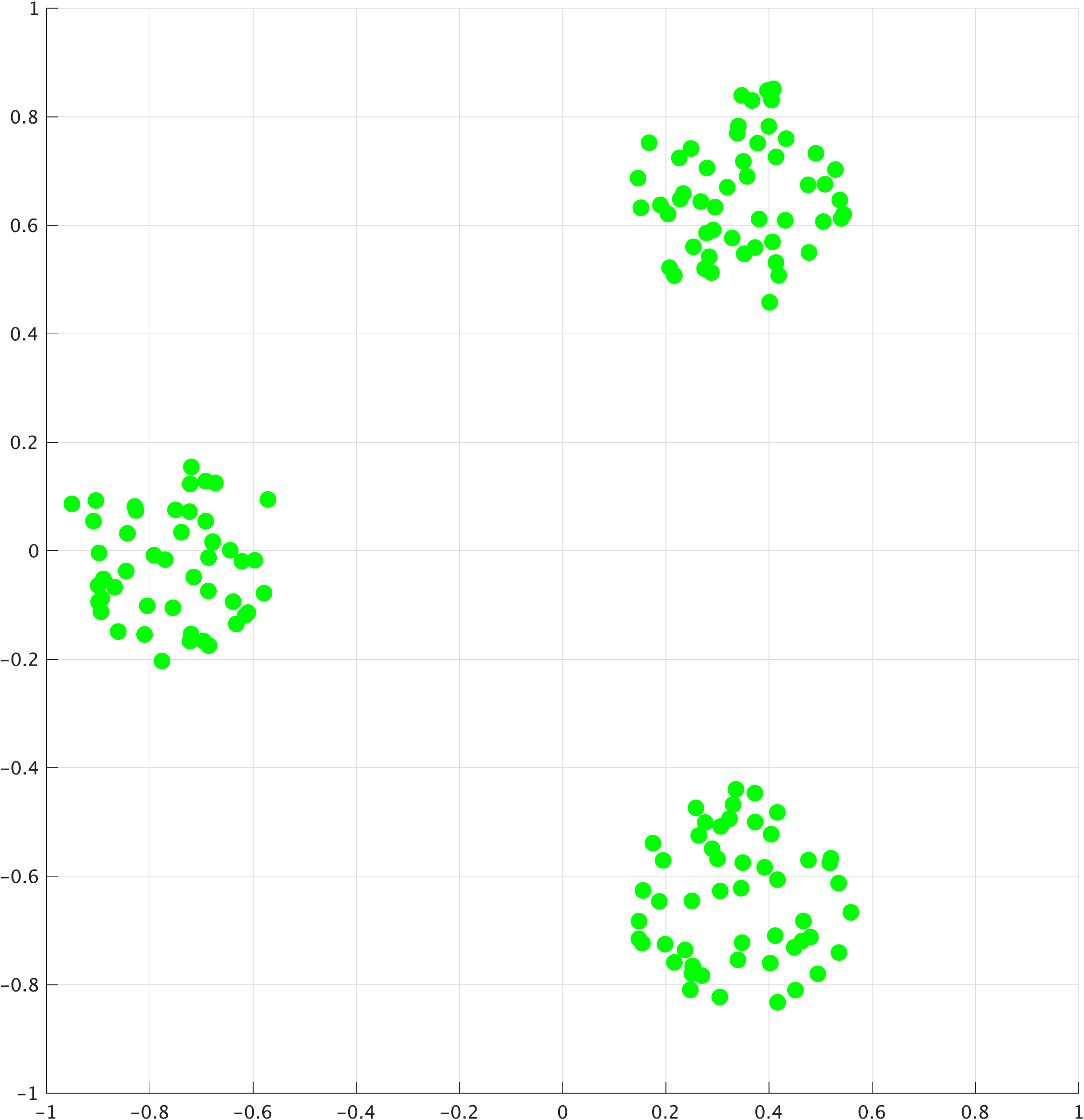}
        }
        \hspace{10pt}
    \subfigure[PisCES ($K$ is fixed)]{
        \includegraphics[clip, width=0.32\columnwidth]
        {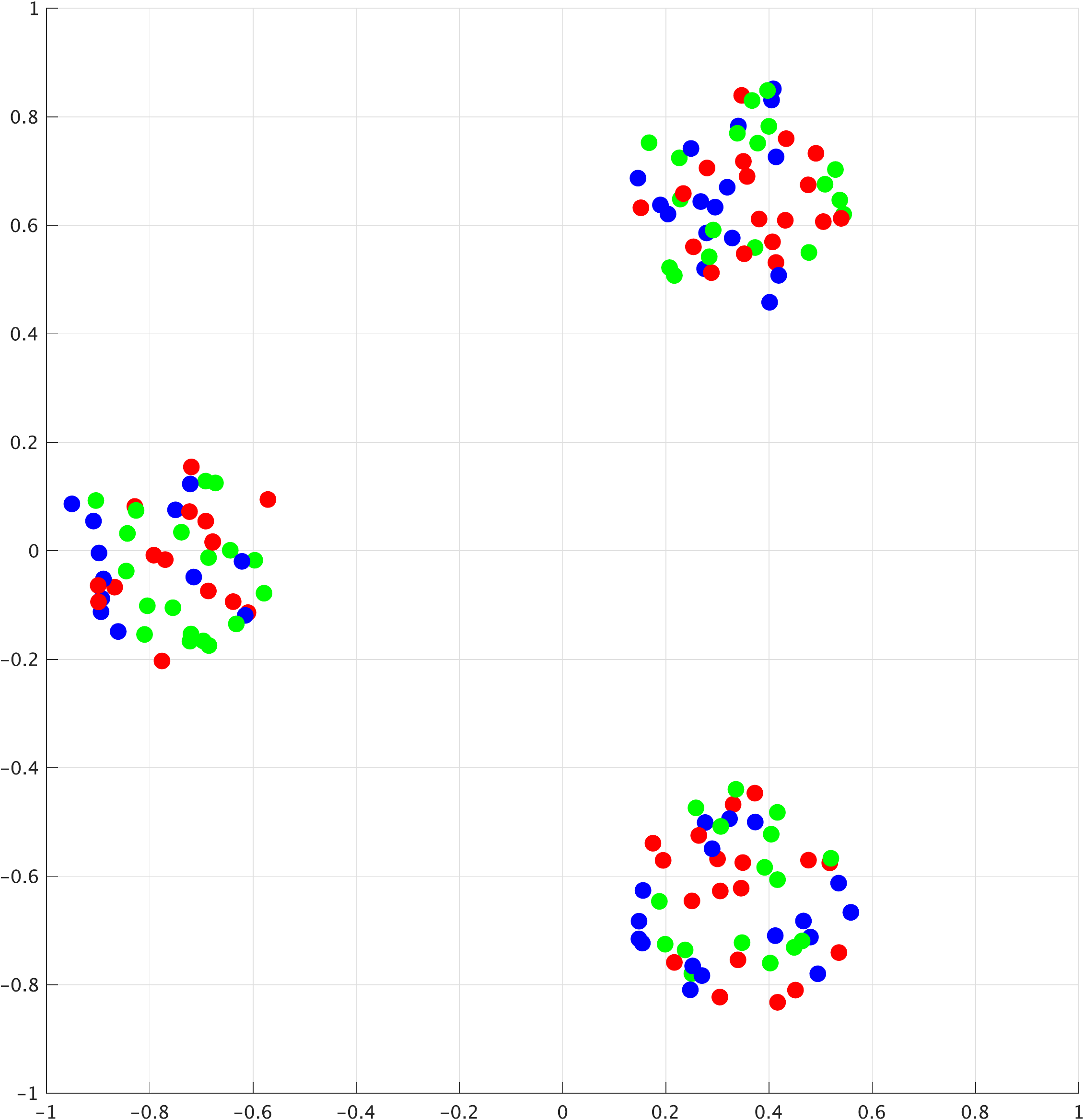}
        }
        \hspace{10pt}
    \centering
    \subfigure[Proposed method]{
        \includegraphics[clip, width=0.32\columnwidth]
        {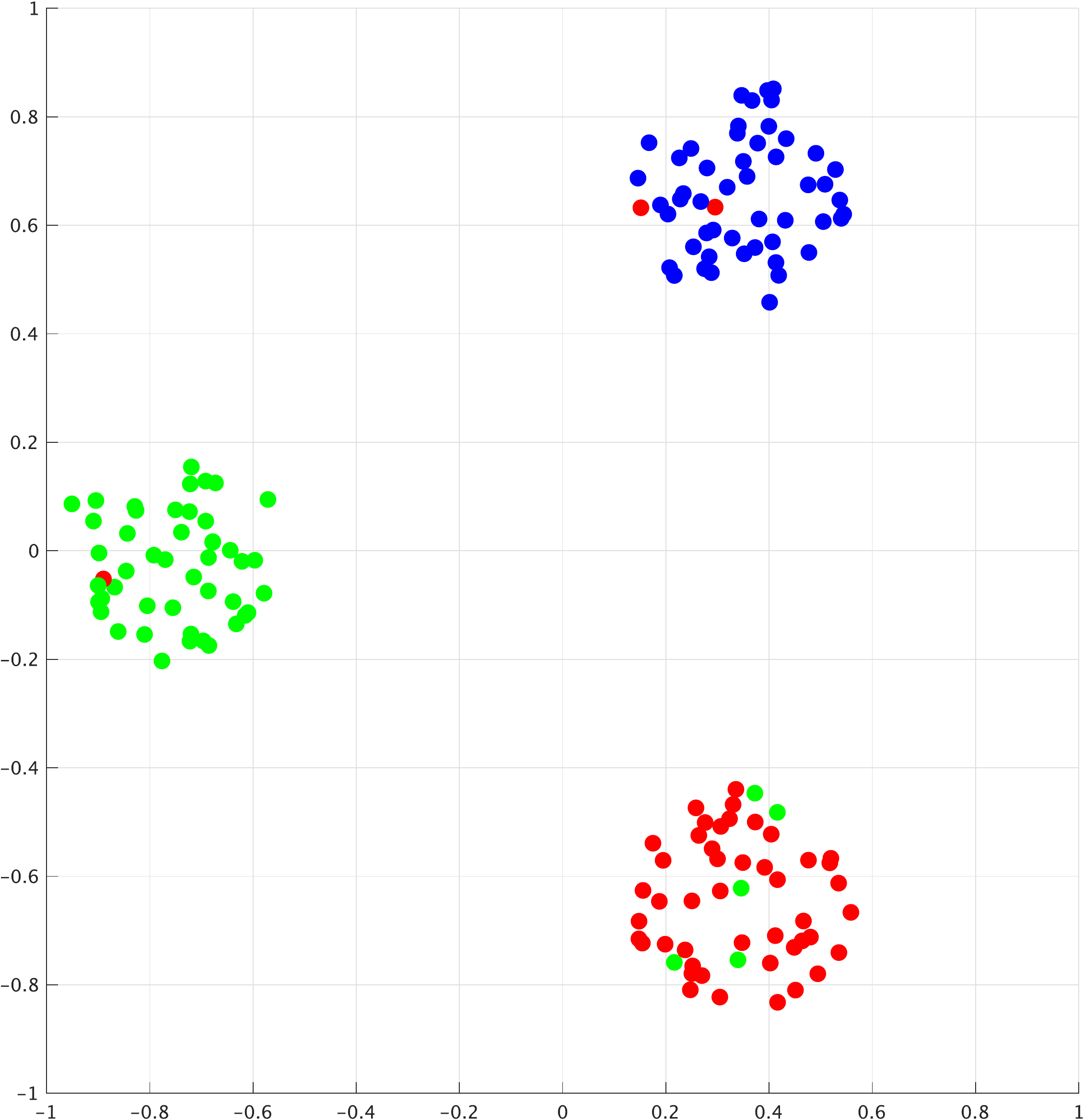}
        }
\\
    \subfigure[Graph at $t=54$]{
        \includegraphics[clip, width=0.32\columnwidth]
        {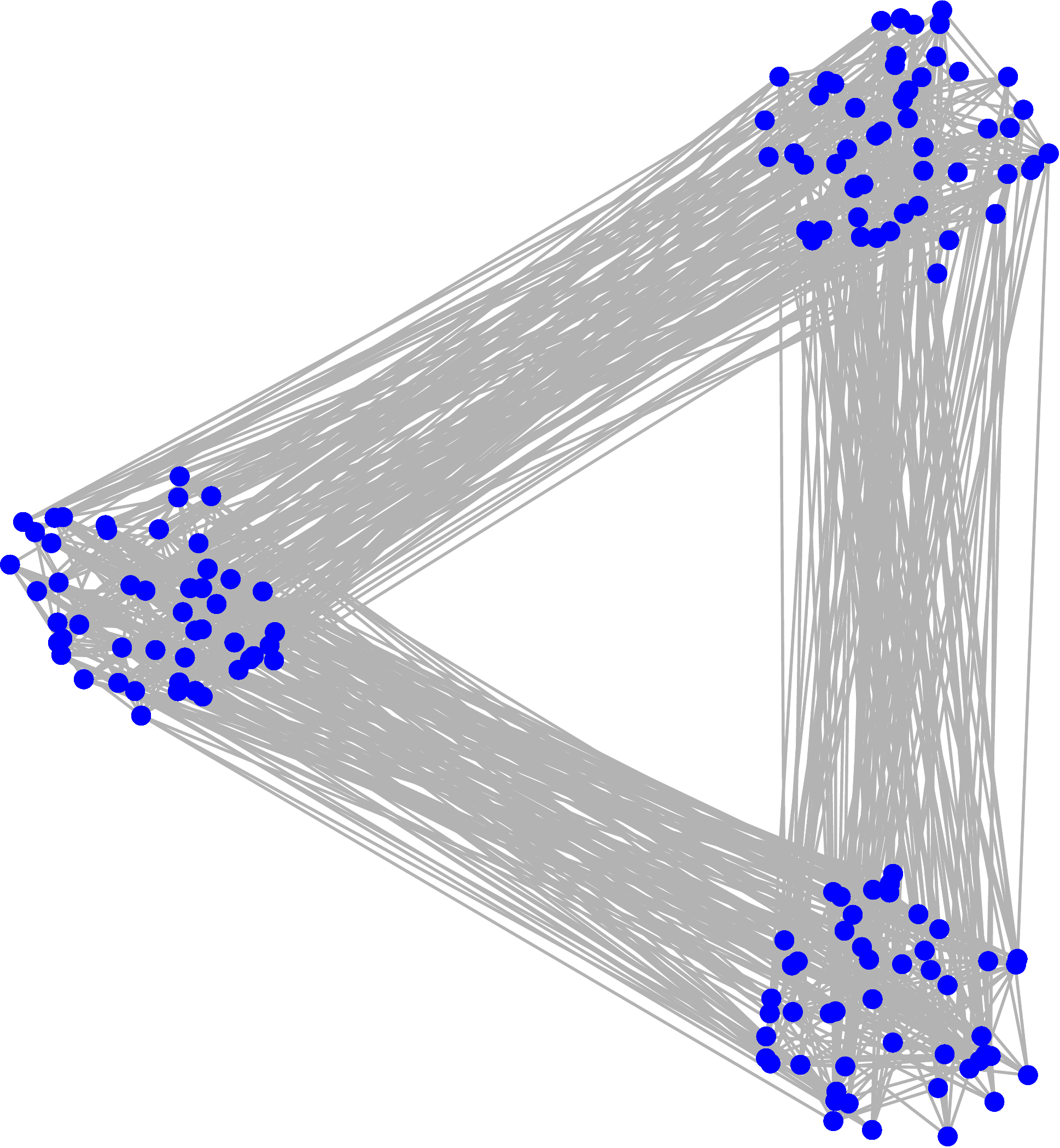}
        }
        \hspace{10pt}
    \subfigure[SC for static graph]{
        \includegraphics[clip, width=0.32\columnwidth]
        {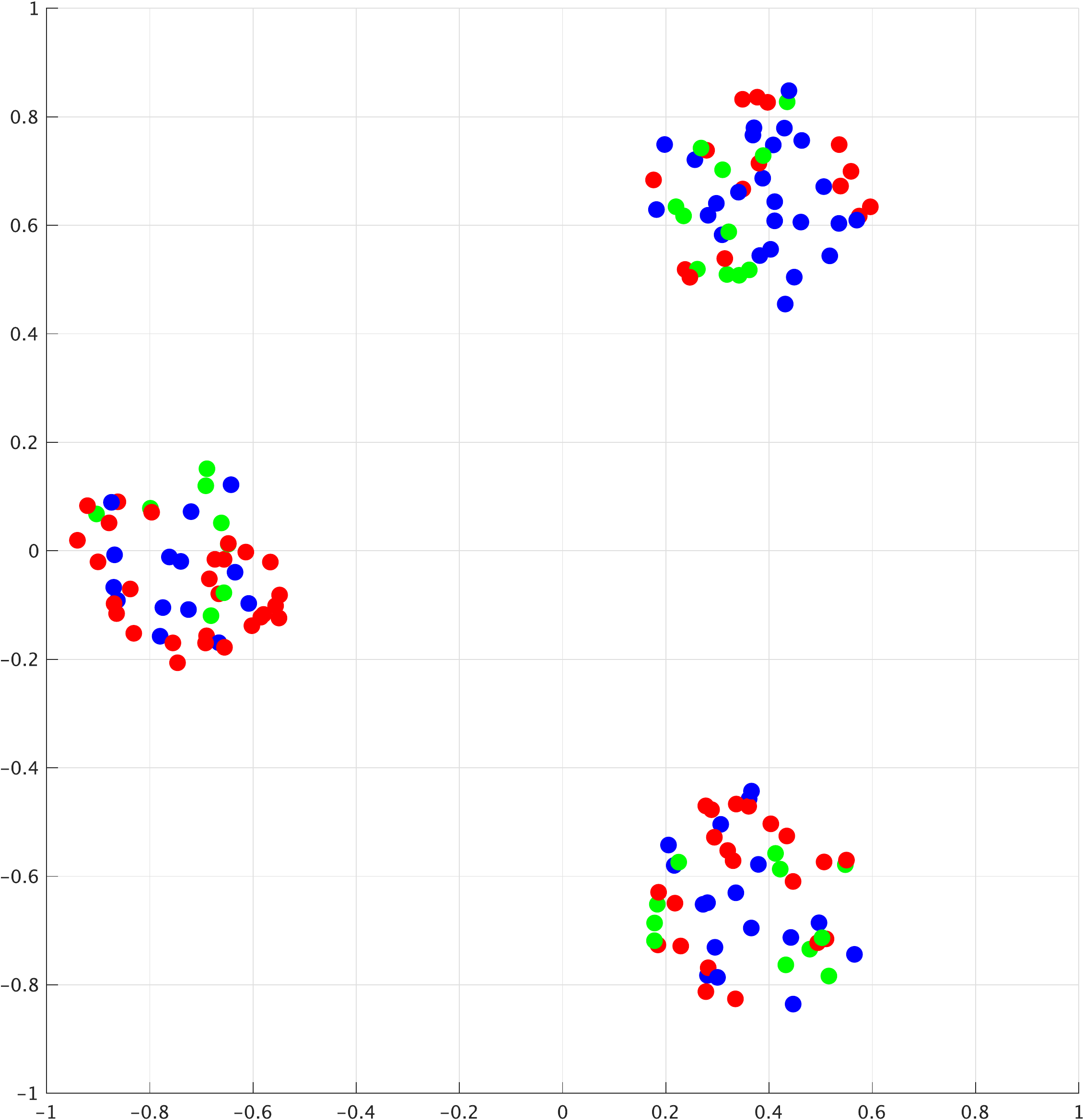}
        }
        \hspace{10pt}
    \subfigure[PisCES]{
        \includegraphics[clip, width=0.32\columnwidth]
        {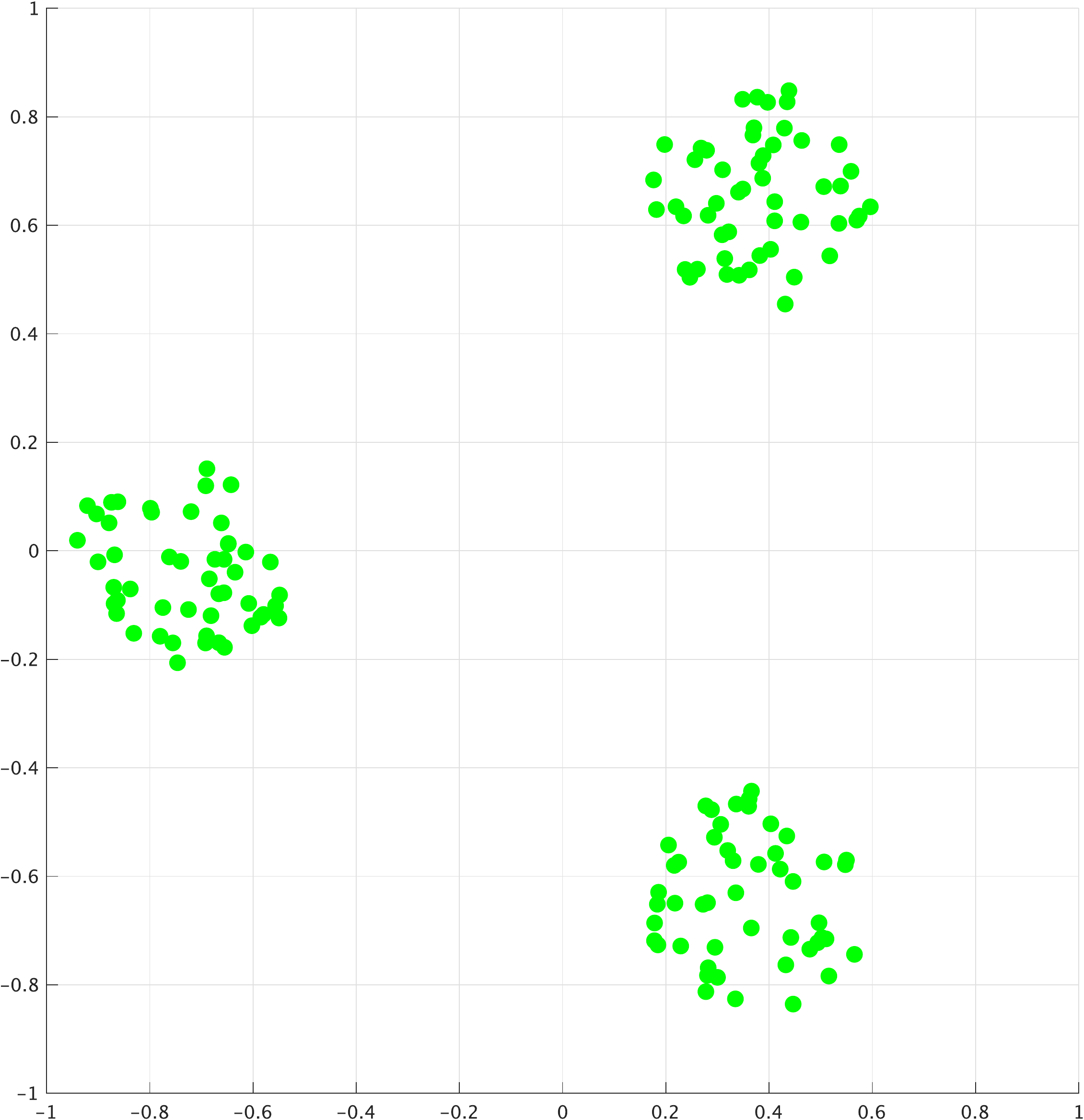}
        }
        \hspace{10pt}
    \centering
    \subfigure[PisCES ($K$ is fixed) ]{
        \includegraphics[clip, width=0.32\columnwidth]
        {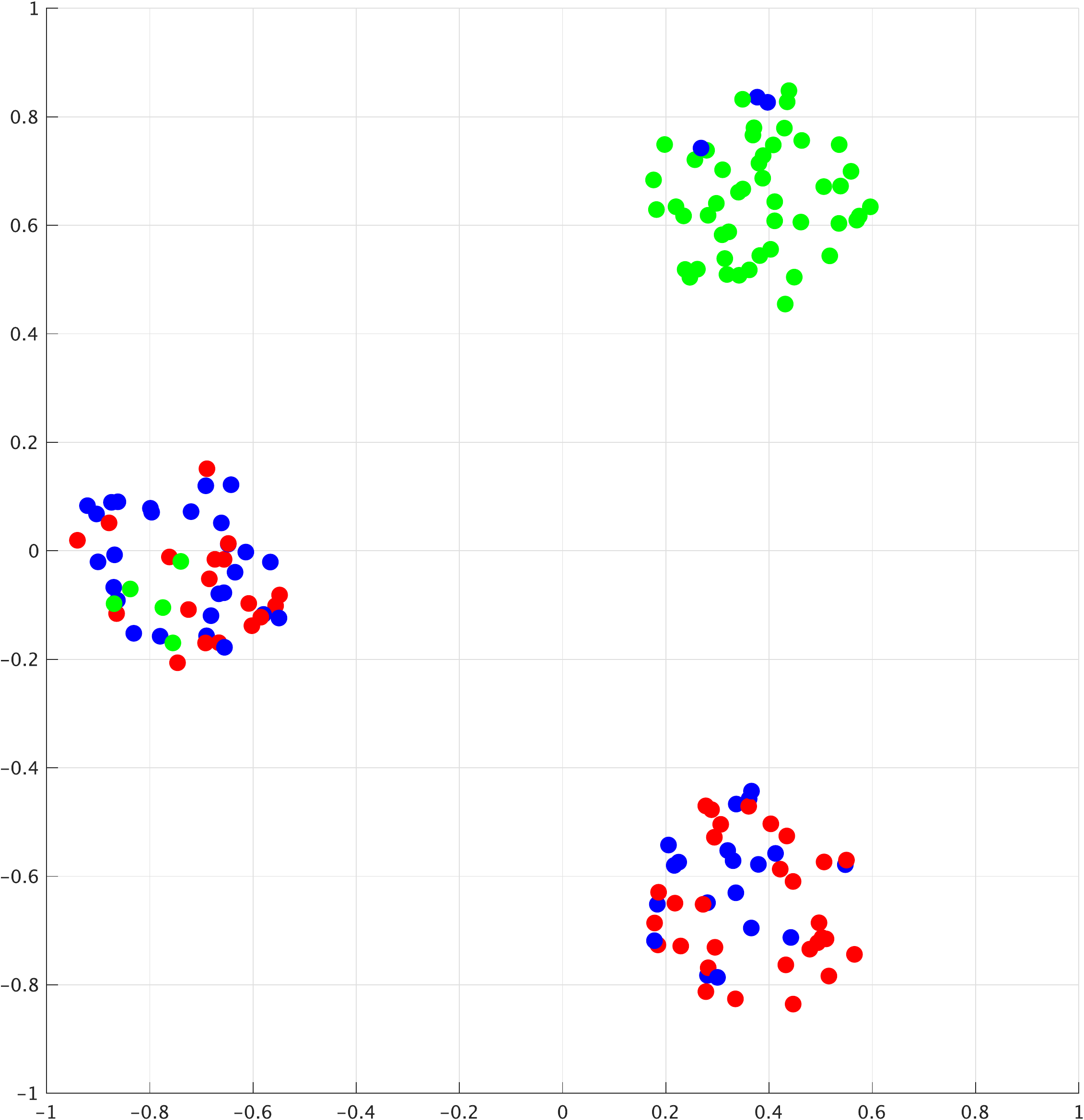}
        }
        \hspace{10pt}
    \subfigure[Proposed method]{
        \includegraphics[clip, width=0.32\columnwidth]
        {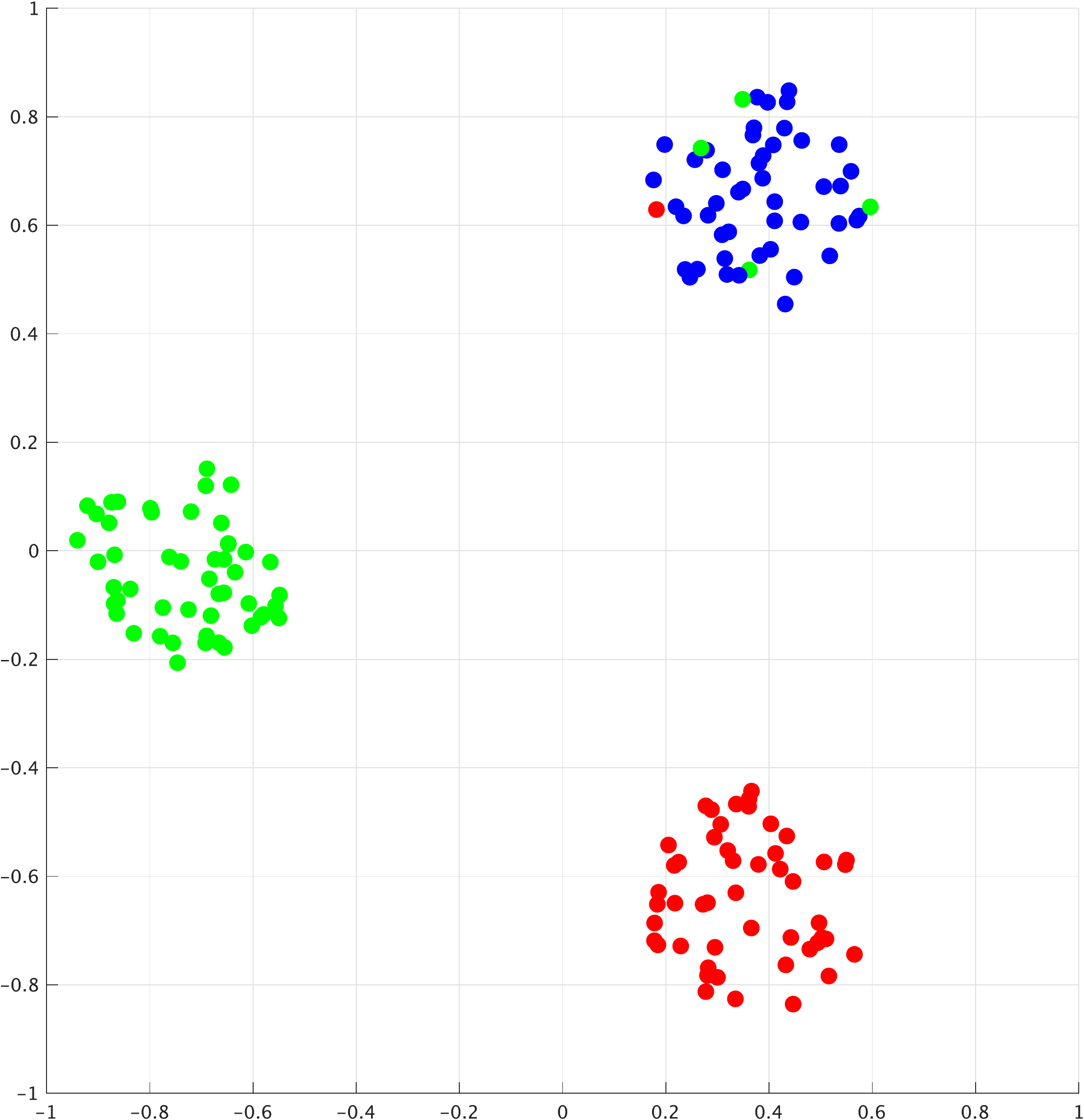}
        }
    
    \caption{Experimental results of the sparse synthetic data. (a)-(e): $t=1$. (f)-(j): $t=54$.}
    \label{fig:sy2}
\end{figure*}

\begin{figure}[t]
\centering \relax
\includegraphics[width=80mm]{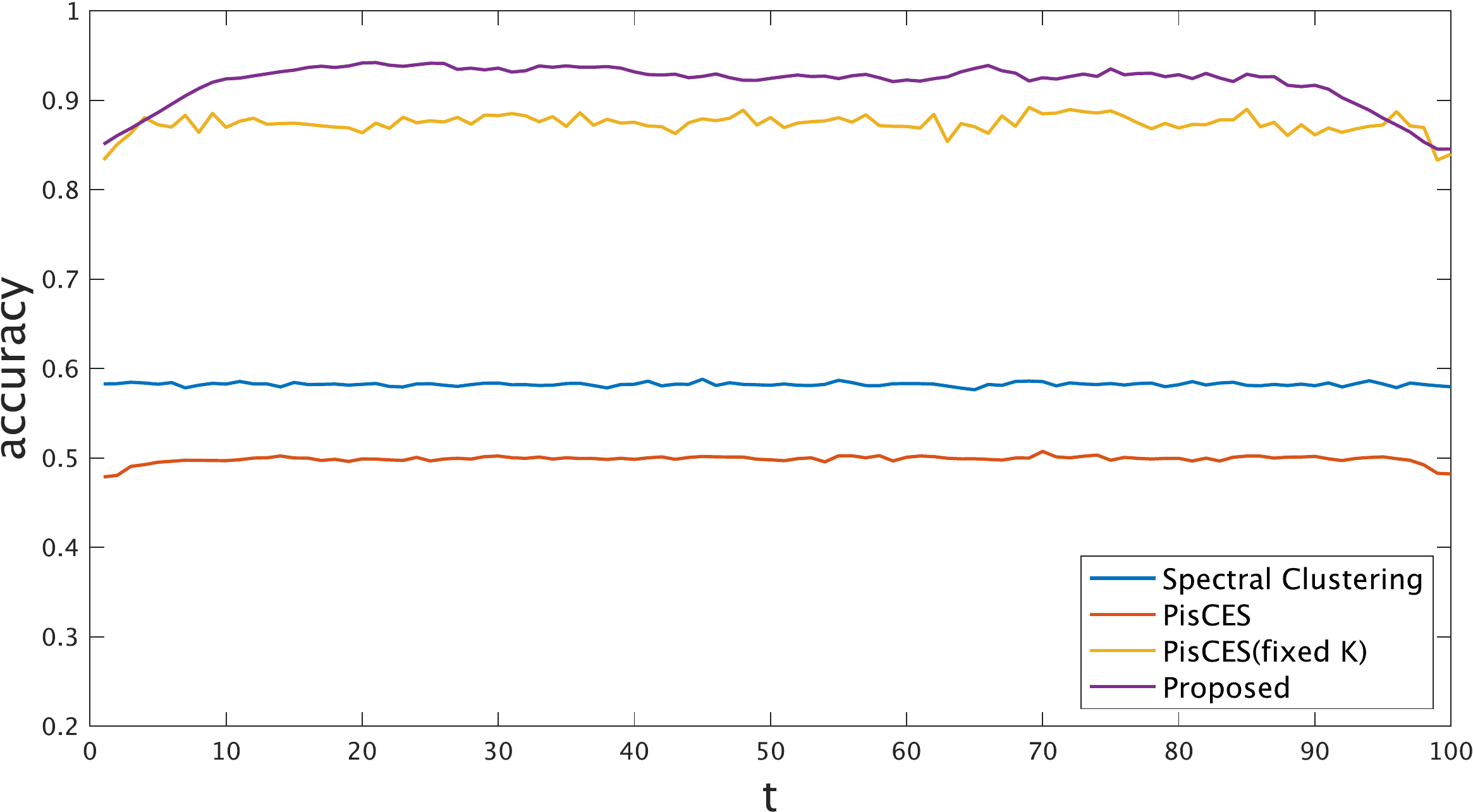}
\caption{Clustering accuracy for sparse TV graphs. Averages for 100 trials are shown.}
\label{fig:acc}
\end{figure}

\begin{figure}[t]
\centering \relax
\includegraphics[width=80mm]{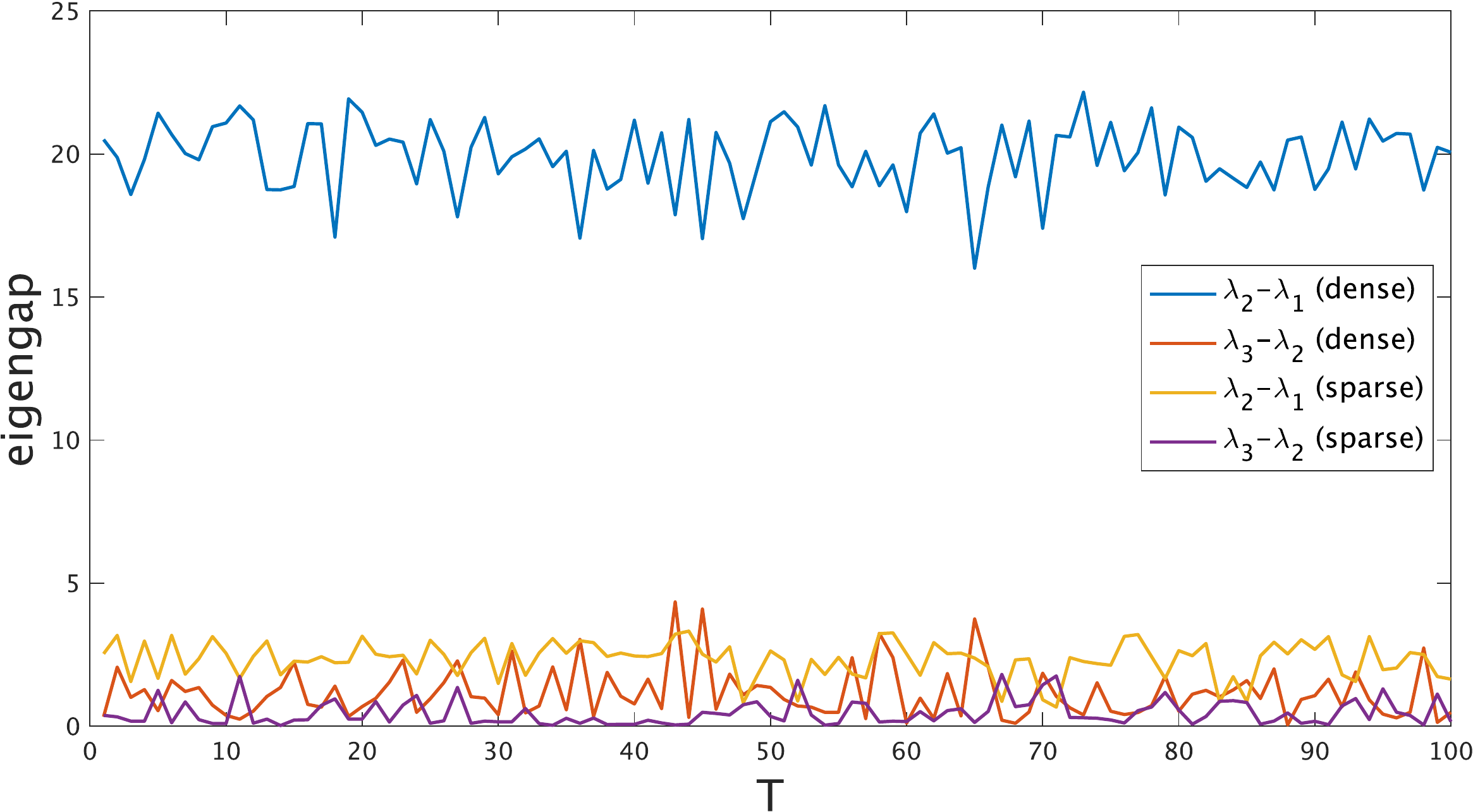}
\caption{Examples of eigengap of graph in Fig. \ref{fig:sy1} and \ref{fig:sy2}. $\lambda_i$ is the $i$-th smallest eigenvalue.}
\label{fig:egap1}
\end{figure}

\begin{figure*}[tbp]
    \centering
    \subfigure[Graph at $t = 1$]{%
        \includegraphics[clip, width=0.35\columnwidth]%
    {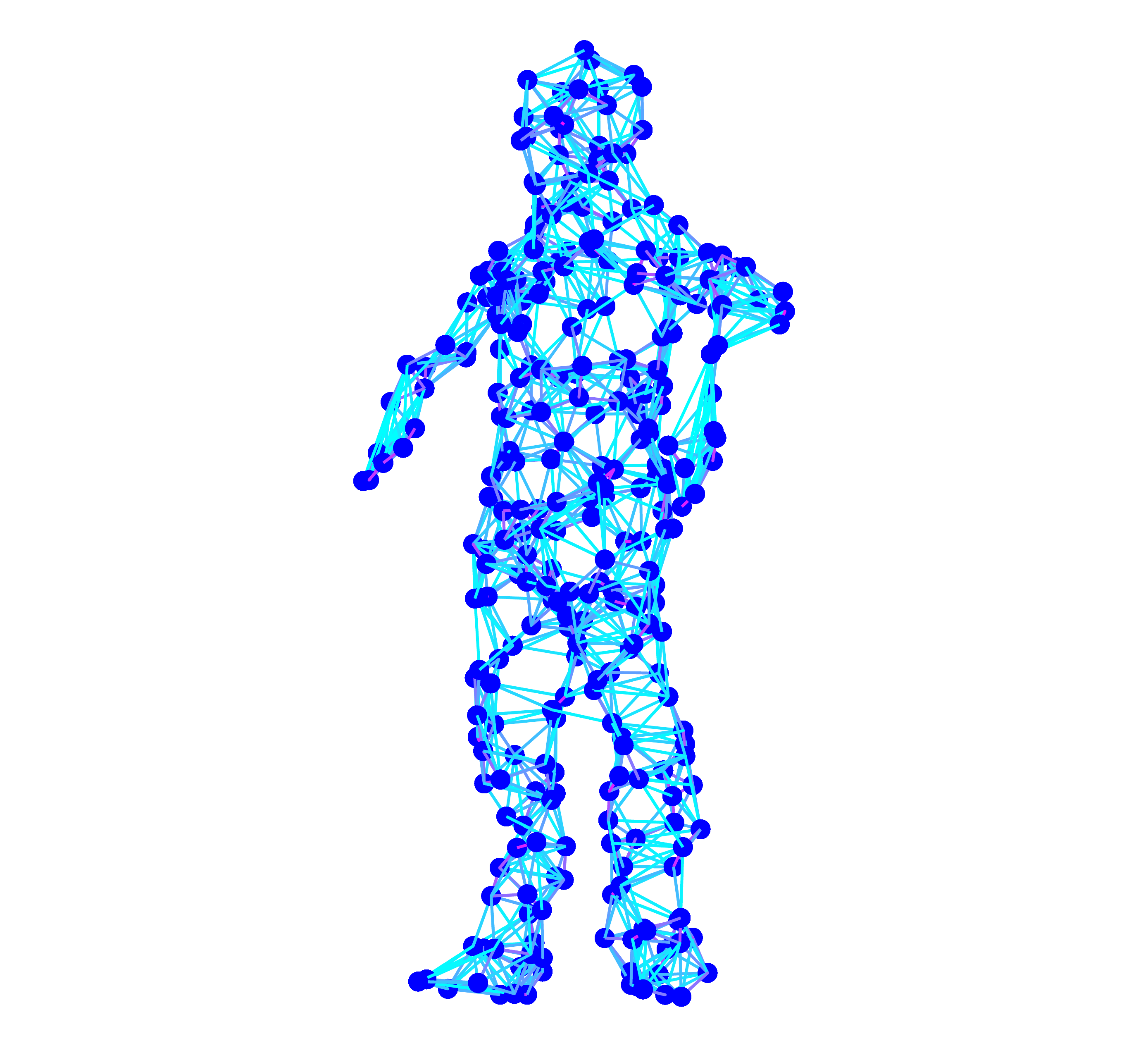}%
        }%
        \hspace{10pt}
    \subfigure[SC for static graph]{%
        \includegraphics[clip, width=0.35\columnwidth]%
        {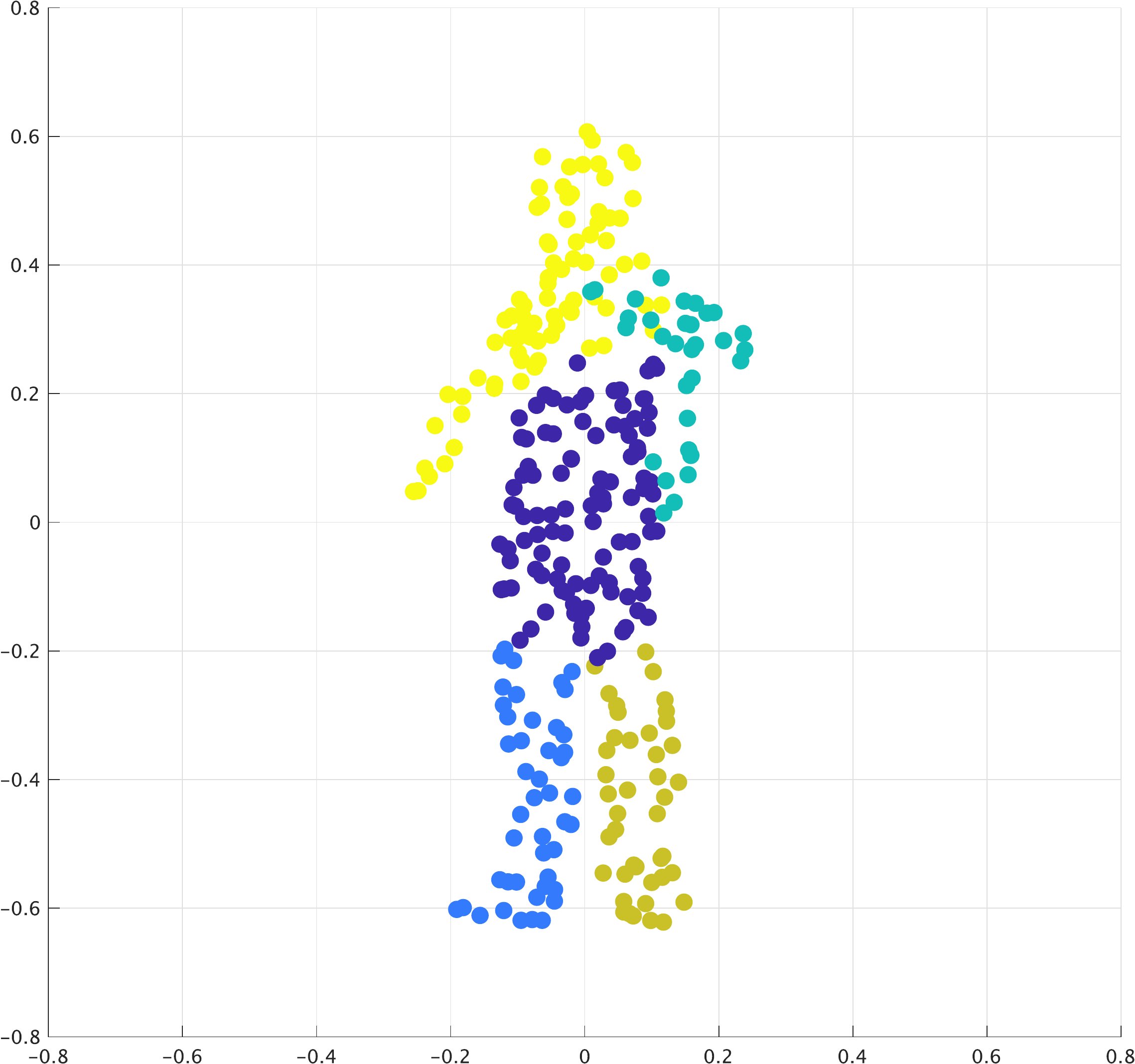}%
        }%
        \hspace{10pt}
            \subfigure[PisCES]{%
        \includegraphics[clip, width=0.35\columnwidth]%
        {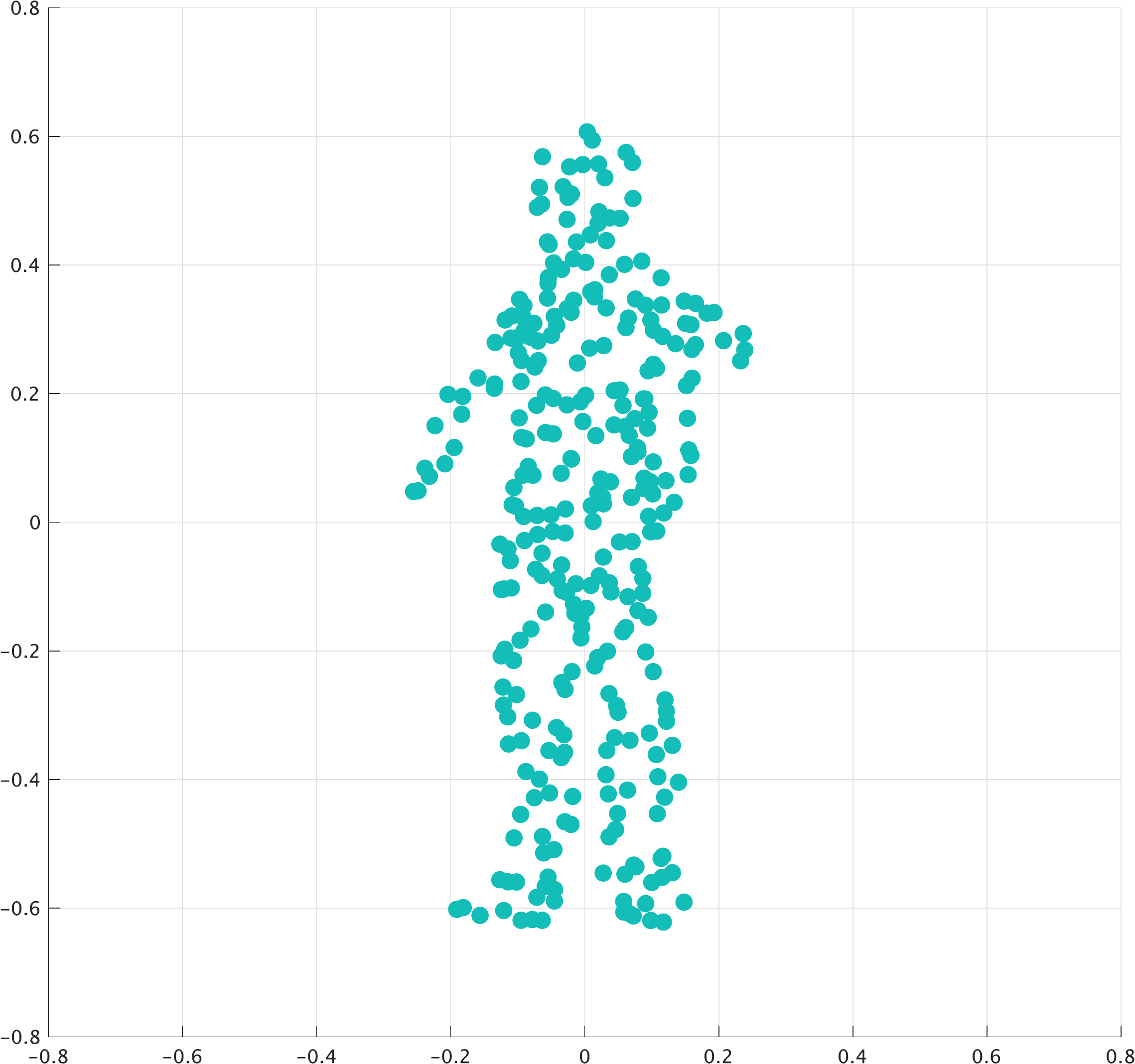}%
        }%
        \hspace{10pt}
    \subfigure[PisCES ($K$ is fixed)]{%
        \includegraphics[clip, width=0.35\columnwidth]%
        {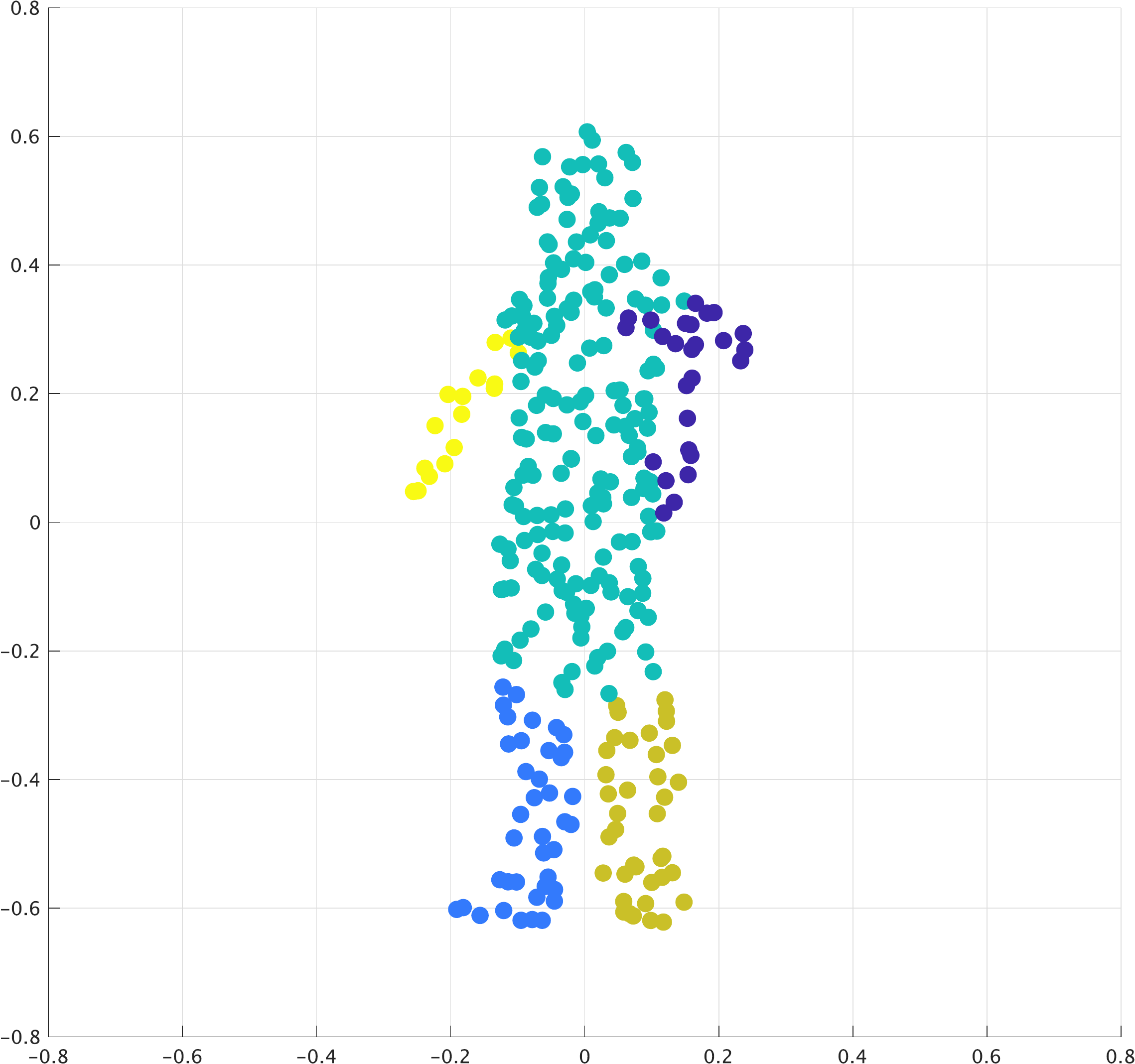}%
        }%
        \hspace{10pt}
    \subfigure[Proposed method]{%
        \includegraphics[clip, width=0.35\columnwidth]%
        {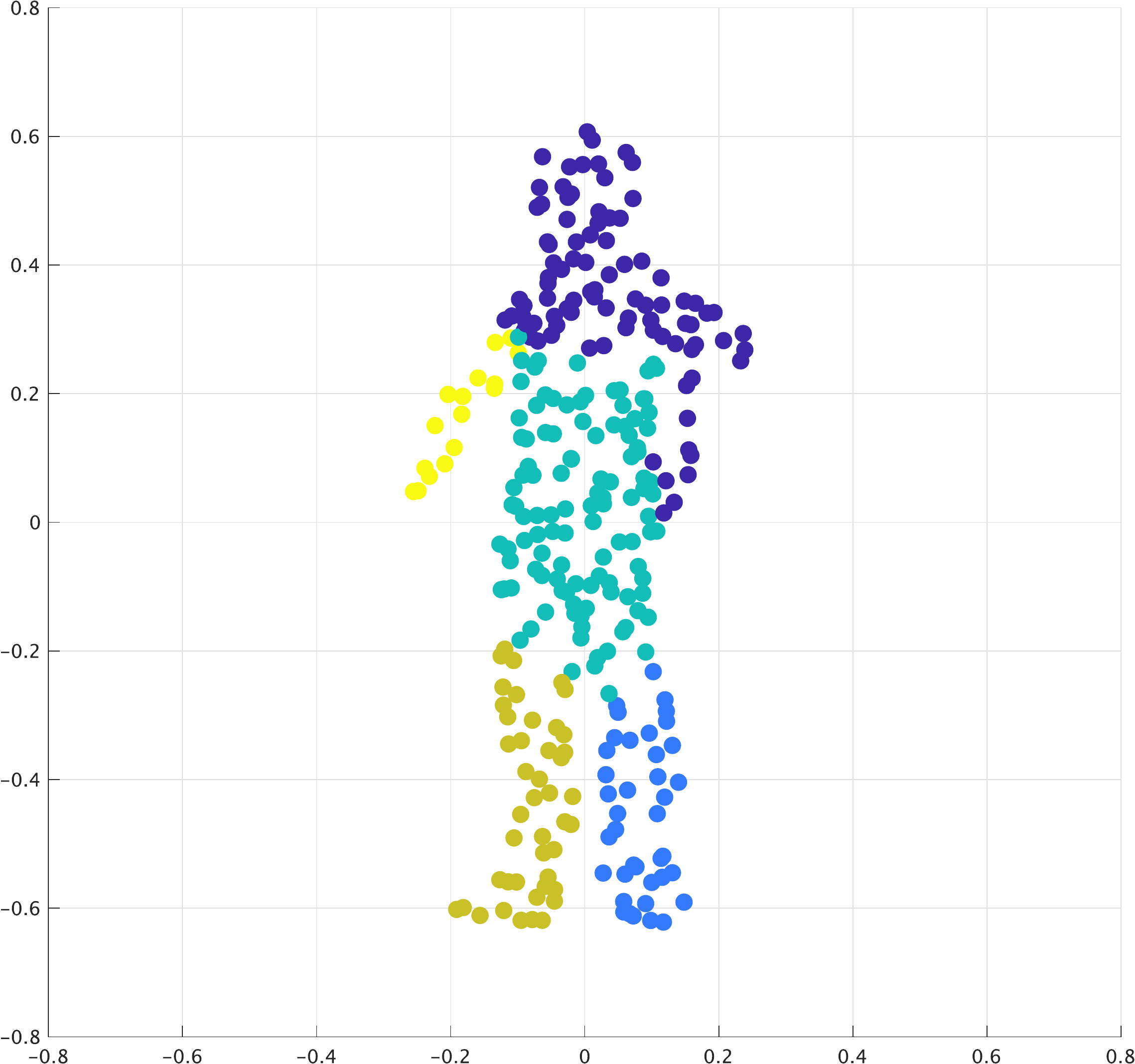}%
        }%
\\
    \subfigure[Graph at $t = 47$]{%
        \includegraphics[clip, width=0.35\columnwidth]%
        {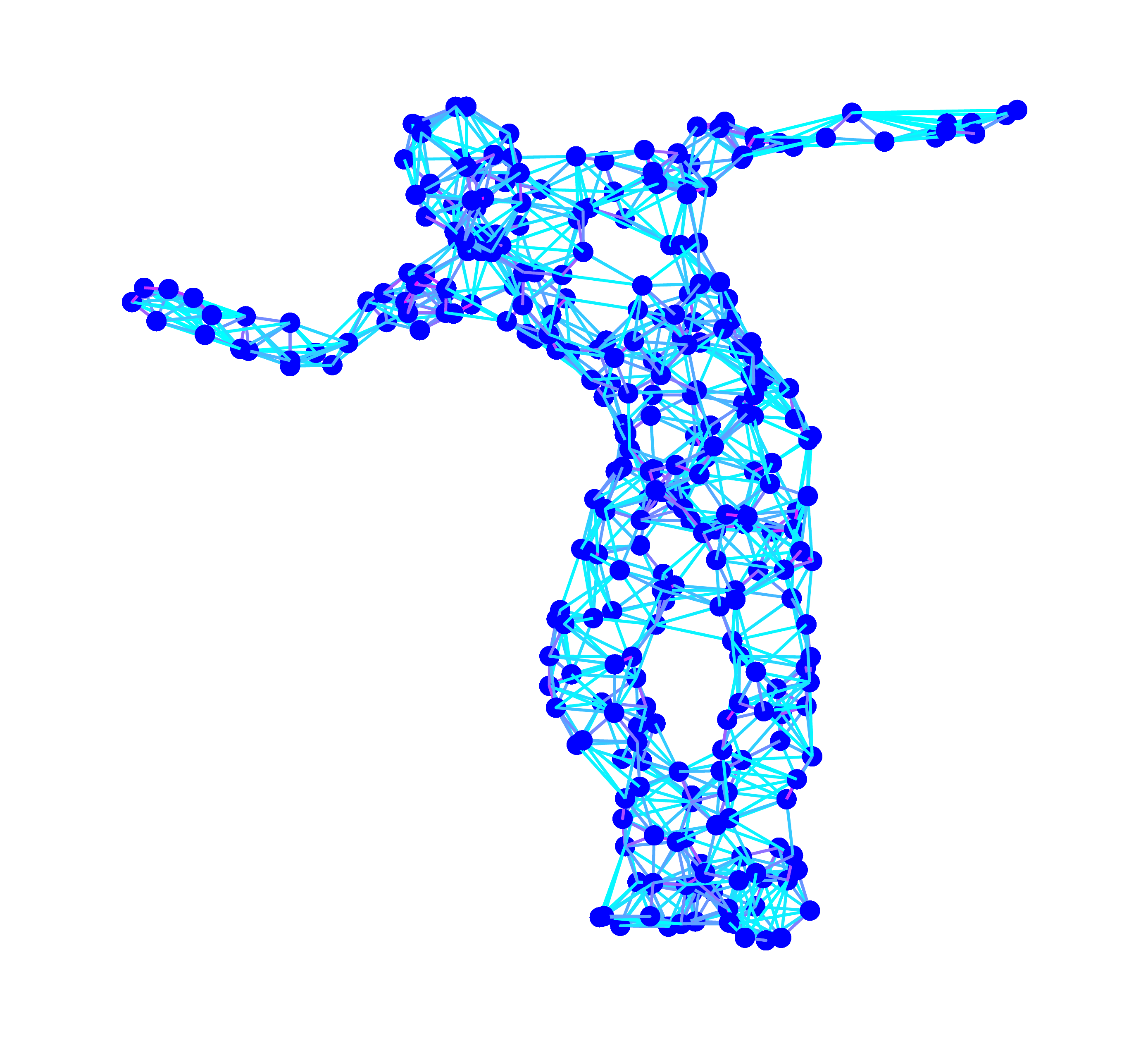}%
        }%
        \hspace{10pt}
    \subfigure[SC for static graph]{%
        \includegraphics[clip, width=0.35\columnwidth]%
        {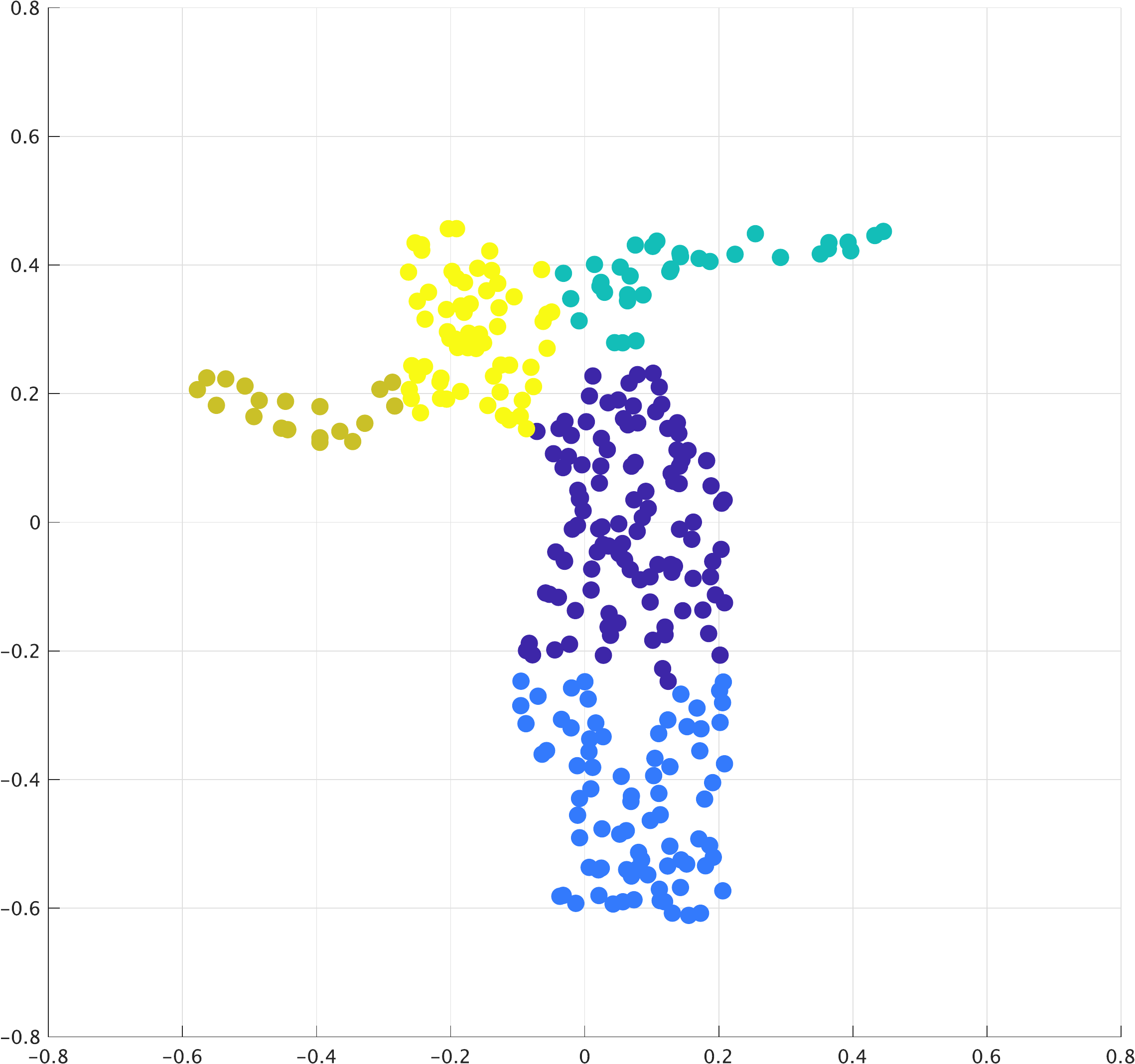}%
        }%
        \hspace{10pt}
            \subfigure[PisCES]{%
        \includegraphics[clip, width=0.35\columnwidth]%
        {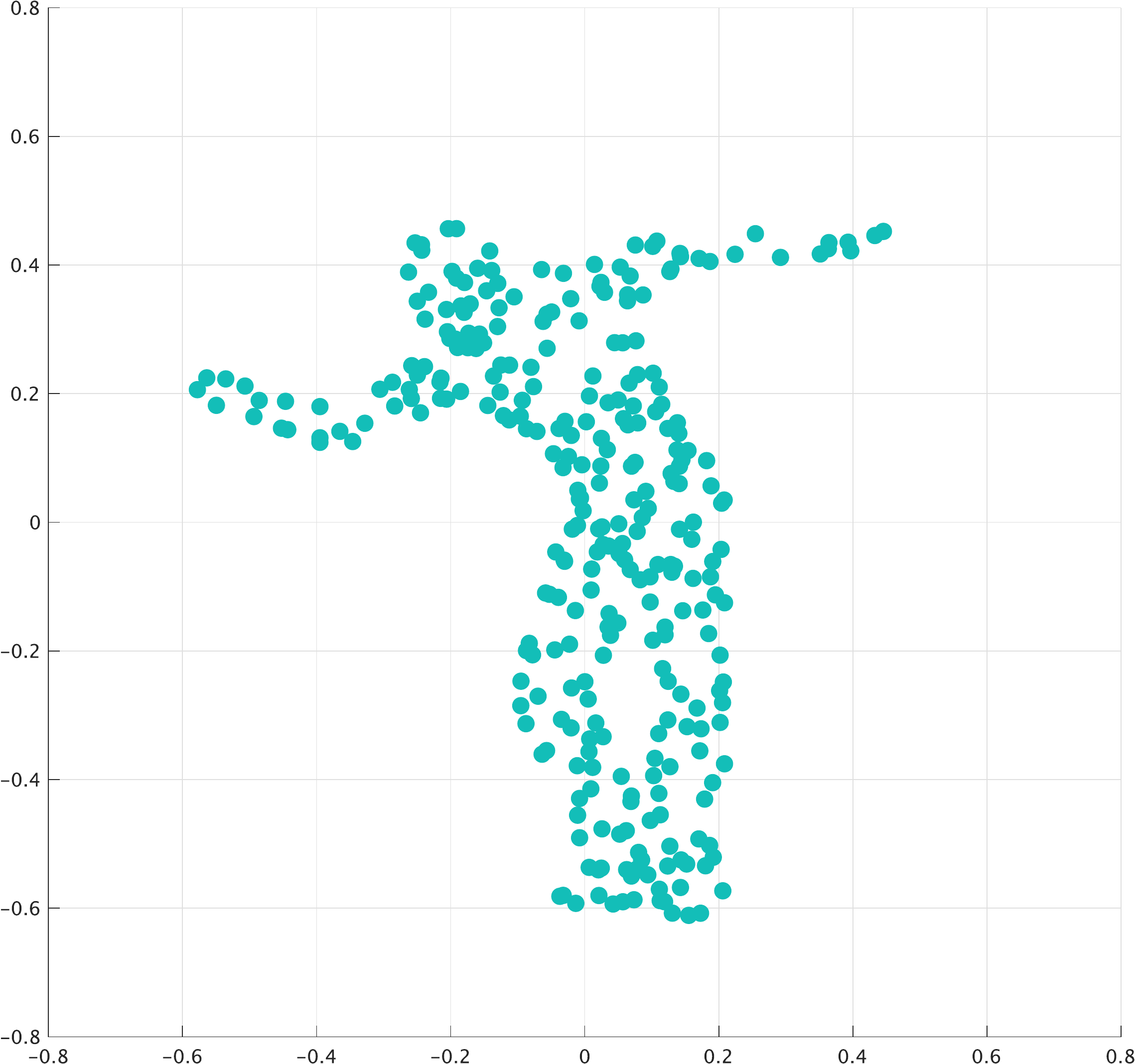}%
        }%
        \hspace{10pt}
    \subfigure[PisCES ($K$ is fixed)]{%
        \includegraphics[clip, width=0.35\columnwidth]%
        {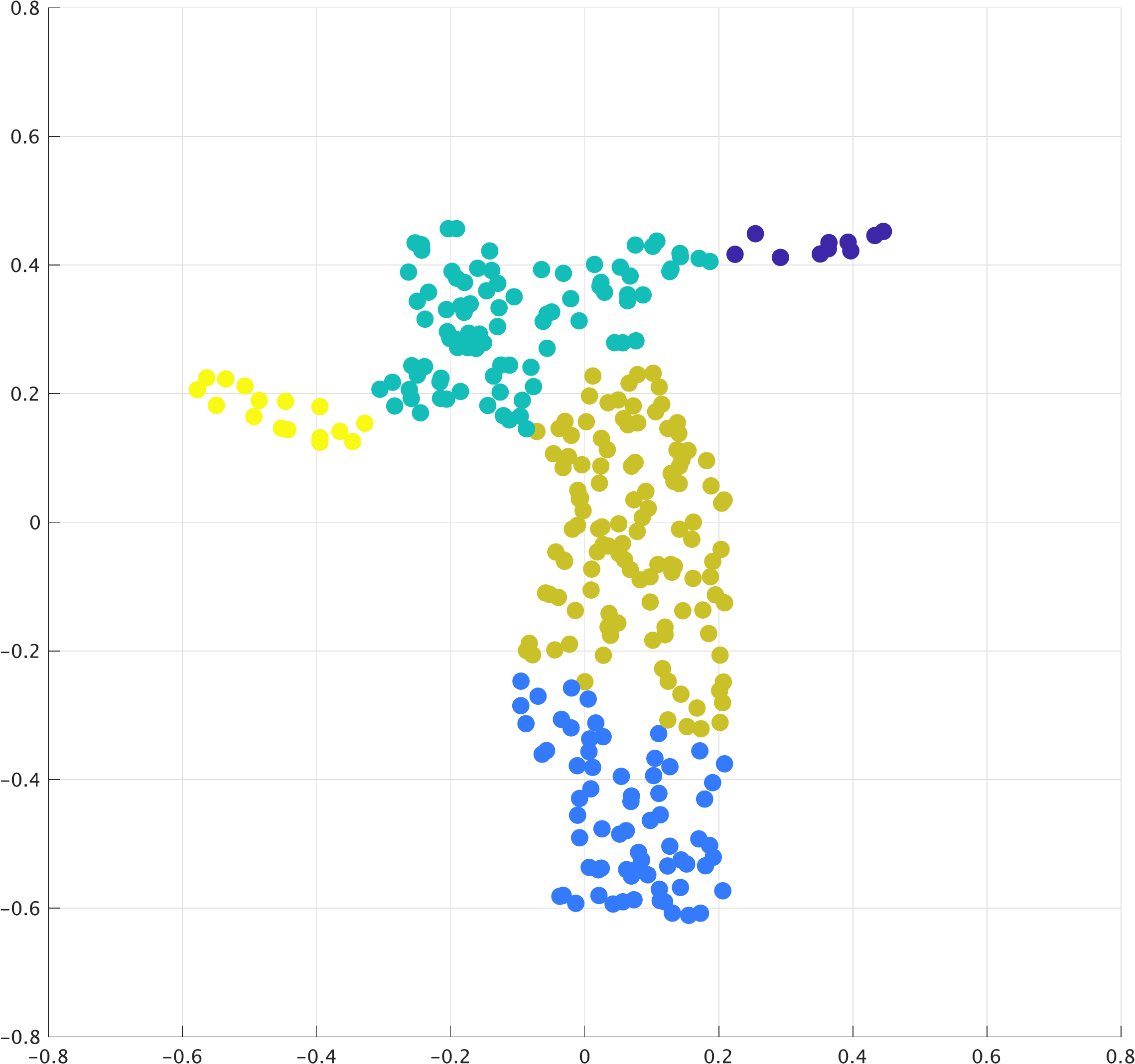}%
        }%
        \hspace{10pt}
    \subfigure[Proposed method]{%
        \includegraphics[clip, width=0.35\columnwidth]%
        {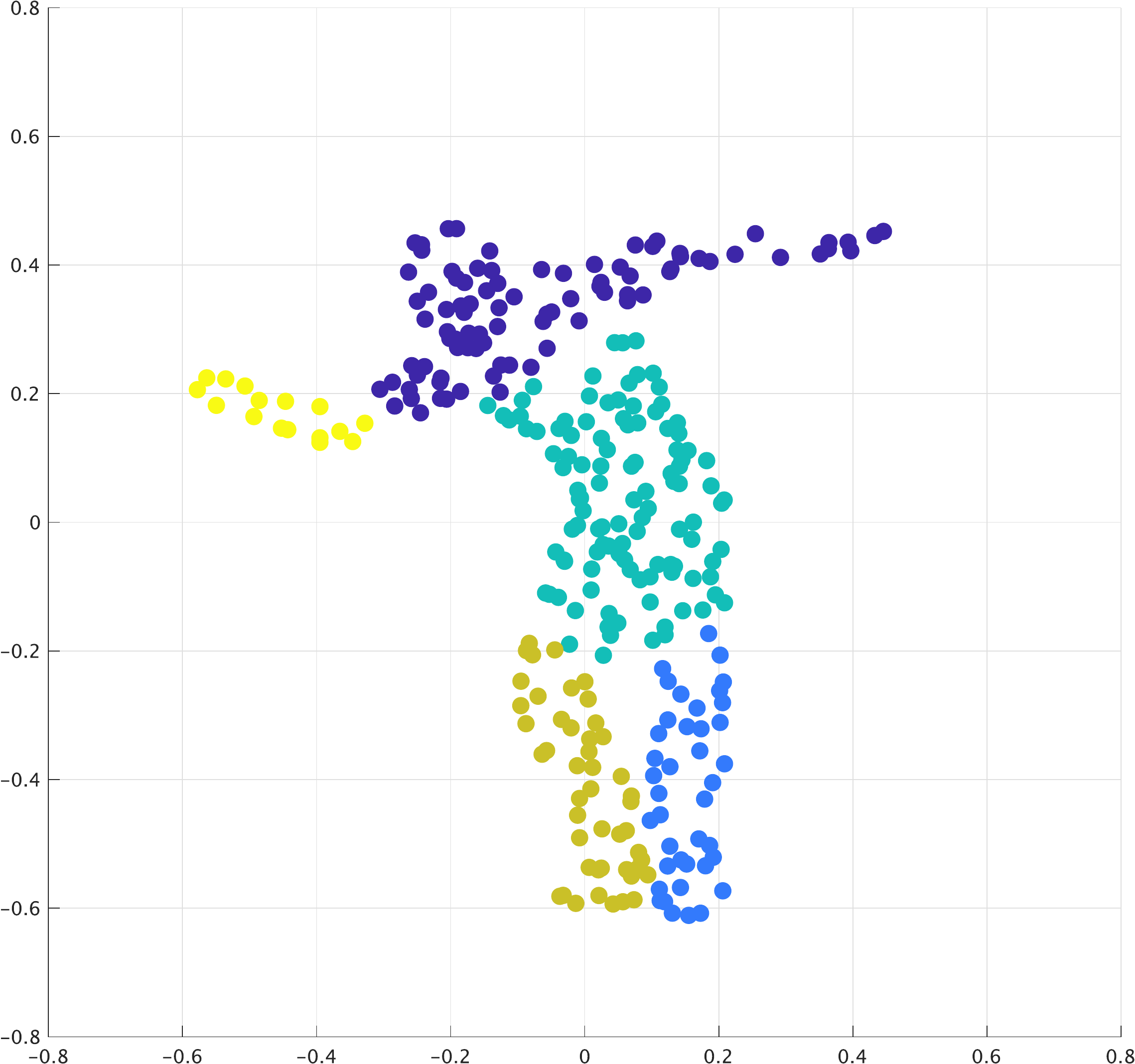}%
        }%
    \caption{Experimental results for dynamic point clouds (cartwheel). Node colors indicate the cluster labels. (a)-(e): $t=1$. (f)-(j): $t=47$.}
    \label{fig:wheel}
\end{figure*}

\begin{figure*}[tbp]
    \centering
    \subfigure[Graph at $t = 1$]{%
        \includegraphics[clip, width=0.35\columnwidth]%
        {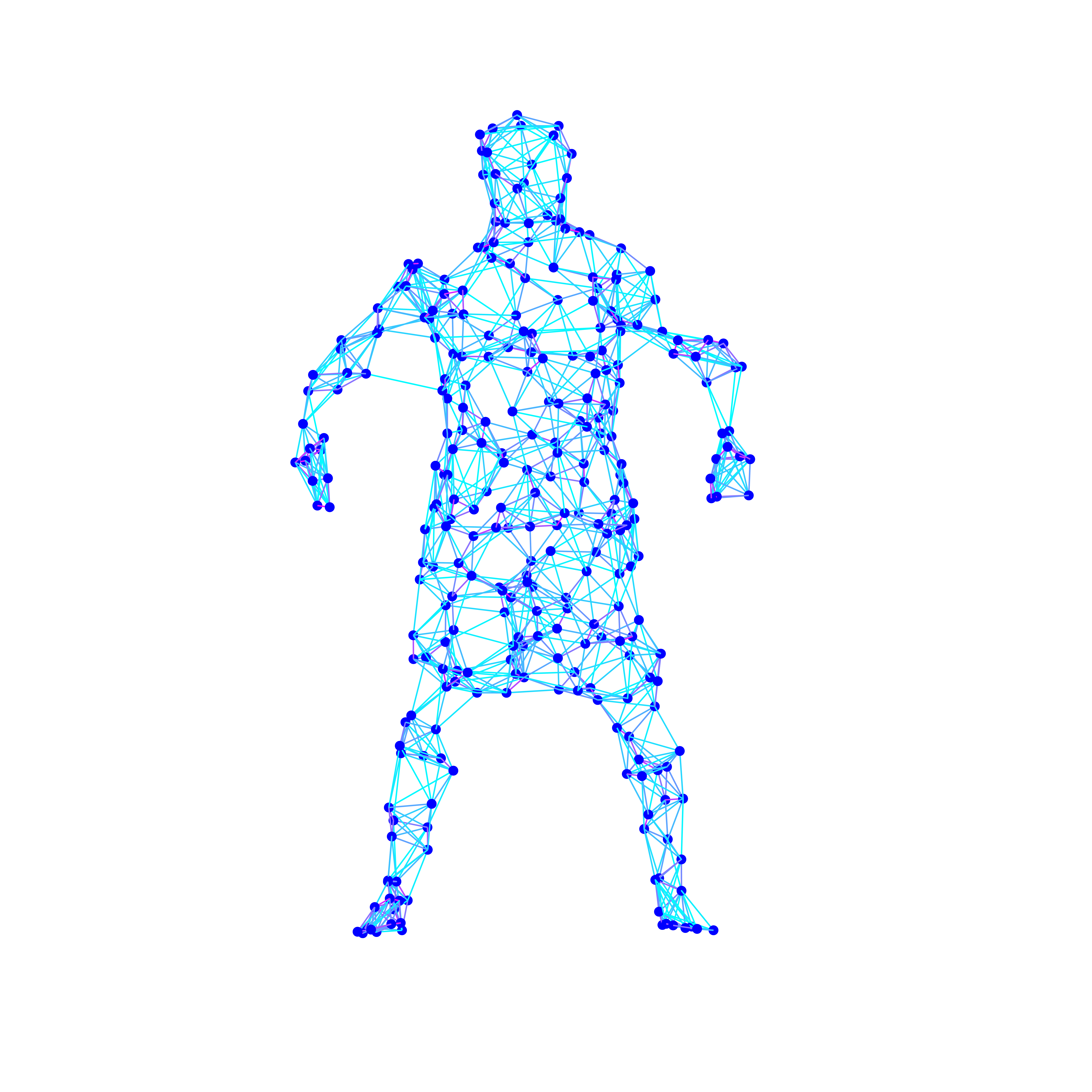}%
        }%
        \hspace{10pt}
    \subfigure[SC for static graph]{%
        \includegraphics[clip, width=0.35\columnwidth]%
        {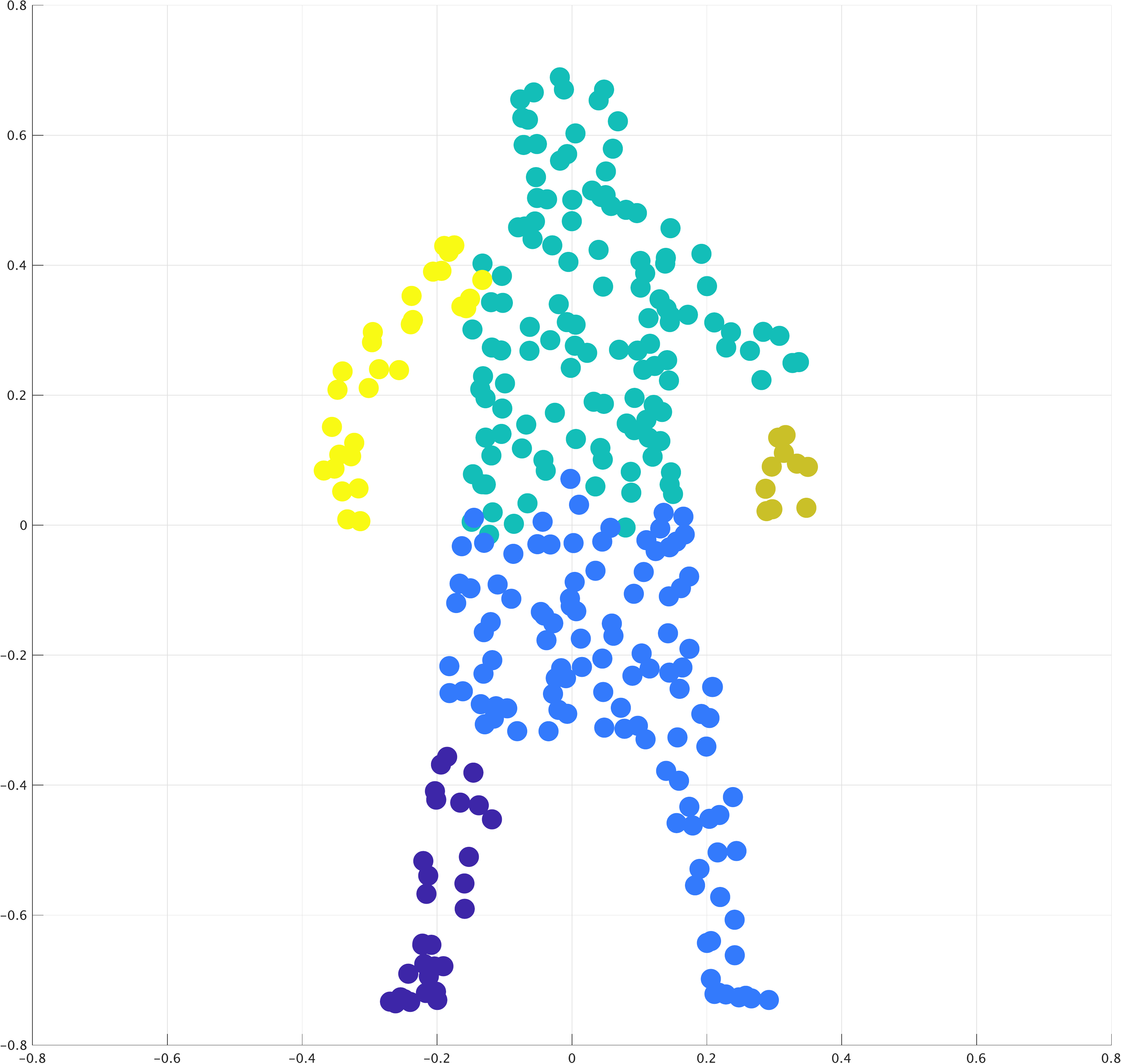}%
        }%
        \hspace{10pt}
    \subfigure[PisCES]{%
        \includegraphics[clip, width=0.35\columnwidth]%
        {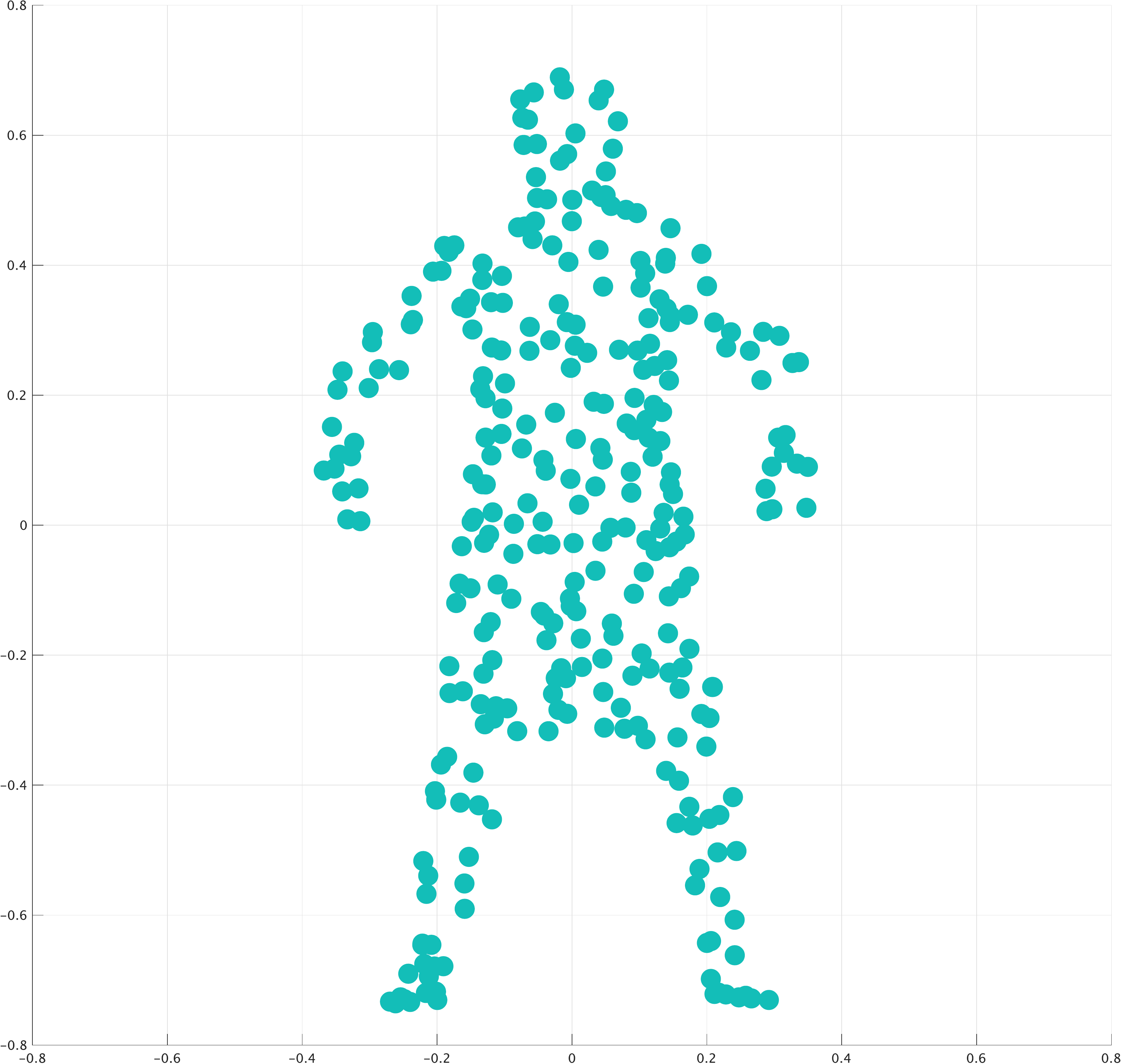}%
        }%
        \hspace{10pt}
            \subfigure[PisCES ($K$ is fixed)]{%
        \includegraphics[clip, width=0.35\columnwidth]%
        {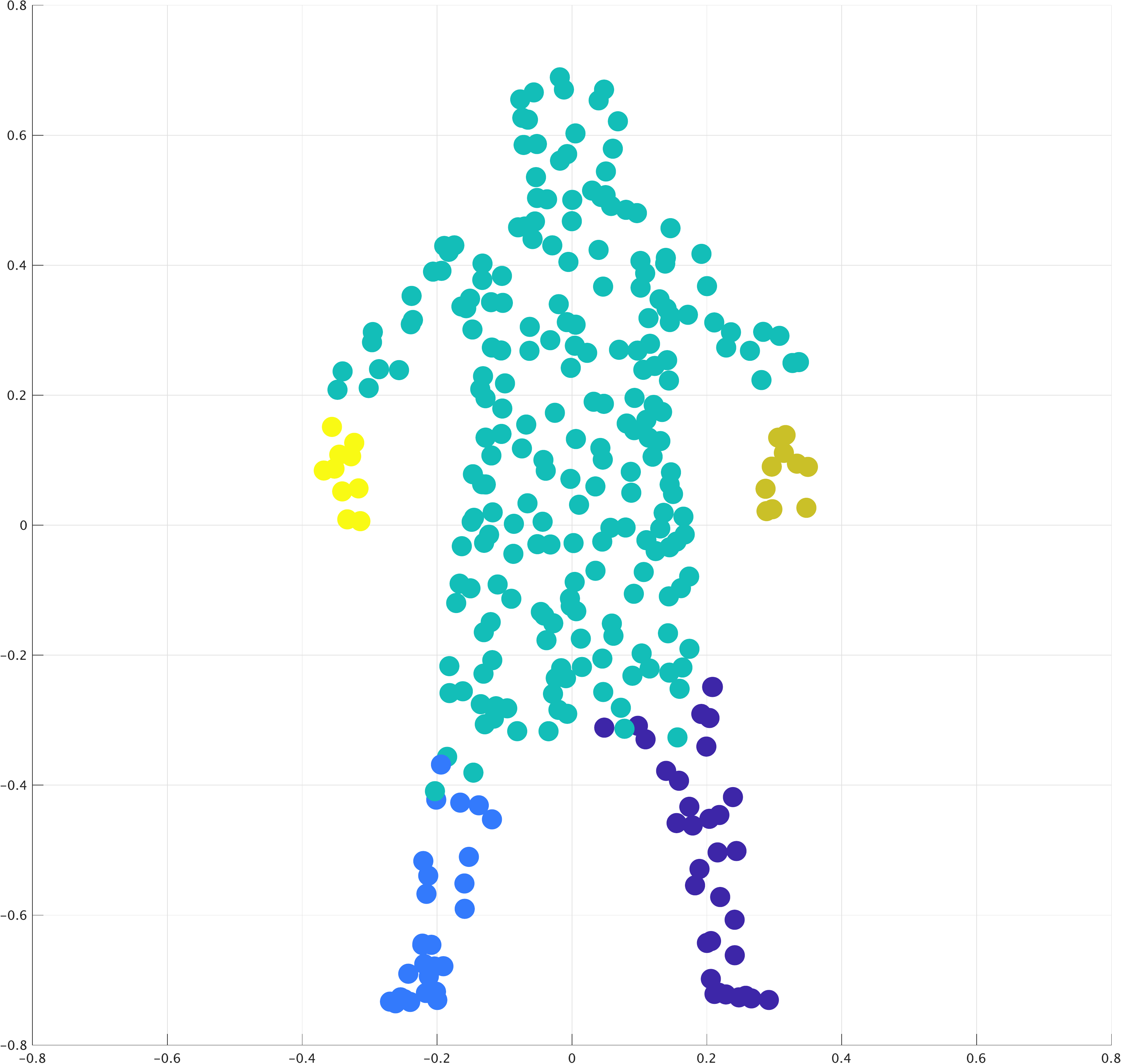}%
        }%
        \hspace{10pt}
    \subfigure[Proposed method]{%
        \includegraphics[clip, width=0.35\columnwidth]%
        {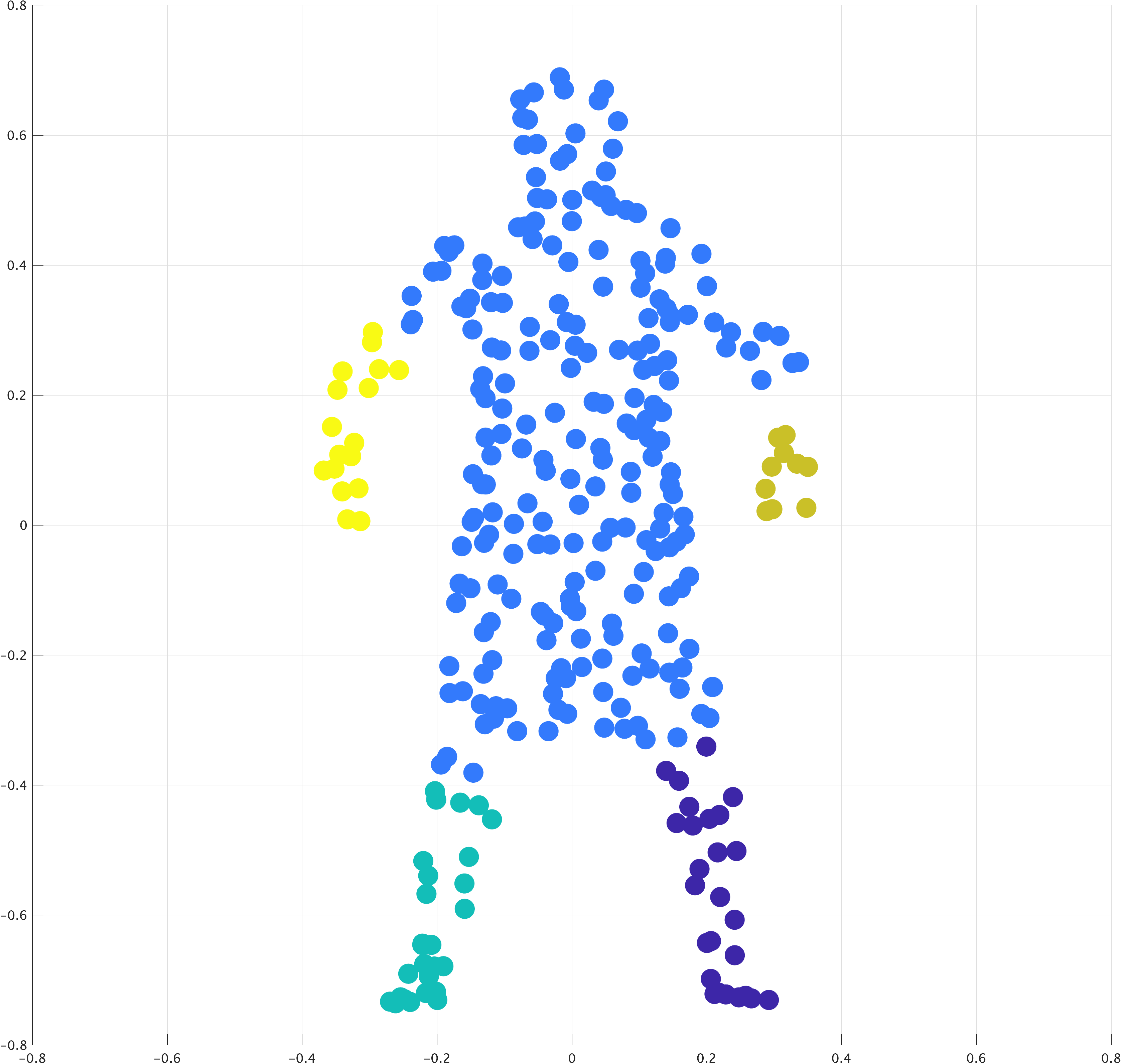}%
        }%
\\     
        \centering
    \subfigure[Graph at $t = 72$]{%
        \includegraphics[clip, width=0.35\columnwidth]%
        {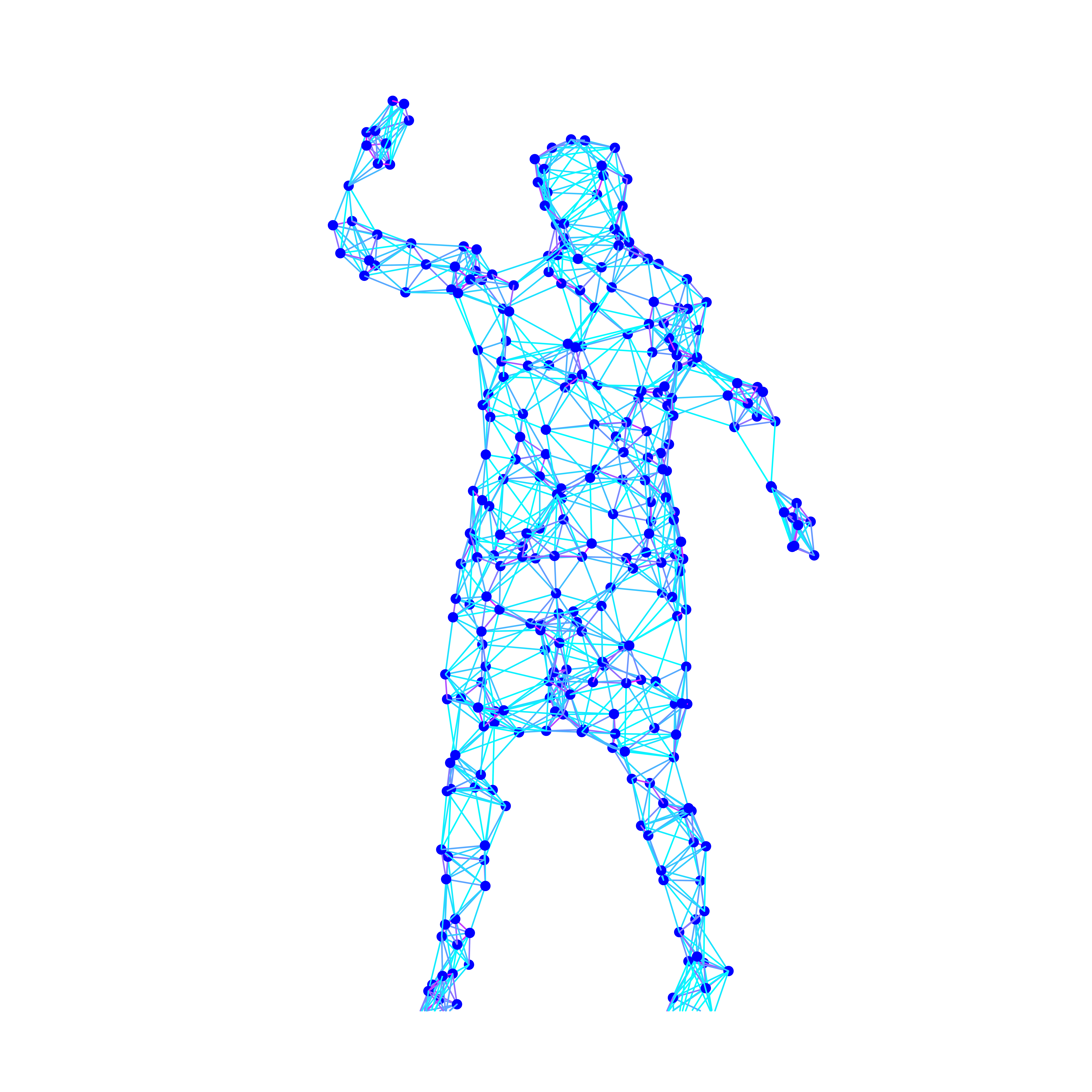}%
        }%
        \hspace{10pt}
    \subfigure[SC for static graph]{%
        \includegraphics[clip, width=0.35\columnwidth]%
        {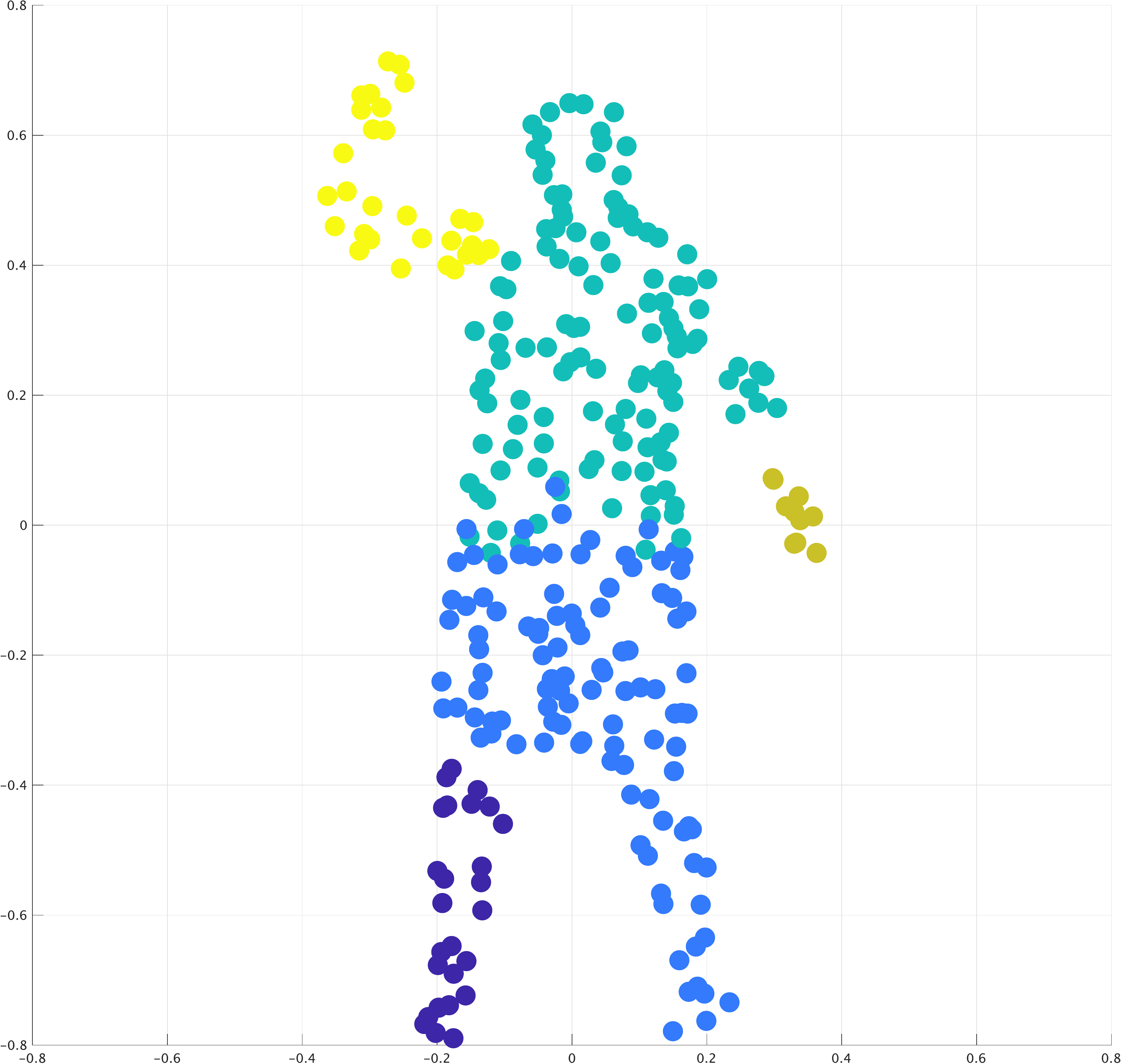}%
        }%
        \hspace{10pt}
    \subfigure[PisCES]{%
        \includegraphics[clip, width=0.35\columnwidth]%
        {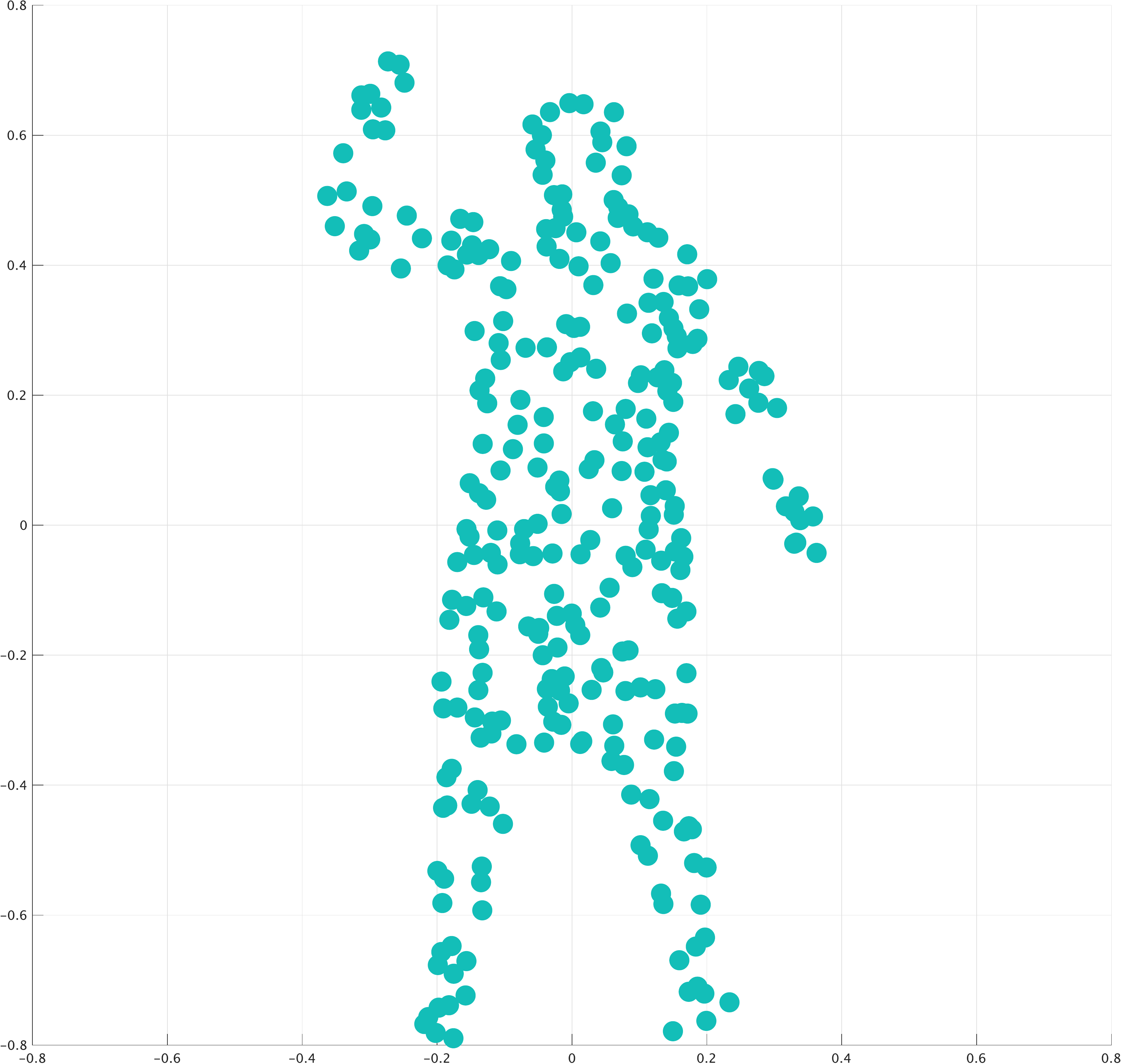}%
        }%
        \hspace{10pt}
            \subfigure[PisCES ($K$ is fixed)]{%
        \includegraphics[clip, width=0.35\columnwidth]%
        {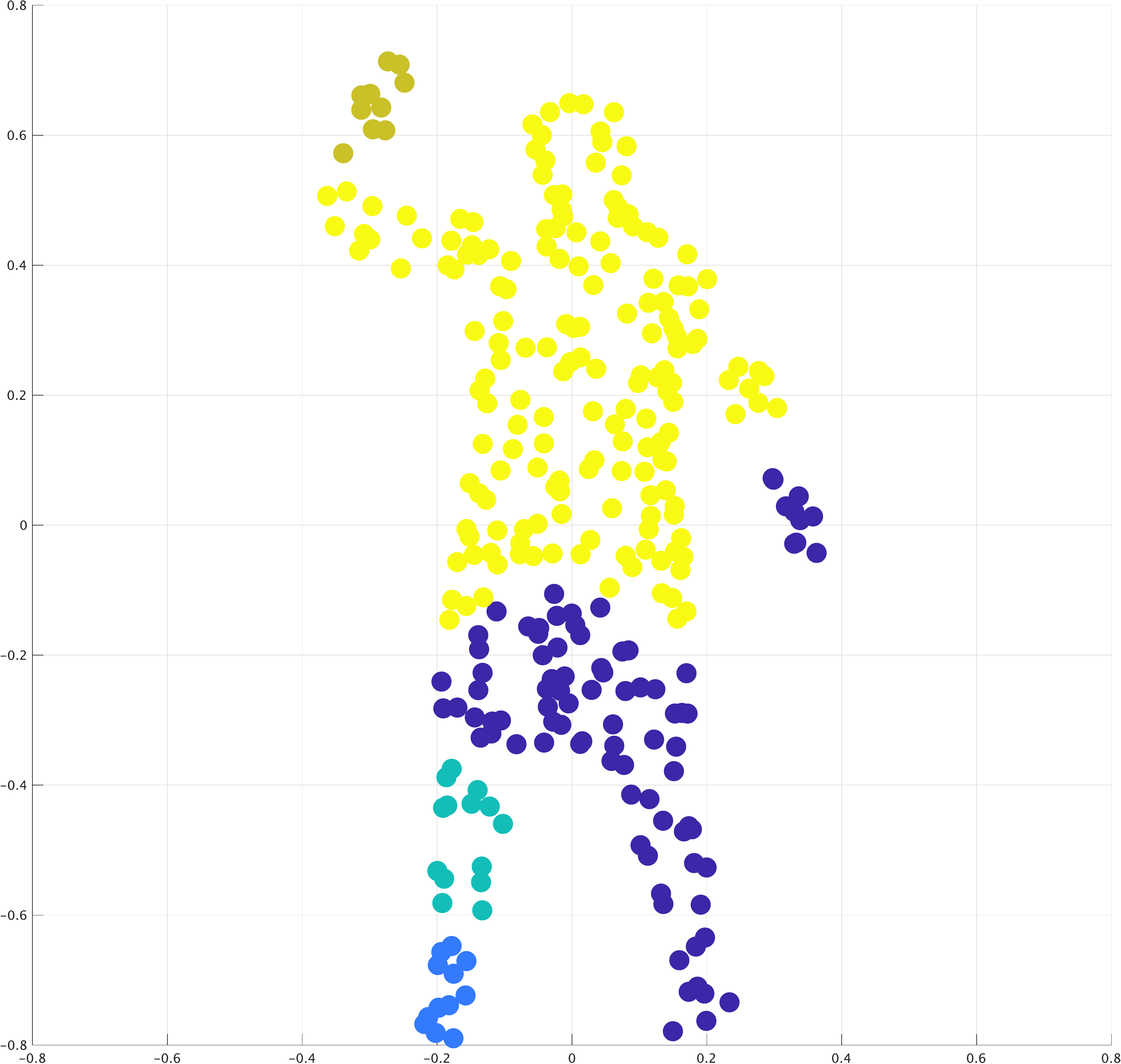}%
        }%
        \hspace{10pt}
    \subfigure[Proposed method]{%
        \includegraphics[clip, width=0.35\columnwidth]%
        {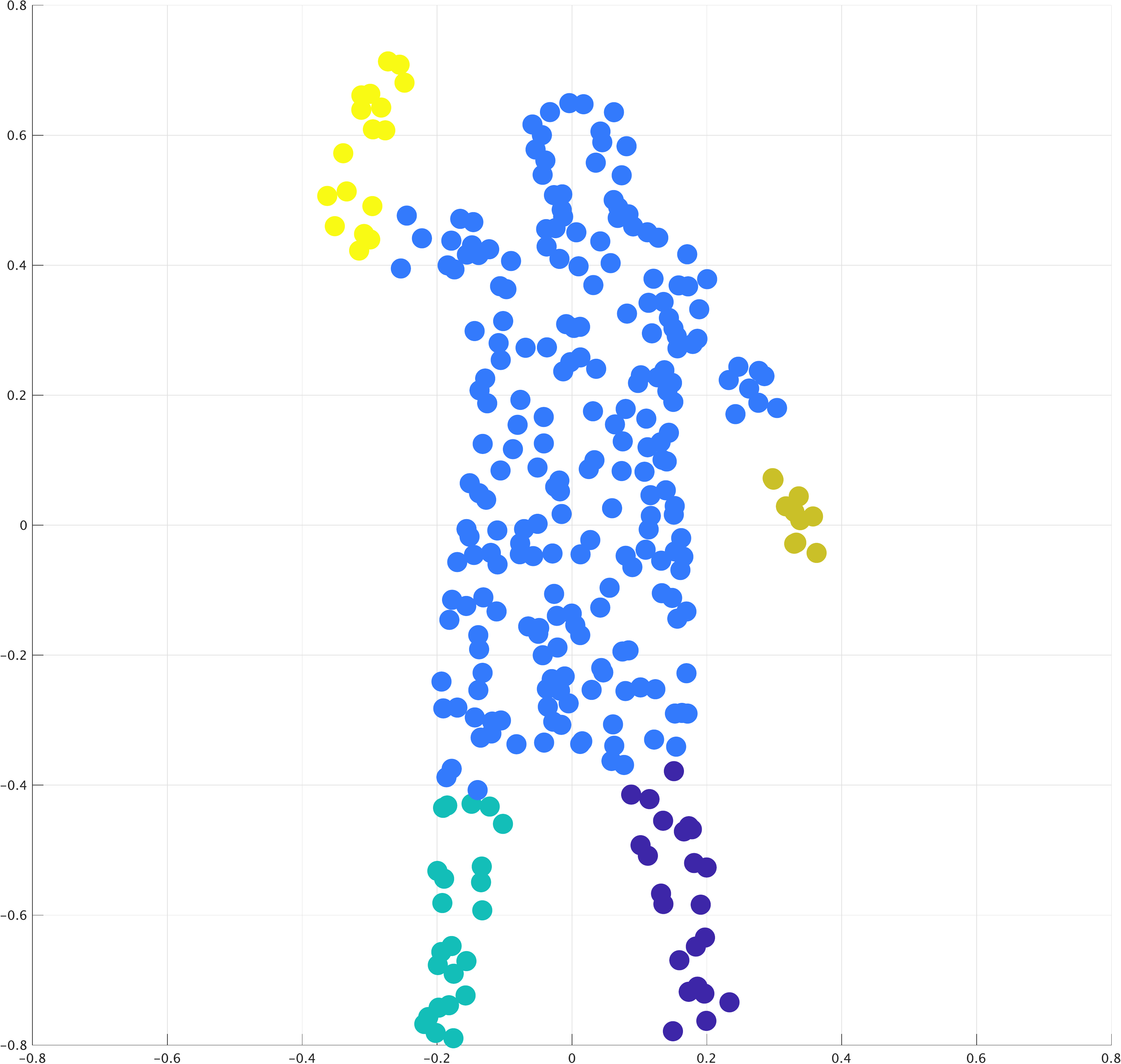}%
        }%
\\        
        \centering
    \subfigure[Graph at $t = 144$]{%
        \includegraphics[clip, width=0.35\columnwidth]%
        {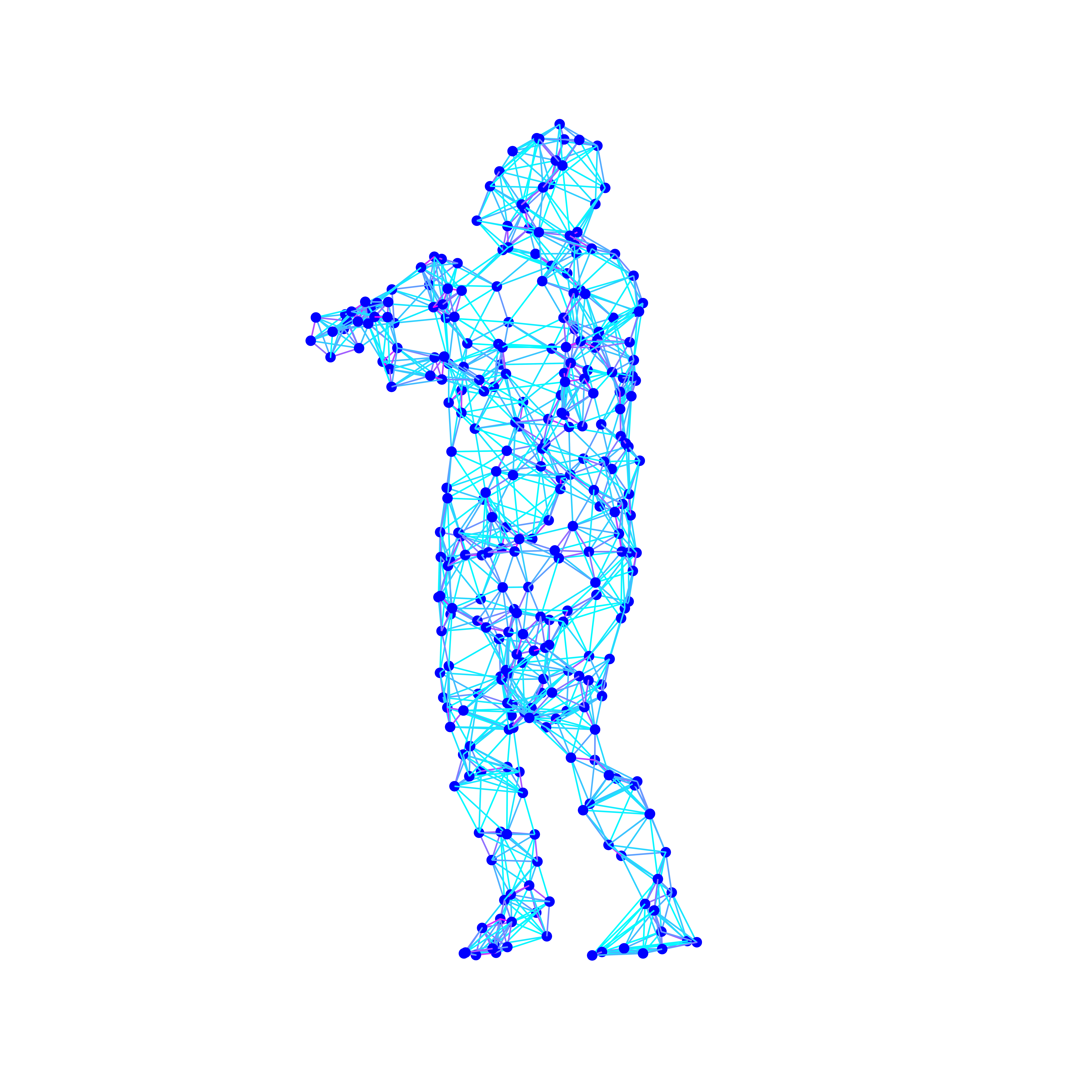}%
        }%
        \hspace{10pt}
    \subfigure[SC for static graph]{%
        \includegraphics[clip, width=0.35\columnwidth]%
        {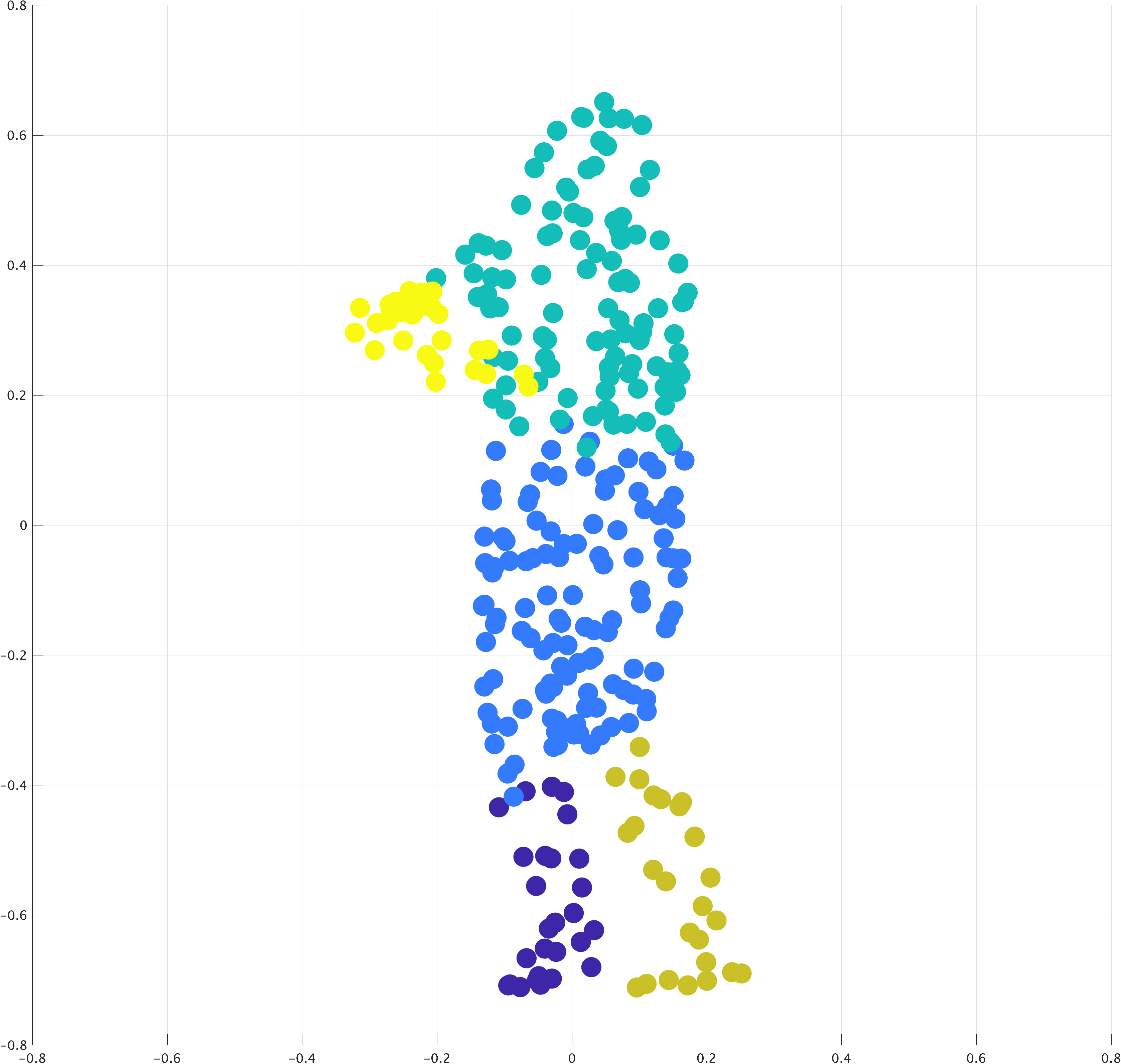}%
        }%
        \hspace{10pt}
    \subfigure[PisCES]{%
        \includegraphics[clip, width=0.35\columnwidth]%
        {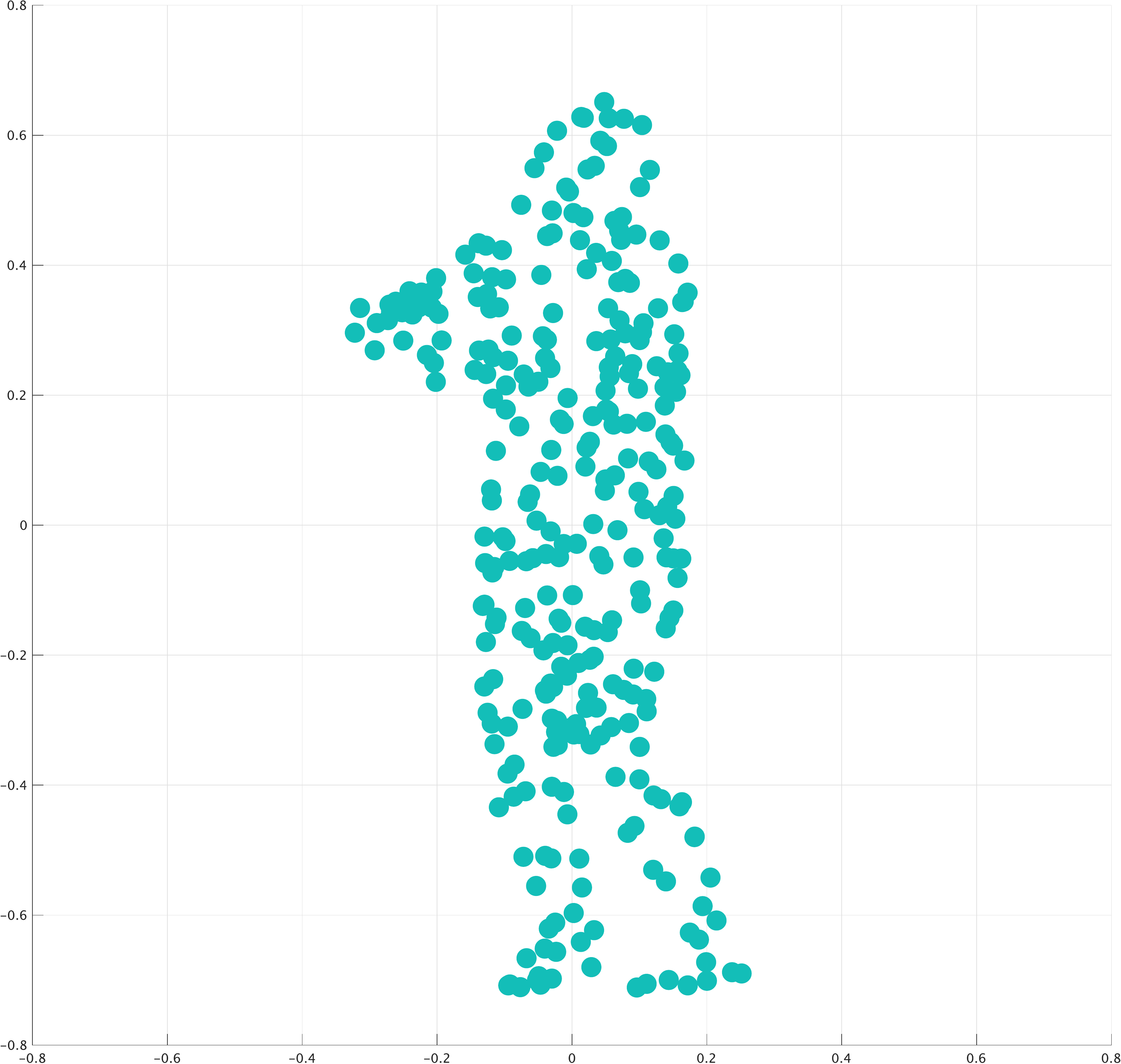}%
        }%
        \hspace{10pt}
            \subfigure[PisCES ($K$ is fixed)]{%
        \includegraphics[clip, width=0.35\columnwidth]%
        {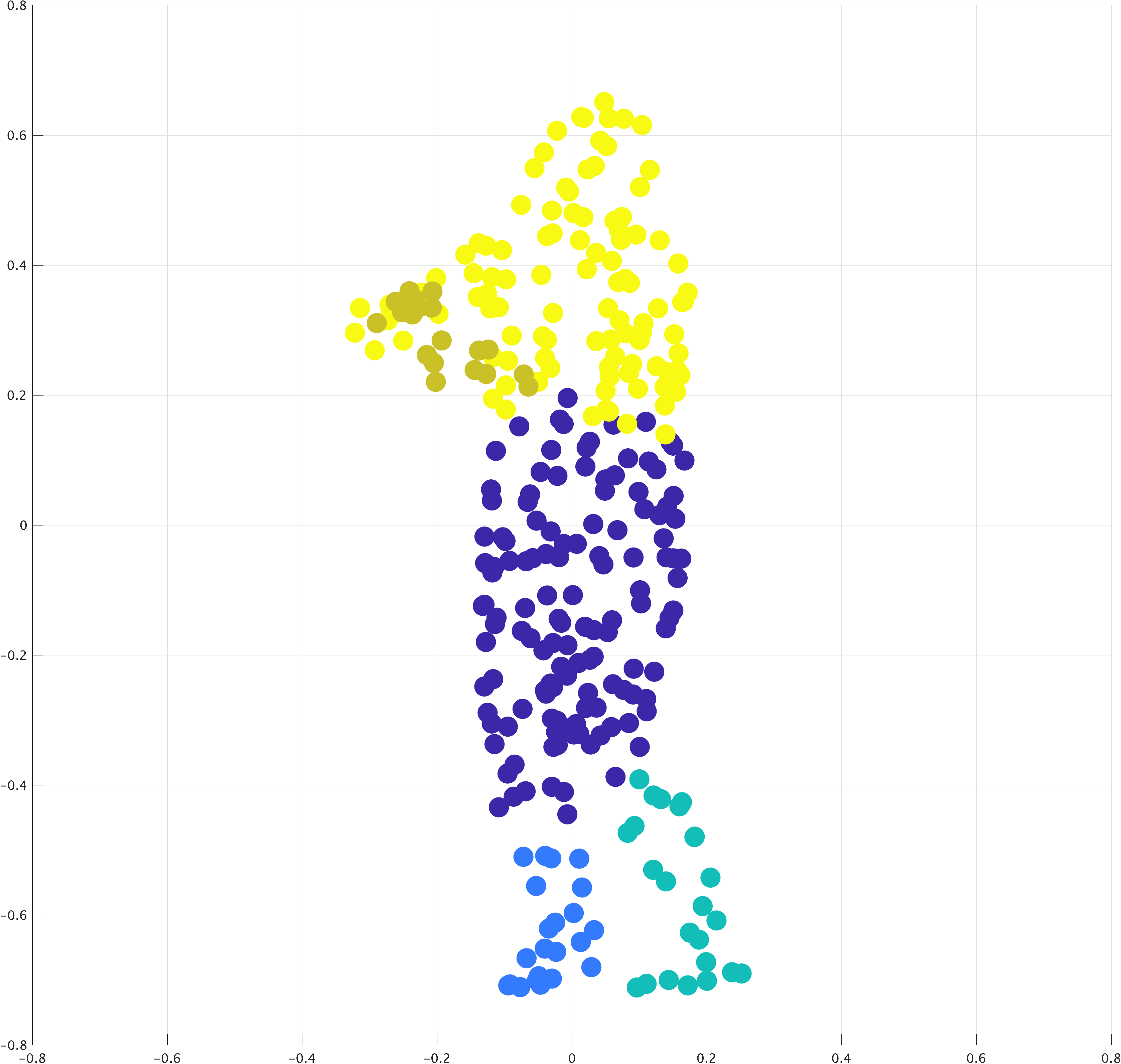}%
        }%
        \hspace{10pt}
    \subfigure[Proposed method]{%
        \includegraphics[clip, width=0.35\columnwidth]%
        {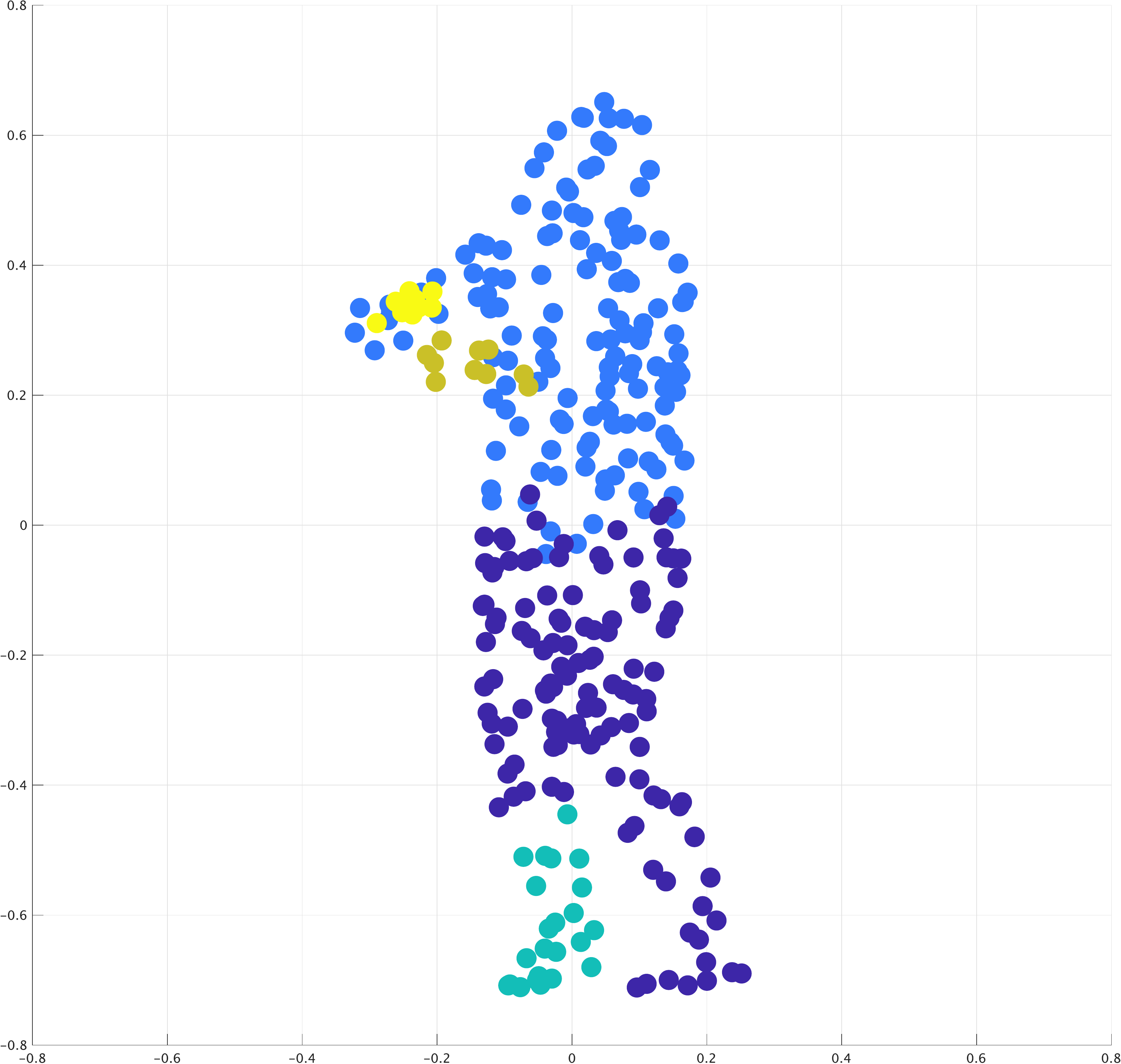}%
        }%
    \caption{Experimental results for dynamic point clouds (dancing). Node colors indicate the cluster labels. (a)-(e): $t=1$. (f)-(j): $t=72$. (k)-(o): $t=144$.}
    \label{fig:dance}
\end{figure*}

\begin{figure}[t]
\centering \relax
\includegraphics[width=80mm]{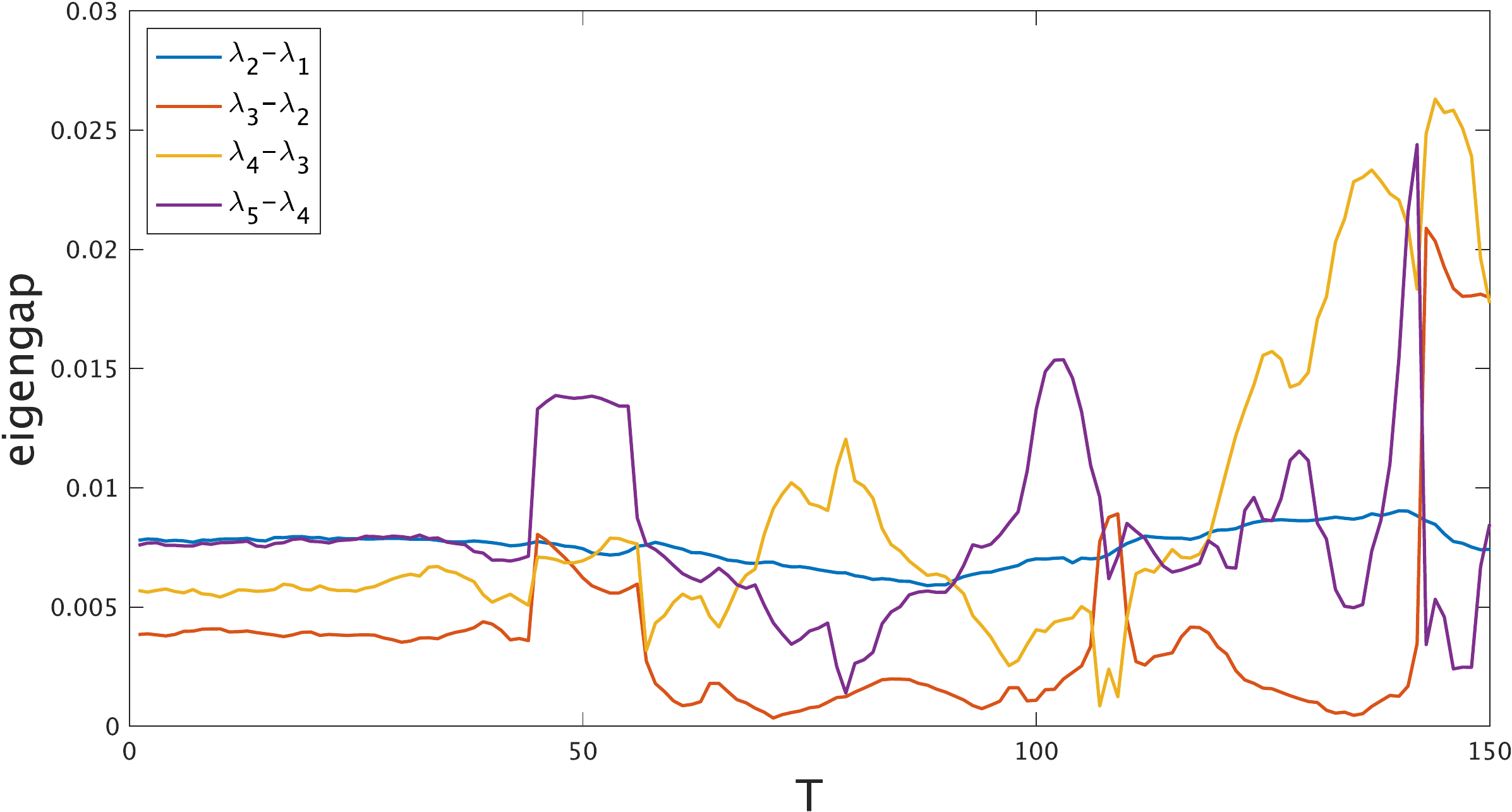}
\caption{Examples of eigengap of graph in Fig. \ref{fig:wheel}. $\lambda_i$ is the $i$-th smallest eigenvalue.}
\label{fig:egapwheel}
\end{figure}

 In this section, we evaluate the performance of the proposed method through experiments using synthetic and real-world data.

\subsection{SYNTHETIC DATA}
\subsubsection{SETUP}
First, we perform TV graph clustering for synthetic graphs.
For a clustering purpose, TV graphs based on the stochastic block model (SBM) \cite{hollandStochasticBlockmodelsFirst1983} are generated in the following manner.
SBM is a well-known random graph model where intra- and inter-cluster edges are generated randomly according to the predefined edge probabilities.
The intra-cluster edge probability $p_{\text{intra}}$ is basically larger than that for inter-clusters $p_{\text{inter}}$.
We set the number of clusters ($K$) and that of frames ($T$) are set to $K=3$ and $T=100$, respectively.

In this experiment, we generated two types of TV graphs with different edge connection probabilities.
A set of TV graphs $\{\mathcal{G}_t = (V_t, E_t)\}_{t=1}^T$ is generated as follows.

\begin{enumerate}
    \item Under given edge connection probabilities $p_{\text{intra}}$ and $p_{\text{inter}}$, $\mathcal{G}_1$ is generated by the (static) SBM model having three equisized clusters with $50$ nodes, i.e., $N=150$.
    \item For $t>1$, $\mathcal{G}_t$ is yielded from $\mathcal{G}_{t-1}$ in the following manner.
    \begin{enumerate}
        \item \textit{Node labels} of $V_t$ are randomly changed from that of $V_{t-1}$ with a probability of $0.01$.
        \item The edge set $E_t$ is obtained by the static SBM model with $p_{\text{intra}}$ and $p_{\text{inter}}$.
    \end{enumerate}
\end{enumerate}
Therefore, the synthesized TV graphs have a small $\operatorname{mismatch}(A_t,A_{t-1})$ in \eqref{mismatch} while the edges are randomly changed between $\mathcal{G}_t$ and $\mathcal{G}_{t-1}$.
Fig. \ref{fig:generate} shows overview of synthetic TV graphs.

For the first TV graph, the edge connection probabilities are set to $p_{\text{intra}} = 0.3$ and $p_{\text{inter}} = 0.2$.
This results in a set of TV graphs with high edge density.

For the second graph, those are set to $p_{\text{intra}} = 0.1$ and $p_{\text{inter}} = 0.05$.
The second graph is sparser than the first one.

\subsubsection{ACCURACY MEASURE}
 For calculating the clustering accuracy, we use the method in \cite{banerjeeModelbasedOverlappingClustering2005}.
First, we create a matrix $\mathbf{P} \in \mathbb{R}^{N \times N}$ that indicates clusters as follows:
\begin{equation}
  [\mathbf{P}]_{mn}=\begin{cases}
1 & \text { if $v_m$ and $v_n$ belong to the same cluster,} \\
0 & \text { otherwise}.
\end{cases}\label{eqn:P}
\end{equation}
Second, we calculate the correct classification ratio by comparing $\mathbf{P}$ of the ground-truth graph and its estimation as follows:

\begin{align}
    \text{accuracy} &= \frac{\sum_{i,j=1}^{N} \text{count}_{ij}-N}{N(N-1)}, \label{eqn:accuracy}\\
    \text{count}_{ij} &= \begin{cases}
    1 & \text {if $[\mathbf{P}]_{ij} = [\tilde{\mathbf{P}}]_{ij}$}, \\
    0 & \text {otherwise,}
    \end{cases}
\end{align}
where $\tilde{\mathbf{P}}$ is the same as \eqref{eqn:P} but is created from the estimated clusters.

\subsubsection{RESULTS}
We compare the experimental results of the proposed method with SC for static graphs introduced in Section \ref{rel}-\ref{rel_sc} and PisCES \cite{liuGlobalSpectralClustering2018}: The state-of-the-art TV graph clustering method.

Fig. \ref{fig:sy1} visualizes the clustering results of the first set of TV graphs (denser one) where node colors indicate the cluster labels.
Figs. \ref{fig:sy1}(a) and (e) show the original graphs where each of the three chunks corresponds to one cluster.
Fig. \ref{fig:accdense} also compares the accuracy of clustering 100 randomly generated time-varying graphs.
The figure plots the average accuracy of each TV graph as a function of $t$.

As shown in Figs. \ref{fig:sy1}(b) and (f), SC for static graphs fail to extract accurate clusters because edge connection probabilities of intra- and inter-clusters are close to each other.
In contrast, the proposed method and PisCES estimate the almost correct clusters, which reflects the power of temporal evolution.
Fig. \ref{fig:accdense} also indicates that the accuracy of the proposed method is consistently higher than SC and comparable to PisCES.

Fig. \ref{fig:sy2} shows the clustering results of the set of the sparser TV graphs and the accuracy is compared in Fig. \ref{fig:acc}.
As in the denser version, SC for static graphs does not work accurately.
In contrast to the dense one, Figs. \ref{fig:sy2}(c) and (h) indicate PisCES estimates all nodes as one cluster.
The proposed method extracts the almost correct clusters as in the previous experiment.
In Fig. \ref{fig:acc}, the accuracy of the proposed method is consistently higher than existing methods.

\subsubsection{Discussion}
Here, we investigate the performance difference between the proposed method and PisCES.
PisCES automatically determines the number of clusters for each frame on the basis of the difference between the eigenvalues of the graph Laplacian \cite{liuGlobalSpectralClustering2018}.
Specifically, PisCES splits the graph into multiple clusters when \textit{eigengap} exceeds a certain threshold.
While the two sets of synthetic TV graphs only differ their edge densities, their eigenvalue characteristics are significantly different.

We compare the eigengaps of two versions of TV graphs
in Fig. \ref{fig:egap1}.
As shown in the figure, the eigengap in the sparser TV graphs is much smaller than that in the denser ones.

Here, we manually force to set $K = 3$ for PisCES as SC and the proposed method.
Figs. \ref{fig:sy2}(d) and (i) show the clustering results by PisCES with the manually set $K$.
Even with the fixed $K$, the clustering results for PisCES for the set of sparser TV graphs are similar to the static SC, which is also confirmed from Fig. \ref{fig:acc}.

\subsection{REAL-WORLD DATA}
\subsubsection{SETUP}
 For an experiment using real-world data, we perform clustering of dynamic point clouds.
We use dynamic point clouds in the dataset of \cite{gallMotionCaptureUsing2009}.
It contains point cloud data capturing several types of human motion.
In this experiment, we used two point cloud data: \textit{cartwheel} and \textit{dancing}.

First, the both point clouds are downsampled to $N = 301$.
Since the point cloud does not have a ground-truth graph, we connect the points by the $k$-nearest neighbor graph based on the 3-D Euclidean distance.
We set $k=8$ for both point clouds.
We use $T = 200$ consecutive frames for the experiment.
Examples of the TV graphs are shown in Figs. \ref{fig:wheel}(a) and (e) for \textit{cartwheel}, and Figs. \ref{fig:dance}(a), (e) and (i) for \textit{dancing}.
We set the number of clusters to $K=5$.

\subsubsection{RESULTS}
Figs. \ref{fig:wheel} and \ref{fig:dance} visualize the clustering results for the dynamic point clouds.

As shown in the figures, the proposed method almost extracts reasonable clusters for the both point clouds.
In contrast, SC sometimes merges legs and hands close to each other as shown in Figs. \ref{fig:wheel}(g) and \ref{fig:dance}(l).
As in the synthetic sparser graphs, PisCES predicts the number of clusters as one in its default setting.
This is due to the small eigengaps as shown in Fig. \ref{fig:egapwheel}.
Even if we force to set $K=5$ for PisCES, it merges legs (Fig. \ref{fig:wheel}(i)) and splits one leg into multiple clusters (Fig. \ref{fig:dance}(i)).

\section{CONCLUSION}
\label{con}
 In this paper, we propose a clustering method for a set of TV graphs taking into consideration the temporal label smoothness.
We formulate an optimization problem based on spectral clustering with a regularization term for the label temporal evolution.
Through the experiments with synthetic and real world point cloud data, it is observed that the proposed method extracts accurate clusters compared to the existing clustering methods.

\end{document}